%% file: main.tex
\documentclass[10pt]{article} 
\usepackage[accepted]{tmlr}

\input{math_commands.tex}

\usepackage{hyperref}
\usepackage{url}

\usepackage{natbib}
\usepackage{amsfonts}       
\usepackage{amsmath}        
\usepackage{amssymb}        
\usepackage{graphicx}       
\usepackage{bm}             
\usepackage{url}            
\usepackage{nicefrac}       
\usepackage{booktabs}       
\usepackage{microtype}      
\usepackage[symbol]{footmisc}
\usepackage{float}
\usepackage{subfigure}
\usepackage{wrapfig}
\usepackage{algorithm}
\usepackage{algorithmic}
\usepackage{array}          
\usepackage{enumitem}       

\usepackage{amsthm}
\usepackage{mathtools}
\usepackage{nccmath}

\newtheorem{definition}{Definition}

\title{DORA: Exploring Outlier Representations in Deep Neural Networks}

\author{\name Kirill Bykov \email kbykov@atb-potsdam.de\\\addr Understandable Machine Intelligence Lab\\
      Leibniz Institute for Agriculture and Bioeconomy (ATB), Potsdam, Germany\\
      Technical University of Berlin, Berlin, Germany
      \AND
      \name Mayukh Deb \email mayukhmainak2000@gmail.com \\
      \addr Independent\\
      \AND
      \name Dennis Grinwald 
      \email dennis.grinwald@tu-berlin.de \\
      \addr 
      Machine Learning Group \\
      Technical University of Berlin, Berlin, Germany\\
      BIFOLD -- Berlin Institute for the Foundations of Learning and Data, Berlin, Germany
      \AND
      \name Klaus-Robert M\"uller \email klaus-robert.mueller@tu-berlin.de \\
      \addr  Machine Learning Group \\
      Technical University of Berlin, Berlin,  Germany\\  
      BIFOLD -- Berlin Institute for the Foundations of Learning and Data, Berlin, Germany\\
      Department of Artificial Intelligence, Korea University, Seoul 136-713, Korea\\
Max Planck Institut für Informatik, 66123 Saarbrücken, Germany\\
Google Research, Brain Team, Berlin, Germany      
      \AND
      \name Marina M.-C. H\"ohne 
      \email mhoehne@atb-potsdam.de \\
      \addr Understandable Machine Intelligence Lab\\
      Leibniz Institute for Agriculture and Bioeconomy (ATB), Potsdam, Germany\\
      Department of Computer Science, University of Potsdam\\
      Department of Physics and Technology, UiT Arctic University of Norway\\
      BIFOLD -- Berlin Institute for the Foundations of Learning and Data, Berlin, Germany}

\def\month{06}  
\def\year{2023} 
\def\openreview{\url{https://openreview.net/forum?id=nfYwRIezvg}} 

\begin{document}

\maketitle
\begin{abstract}
Deep Neural Networks (DNNs) excel at learning complex abstractions within their internal representations. However, the concepts they learn remain opaque, a problem that becomes particularly acute when models unintentionally learn spurious correlations. In this work, we present \textit{DORA} (Data-agnOstic Representation Analysis), the first data-agnostic framework for analyzing the representational space of DNNs. Central to our framework is the proposed \textit{Extreme-Activation} (EA) distance measure, which assesses similarities between representations by analyzing their activation patterns on data points that cause the highest level of activation. As spurious correlations often manifest in features of data that are anomalous to the desired task, such as watermarks or artifacts, we demonstrate that internal representations capable of detecting such artifactual concepts can be found by analyzing relationships within neural representations. We validate the EA metric quantitatively, demonstrating its effectiveness both in controlled scenarios and real-world applications. Finally, we provide practical examples from popular Computer Vision models to illustrate that representations identified as outliers using the EA metric often correspond to undesired and spurious concepts.
\end{abstract}

\section{Introduction}

\label{introduction}

The ability of Deep Neural Networks (DNNs) to perform complex tasks and achieve \textit{state-of-the-art} performance in various fields can be attributed to the rich and hierarchical representations that they learn \cite{bengio2013representation}. Far beyond the handcrafted features that were inductively imposed by humans on learning machines in classical Machine Learning methods \cite{marr1978representation, jackson1986introduction, fogel1989gabor}, Deep Learning approaches exploit the network's freedom to learn complex abstractions. However, a prevalent concern remains that the nature of concepts, learned by the model remains unknown. The rapid progress in representation learning only exacerbates the issue of interpretability, since 
DNNs are frequently trained using self-supervised methodologies \cite{jaiswal2020survey, lecun2021self} on immense volumes of data \cite{brown2020language, bommasani2021opportunities}, which accelerated the unpredictability concerning the scope of possible learned concepts and their mutual relations \cite{goh2021multimodal}.

The increasing popularity of Deep Learning techniques across various fields, coupled with the difficulty of interpreting the decision-making processes of complex models, has led to the emergence of the field of Explainable AI (XAI) (e.g.~\cite{montavon2018methods,samek2019explainable, xu2019explainable, gade2019explainable,rudin2019stop,samek2021explaining}). Research within XAI has revealed that the internal representations that form the basis of DNNs are susceptible to learning harmful and undesired concepts, such as biases \cite{guidotti2018survey, pmlr-v108-jiang20a}, Clever Hans (CH) effects \cite{lapuschkin2019unmasking}, and backdoors \cite{anders2022finding}. These malicious concepts often are unnatural or anomalous in relation to the relevant concepts within the dataset. Examples include watermarks in the PASCAL 2007 image classification task \cite{lapuschkin2019unmasking}, Chinese logographic watermarks in ImageNet dataset \cite{li2022Dilemma}, colored band-aids in skin-cancer detection problem \cite{anders2022finding} or tokens in a pneumonia detection problem \cite{zech2018variable}.

To enhance our understanding of the decision-making processes within complex machines and to prevent biased or potentially harmful decisions, it is crucial to explain the concepts learned during training. By analyzing the relationships between internal neural representations, we can gain insights into the model's predictive strategies. In this work, we introduce \textit{Representation Analysis}, a framework dedicated to exploring the representations of a particular model layer. Our approach utilizes a proposed \textit{Extreme-Activation} (EA) metric, which measures the similarity between various learned representations within the networks by examining the common activation patterns on Activation-Maximisation Signals (AMS). These signals represent data points where the representations exhibit their highest activations and can be identified through either a \textit{data-aware} (natural) process from an existing data corpus \cite{borowski2020natural} or a \textit{data-agnostic} process, in which the signals are synthetically generated \cite{erhan2009visualizing, olah2017feature}. We refer to the representation analysis conducted with the latter method as \textit{DORA}\footnote{PyTorch implementation of the proposed method can be found by the following link:
\url{https://github.com/lapalap/dora}
.} (Data-agnOstic Representation Analysis). We demonstrate the interpretability of our proposed distance measure and study the connections between natural and synthetic Activation-Maximisation Signals. Moreover, we quantitatively assess the alignment between the functional EA distance and human perception --- we demonstrate that EA distances between representations generally align with human judgment regarding the similarity of concepts, particularly in scenarios where the concepts underlying the representations are known. Additionally, we highlight our proposed distance measure's ability to establish a robust baseline for detecting inserted anomalous concepts in controlled scenarios. Lastly, through practical experiments conducted on popular Computer Vision models, we reveal that anomalous representations identified by our framework often correspond to undesirable spurious concepts.

\section{Related Work}

To address the concerns regarding the black-box nature of complex learning machines \cite{baehrens2010explain,vidovic2015opening,buhrmester2019analysis,samek2021explaining}, the field of \textit{Explainable AI (XAI)} has emerged. While some recent research focuses on inducing the self-explaining capabilities through changes in the architecture and the learning process \cite{gautam2022protovae, sg_isbi, chen2018looks,gautam2021looks}, the majority of XAI methods (typically referred to as \textit{post-hoc} explanation methods) are decoupled from the training procedure. A dichotomy of post-hoc explanation methods could be performed based on the scope of their explanations, i.e., the model behavior can be either explained on a \textit{local} level, where the decision-making strategy of a system is explained for one particular input sample, or on a \textit{global} level, where the aim is to explain the prediction strategy learned by the machine across the population and investigate the purpose of its individual components in a universal fashion detached from single datapoints (similar to feature selection \cite{guyon2003introduction}).

\textit{Local} explanation methods typically interpret the prediction by attributing relevance scores to the features of the input signal, highlighting the influential characteristics that affected the prediction the most. Various methods, such as Layer-wise Relevance Propagation (LRP) \cite{bach2015pixel}, GradCAM \cite{Selvaraju_2019}, Occlusion \cite{zeiler2014visualizing}, MFI \cite{vidovic2016feature}, Integrated Gradient \cite{sundararajan2017axiomatic}, have proven effective in explaining Graph Neural Networks \cite{wang2021towards,tiddi2020directions} as well as Bayesian Neural Networks \cite{bykov2021explaining, brown2022using}. To further boost the quality of interpretations, several enhancing techniques were introduced, such as SmoothGrad \cite{smilkov2017smoothgrad, omeiza2019smooth}, NoiseGrad and FusionGrad \cite{bykov2021noisegrad}. Considerable attention also has been paid to analyzing and evaluating the quality of local explanation methods (e.g.~\cite{samek2016evaluating,hedstrom2022quantus, guidotti2021evaluating,binder2023shortcomings}). 
However, while the local explanation paradigm is incredibly powerful in explaining the decision-making strategies for a particular data sample, the main limitation of such methods is their inability to effectively investigate the unexplored behaviors of the models, such as the detection of previously unknown spurious correlations and computational shortcuts \cite{adebayo2022post}.

\textit{Global} explanation methods aim to interpret the general behavior of learning machines by investigating the role of particular components, such as neurons, channels, or output logits, which we refer to as representations. Existing methods mainly aim to connect internal representations to human understandable concepts, making the purpose and semantics of particular network sub-function transparent to humans. Methods such as Network Dissection \cite{bau2017network, bau2018gan} and Compositional Explanations of Neurons \cite{mu2020compositional} aim to associate representations with human-understandable concepts. They achieve this by examining the intersection between the concept-relevant information provided by a binary mask and the activation map of the corresponding representation. The MILAN method \cite{hernandez2021natural} generates a text description of the representation by searching for a text string that maximizes the mutual information with the image regions in which the neuron is active. 

\subsection{Activation-Maximisation Methods}

The family of Activation-Maximization (AM) \cite{erhan2009visualizing} methods aims to globally explain the concepts behind neurons by identifying the input that triggers maximal activation in a particular neuron or network layer, thereby visualizing the features learned. These inputs, which we will refer to as Activation-Maximization Signals (AMS), could be either natural, found in a \textit{data-aware} fashion by selecting a ``real'' example from an existing data corpus \cite{borowski2020natural}, or artificial, found in a \textit{data-agnostic} mode by generating a synthetic input through optimization \cite{erhan2009visualizing, olah2017feature, szegedy2013intriguing}.

In comparison to earlier synthetic AM methods, Feature Visualization (FV) \cite{olah2017feature} performs optimization in the frequency domain by parametrizing the image with frequencies obtained from the Fourier transformation. This reduces adversarial noise in resulting explanations (e.g. \cite{erhan2009visualizing, szegedy2013intriguing}) --- improving the interpretability of the obtained signals. Additionally, the FV method applies multiple stochastic image transformations, such as jittering, rotating, or scaling, before each optimization step, as well as frequency penalization, which either explicitly penalizes the variance between neighboring pixels or applies bilateral filters on the input.

\subsection{Spurious Correlations}

Deep Neural Networks are prone to learn spurious representations --- patterns that are correlated with a target class on the training data but not inherently relevant to the learning problem \cite{izmailov2022spurious}. Reliance on spurious features prevents the model from generalizing, which subsequently leads to poor performance on sub-groups of the data where the spurious correlation is absent (cf.~\cite{lapuschkin2016analyzing,lapuschkin2019unmasking,geirhos2020shortcut}). In the field of Computer Vision, such behavior could be characterized by the model's reliance on aspects such as an image’s background \cite{xiao2020noise}, object textures \cite{geirhos2018imagenet}, or the presence of semantic artifacts in the training data \cite{wallis2022clever, lapuschkin2019unmasking, geirhos2020shortcut, anders2022finding}. Artifacts can be added to the training data on purpose as Backdoor attacks \cite{gu2017badnets, tran2018spectral}, or emerge ‘‘naturally’’ and might persist unnoticed in the training corpus, resulting in \textit{Clever Hans effects} \cite{lapuschkin2019unmasking}.

Recently, XAI methods have demonstrated their potential in revealing the underlying mechanisms of predictions made by models, particularly in the presence of artifacts such as Clever Hans or Backdoor artifacts. Spectral Relevance analysis (SpRAy) aims to provide a global explanation of the model by analyzing local explanations across the dataset and clustering them for manual inspection \cite{lapuschkin2019unmasking}. While successful in certain cases \cite{schramowski2020making}, SpRAy requires a substantial amount of human supervision and may not detect artifacts that do not exhibit consistent shape and position in the original images. SpRAY-based Class Artifact Compensation \cite{anders2022finding} method allowed for less human supervision and demonstrated its capability to suppress the artifactual behavior of DNNs.
 
\subsection{Comparison of Representations}

The study of representation similarity in DNN architectures is a topic of active research. Numerous methods comparing network representations have been applied to different architectures, including Neural Networks of varying width and depth \cite{Nguyen2020}, Bayesian Neural Networks \cite{Grinwald2022}, and Transformer Neural Networks \cite{Raghu2021}. Some works \cite{Ramsay1984, laakso2000, kornblith2019, Nguyen2022} argue that the representation similarity should be based on the correlation of a distance measure applied to layer activations on training data. Other works \cite{Raghu2017, Morcos2018} compute similarity values by applying variants of Canonical Correlation Analysis (CCA) \cite{Hardoon2005, biessmann2010temporal} on the activations or by calculating mutual information \cite{Li2015}, or employ kernel methods to quantify the evolution of the representations \cite{montavon2011kernel,braun2008relevant}. However, these methods are predominantly utilized for the comparison of whole representation spaces, e.g. layers, and not individual components, and as a result often overlook the semantics of learned concepts. Furthermore, those methods are dependent on the availability of data.

\section{Distance Metrics between Neural Representations}
\label{sec:distance_metrics}
In the following, we start with the definition of a \textit{neural representation} as a sub-function of a given network that depicts the computation graph, from the input of the model to the output of a specific neuron.

\begin{definition}[Neural representation]
We define a neural representation $f$ as a real-valued function $\displaystyle f: \mathbb{D} \rightarrow \mathbb{R},$ mapping from the data domain $\mathbb{D}$ to the real numbers $\mathbb{R}$.
\end{definition}

The following definition is introduced to highlight the distinction between the traditional notion of a neuron and the broader computational process encapsulated in the term \textit{neural representation}. Conventionally, a neuron is defined as a function that takes inputs from its preceding neurons. In contrast, a neural representation describes the entire computational process, starting from the input and yielding the activation of a specific neuron (unit). While some neurons in DNNs produce multidimensional outputs, depending on the specific use cases, multidimensional functions could be regarded either as a set of individual representations or alternatively could be aggregated to achieve scalar output. For example, in the case of convolutional neurons that output activation maps containing the dot product between filter weights and input data at each location, activation maps could be aggregated by average- or max-pool operations for the sake of simplifying the explanation of the semantic concept underlying the function. The choice depends on the particular aim and scope of the analysis and does not alter the network itself.

The scalar output of representations often corresponds to the amount of evidence or similarity between concepts present in the input and internally learned abstractions. Various sub-functions within the model could be considered as neural representations, ranging from the neurons in the initial layers that are often regarded as elementary edge or color detectors \cite{le2021revisiting}, to the model output. Throughout this work, we primarily focus on the high-level abstractions that emerge in the latest layers of networks, such as the feature-extractor layers in well-known Computer Vision architectures, as they are frequently employed for transfer learning \cite{zhuang2020comprehensive}.

In DNNs, neural representations are combined into layers --- collections of individual neural representations that typically share the same computational architecture and learn abstractions of similar complexity. In the scope of the following work, we mainly focused on the analysis of the relations between representations within one selected layer from the network.

\begin{definition}[Layer]
We define a layer $\mathcal{F} = \left\{f_1, ..., f_k\right\}$ as a set comprising $k$ individual neural representations.
\end{definition}

To examine the relationships between representations, we can begin by analyzing the behavior of functions with respect to a given dataset. We define a dataset, $D$, consisting of $N$ data points denoted as $D = \{x_1, ..., x_N\}$. This set is referred to as the \textit{evaluation dataset} and is used to measure the relationship between two neural representations. We make the assumption that these data points, $x_1, ..., x_N$, are independently and identically distributed (i.i.d.) samples from the overall data distribution $\mathcal{D}$. Additionally, we standardize the activations of representations on this evaluation dataset, resulting in a mean of 0 and a standard deviation of 1. 

For a neural representation $f_i \in \mathcal{F}$ and an evaluation dataset $D= \left\{x_1,..., x_N\right\}$, we define a vector of activations 
\begin{equation}
    \mathbf{a}_i = \left(f_i(x_1), ..., f_i(x_N)\right),
\end{equation}
where
\begin{equation}
    \mu_{i} \vcentcolon= \frac{1}{N}\sum_{t = 1}^{N} f_i(x_t) = 0, \quad \sigma_i \vcentcolon= \sqrt{\frac{1}{N-1}\sum_{t = 1}^{N}\left(f_i(x_t) - \mu_{i}\right)^2} = 1.
\end{equation}

Standardizing the vectors in this way can help to mitigate any differences in scale between the vector components and ensure that each component contributes equally to the distance calculation. Below, we present three widely recognized metrics that can be utilized to measure the distance between neural representations.

\begin{itemize}
    \item \textbf{Minkowski distance}:    
    \begin{equation}
        d_M\left(f_i, f_j\right) = \left(\sum_{t = 1}^N\left|f_i(x_t)-f_j(x_t)  \right|^{p}\right)^{\frac{1}{p}},
    \end{equation}
    where $p \geq 1, p\in \mathbb{Z}$ determines the degree of the norm, which gauges the sensitivity of the metric to differences between the components of the vectors being compared. In general, larger values of $p$ lead to a greater emphasis on larger differences between the components of the vectors. Conversely, smaller values of p reduce the influence of larger differences, leveraging an increased uniform weighting of all components.
   
    \item \textbf{Pearson distance}:
    \begin{equation}
        d_P\left(f_i, f_j\right) = \frac{1}{\sqrt{2}} \sqrt{ 1 - \rho_p\left(\mathbf{a}_i, \mathbf{a}_j\right)},
    \end{equation}
    where $\rho_p(\mathbf{a}, \mathbf{b})$ is the Pearson correlation coefficient between the vectors $\mathbf{a}$ and $\mathbf{b}$. The Pearson correlation coefficient is a widely used metric for measuring the linear dependence between two random variables. It is an interpretable measure of similarity, however, it is also sensitive to outliers, which can significantly affect the calculated distance.
    
    \item \textbf{Spearman distance}:
    \begin{equation}
        d_S\left(f_i, f_j\right) = \frac{1}{\sqrt{2}} \sqrt{ 1 - \rho_s\left(\mathbf{a}_i, \mathbf{a}_j\right)},
    \end{equation}
    where $\rho_s(\mathbf{a}, \mathbf{b})$ is the Spearman rank-correlation coefficient between vectors $\mathbf{a}$ and $\mathbf{b}$. The Spearman correlation is a non-parametric rank-based metric commonly used to measure the monotonic dependence between two random variables. Its main advantage is that it is robust to outliers and can handle ties in the data.
    
\end{itemize}

\subsection{Data-Aware Extreme-Activation distance}

The meaning and semantics of individual representations are frequently characterized and explained by datapoints where these representations exhibit extreme values. This analysis typically concentrates on the most positively activating signals, primarily due to the prevalent use of bounded activation functions like ReLU \cite{glorot2011deep}. In these functions, positive activation values typically signify the existence of specific patterns within the input signal.

Given the evaluation dataset $D$ and a neural representation $f_i \in \mathcal{F},$ we define a collection of natural Activation-Maximisation signals (n-AMS) as follows:

\begin{definition}[n-AMS]
    \label{def:n_ams}
    Let $f_i \in \mathcal{F}$ be a neural representation, and $D = \left\{x_1,..., x_N\right\} \subset \mathbb{D}$ be an evaluation dataset with $N$ datapoints. Assume that the dataset $D$ could be split in $n$ disjoint blocks $D = \bigcup_{i = 1}^n D_t, D_t = \left\{x_{td+1},..., x_{(t+1)d+1}\right\}, \forall t \in \{0, ..., n-1\}$ of length $d$.
    
    We define a collection of $n$ natural Activation-Maximisation signals (n-AMS) as $S_i = \left\{s^i_1,..., s^i_n\right\}, $ where
    \begin{equation}
        s^i_t = \argmax_{x\in D_t} f_i\left(x\right), \forall t \in \{0, ..., n-1\}.
    \end{equation}
\end{definition}

\begin{figure*}[t]
\begin{center}
\centerline{\includegraphics[width=\textwidth]{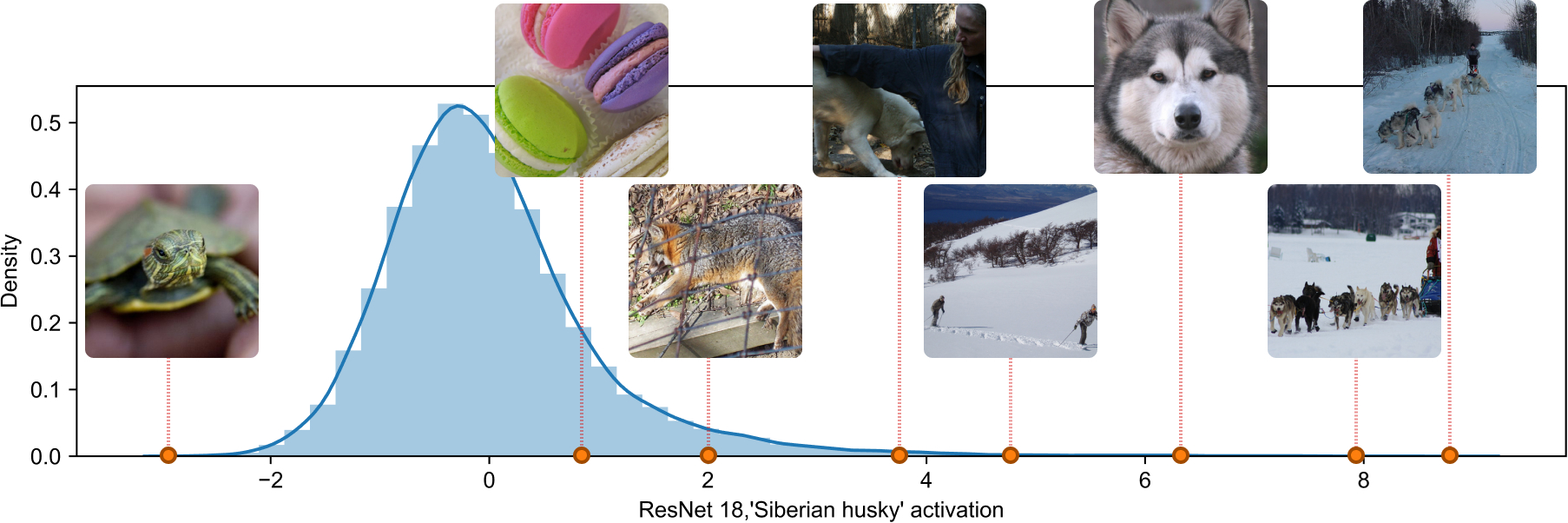}}

\caption{\textbf{Distribution of activations for the ``Siberian husky'' representation.} From the figure we can observe the standardized activation distribution of the ``Siberian husky'' logit from the ResNet18 model trained on ImageNet. The data was collected across the ILSVRC-2012 validation dataset. Additionally, various input images are dispalyed over their respective activations. Analyzing the activation of representations can provide crucial insights into the behavior of the model. For instance, we observe that the model achieves extremely high activations when there are multiple dogs in the image, corresponding to the ``Dogsled'' class. However, we also observe a potential spurious correlation, where the model assigns high scores to images with a snowy background.}
\label{fig:n-ams_husky_distribution}
\end{center}
\end{figure*}

The suggested definition of n-AMS diverges from the conventional method for explaining the concepts behind the representation by analyzing signals with the highest activations, that is, signals whose activations rank highest across the entire dataset. Instead, n-AMS could be seen as samples from the Extreme Value distribution, thereby allowing for statistical properties to be considered. Additionally, such ditribution is parameterized with parameter $d$, referred to as the \textit{depth}, which represents the size of the subset from which the signal is obtained. Note that we could examine the highest activation signal of the whole dataset by setting $n = 1$ and $d = N$, however, interpreting the representation's semantics by using only one signal might be misleading. Figure \ref{fig:n-ams_husky_distribution} illustrates the distribution of activations of the ``Siberian husky'' logit from the ResNet18 model trained on ImageNet \cite{he2016deep} across all the images from the ILSVRC-2012 validation dataset \cite{russakovsky2015imagenet}, where we can observe that the most activating signal corresponds to the ``Dogsled'' class. In light of this, we aim to sample several n-AMS from separate data subsets.

We propose that by examining how two neural representations activate each other's n-AMS, we can gain significant insights into the similarity of the learned abstractions. For this, we first introduce the \textit{representation activation vectors} (RAVs).

\begin{definition}
\label{def:ravs}
Let $\mathcal{F} = \{f_1, ..., f_k\}$ be a layer including $k$ neural representations, and $\mathcal{S} = \{S_1,..., S_k\}$ be a collection of $n$ n-AMS for each of the $k$ representations in the layer. For $\forall i, j \in \{1, ..., k\}$ we define $\mu^{i}_j = \frac{1}{n}\sum_{t=1}^n f_j\left(s^i_t\right)$ as mean activation of $f_j$ given the n-AMS of $f_i.$

For any two representations $f_i, f_j \in \mathcal{F}$, we define their \textit{pair-wise} representation activation vectors (RAVs) $r_{ij}, r_{ji}$ as:

    \begin{equation}
        r_{ij} = \begin{pmatrix}
           \mu^{i}_i \\
           \mu^{i}_j
         \end{pmatrix},
         \quad
        r_{ji} = \begin{pmatrix}
           \mu^{j}_i \\
           \mu^{j}_j
         \end{pmatrix}.
    \end{equation}
    
In addition, for each neural representation $f_i\in \mathcal{F}$, we define the corresponding \textit{layer-wise} RAV as follows:

\begin{equation}
r_{i*} = \begin{pmatrix}
            \mu^{i}_1 \\
           \vdots \\
           \mu^{i}_k
         \end{pmatrix}.
\end{equation}
\end{definition}

Intuitively, the idea behind RAVs is to capture how one representation's n-AMS are perceived by other representations. RAVs capture the direction of n-AMS signals within two-dimensional vectors when dealing with pair-wise vectors, encoding the information about how two neural representations respond to each other's stimuli. In the layer-wise case, the vectors are $k$-dimensional, utilizing all representations within the layer as descriptors. In practice, to compute RAVs, $n$ n-AMS are gathered for each representation within the layer, inferenced by the model. Then, activations across the representations from the layer are collected and averaged.

\begin{figure*}[t]
\begin{center}
\centerline{\includegraphics[width=\textwidth]{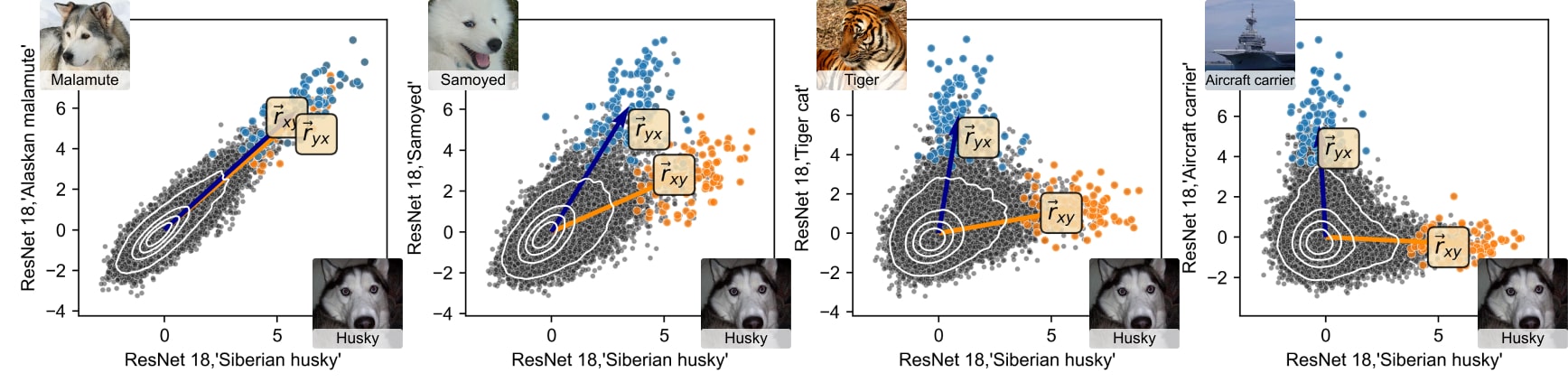}}
\vspace{-0.25cm}
\caption{\textbf{Joint activation and pair-wise RAVs of different ImageNet representations.} Four scatter plots illustrate the joint activations, n-AMS sets, and pair-wise RAVs for distinct pairs of neural representations. Blue points represent the n-AMS of the representation depicted on the vertical axis, while orange points represent the n-AMS of the representation shown on the horizontal axis. The representations, taken from the ResNet18 output logit layer, include ``Alaskan malamute'', ``Samoyed'', ``Tiger cat'', and ``Aircraft carrier'', each compared with the ``Siberian husky'' representation. We can observe that the angle between RAVs reflects the visual similarity between classes, representations were trained to learn: the RAVs of ``Siberian husky'' and ``Alaskan malamute'' are almost collinear due to the high visual similarity between the two dog breeds, while the RAVs of ``Siberian husky'' and ``Aircraft carrier'' are orthogonal, indicating their visual dissimilarity.}
\label{fig:n-ams_husky}
\end{center}
\end{figure*}

 To illustrate the concept of Representation Activation Vectors, we calculated n-AMS for five distinct neural representations extracted from the output layer of the ImageNet pre-trained ResNet18 model. These representations corresponded to the classes ``Siberian husky'', ``Alaskan malamute'', ``Samoyed'', ``Tiger cat'', and ``Aircraft carrier'', which were selected manually to demonstrate the decreasing visual similarity between the classes and the ``Siberian husky'' class. Using the ILSVRC-2012 validation dataset, we computed the signals with a sample size of $n=100$ and a subset size of $d=500$. Figure \ref{fig:n-ams_husky} presents a scatter plot of activation values across datapoints and pair-wise RAVs. Our results indicate that the angle between these vectors increases with the visual dissimilarity between the classes.

To measure the distance between neural representations that reflect the similarity of learned concepts, we introduce a novel distance metric known as the \textit{Extreme-Activation} distance. This metric assesses the similarity between two neural representations based on the angle between their Representation Activation Vectors. Given that the computation of this distance measure is performed in a \textit{data-aware} mode and relies on the presence of a dataset, we refer to this distance as the natural Extreme-Activation distance or EA$_n$.

\begin{definition}[Extreme-Activation distance]
\label{def:EAn}
    Let $f_i, f_j \in \mathcal{F}$ be two neural representations, and $r_{ij}, r_{ji}$ be their pair-wise RAVs. We define a pair-wise Extreme-Activation distance as

    \begin{equation}
        d^p_{{EA}_n}\left(f_i, f_j\right) = \frac{1}{\sqrt{2}}\sqrt{1 - \cos\left(r_{ij}, r_{ji}\right)},
    \end{equation}
    where $\cos(A, B)$ is the cosine of the angle between vectors $A, B.$

    Additionally, we define layer-wise Extreme-Activation distance between $f_i, f_j$ as 

    \begin{equation}
        d^l_{EA_n}\left(f_i, f_j\right) = \frac{1}{\sqrt{2}}\sqrt{1 - \cos\left(r_{i*}, r_{j*}\right)}.
    \end{equation}
\end{definition}

\subsection{Synthetic Extreme-Activation distance}

Although data-aware distance metrics can offer insight into the relationships between representations, their dependence on the data can be viewed as a limitation potentially acting as a bottleneck when analyzing the relationships between a model's internal representations. Modern machine learning models are often trained on closed-source or very large datasets, making it difficult to obtain the exact dataset the model was trained on. If the evaluation dataset, i.e. the dataset utilized for n-AMS sampling lacks concepts that were present in the training data, the resulting n-AMS could potentially be misleading. This is due to the fact that they might not encapsulate the features that the representation has learned to detect, simply because such features are absent in the dataset.

To alleviate the dependence on data, we propose a \textit{data-agnostic} method for computing the Extreme-Activation distance. This approach employs synthetic Activation-Maximization signals, denoted as s-AMS, in place of n-AMS. The s-AMS signals are generated by the model itself through an optimization process, thereby eliminating the need for external generative models or datasets.

\begin{definition}[s-AMS]
Let $f_i \in \mathcal{F}$ be a neural representation.  Synthetic Activation-Maximization (s-AMS) signal $\tilde{s}^i$ is defined as a solution to the following optimization problem:
\begin{equation}
\tilde{s}^i = \argmax_{\tilde{s} \in \Theta} f_i(\tilde{s}),
\end{equation}
where $\Theta$ denotes the set of potential solutions, typically defined by the particular signal parametrization employed for optimization.
\end{definition}

Generating s-AMS for a neural representation is a non-convex optimization problem \cite{nguyen2019understanding} that typically employs gradient-based methods \cite{erhan2009visualizing, nguyen2015innovation, olah2017feature}. Starting from a random noise parametrization of input signals, the gradient-ascend procedure searches for the optimal set of signal parameters that maximize the activation of a given representation. Early methods employed standard pixel parametrization \cite{erhan2009visualizing}, while modern approaches used Generative Adversarial Network (GAN) generators \cite{nguyen2016synthesizing} or Compositional Pattern Producing Networks (CPPNs) \cite{mordvintsev2018differentiable, stanley2007compositional}. In this study, we use the Feature Visualization method \cite{olah2017feature} for s-AMS generation, which parametrizes input signals by frequencies and maps them to the pixel domain using Inverse Fast Fourier Transformation (IFFT). This method is popular for its simplicity and independence from external generative models, as well as for its ability to be human-interpretable \cite{olah2020naturally, goh2021multimodal, cammarata2020curve}.

\begin{wrapfigure}{r}{0.5\textwidth}
\vspace{-0.4cm}
\includegraphics[width=0.5\textwidth]{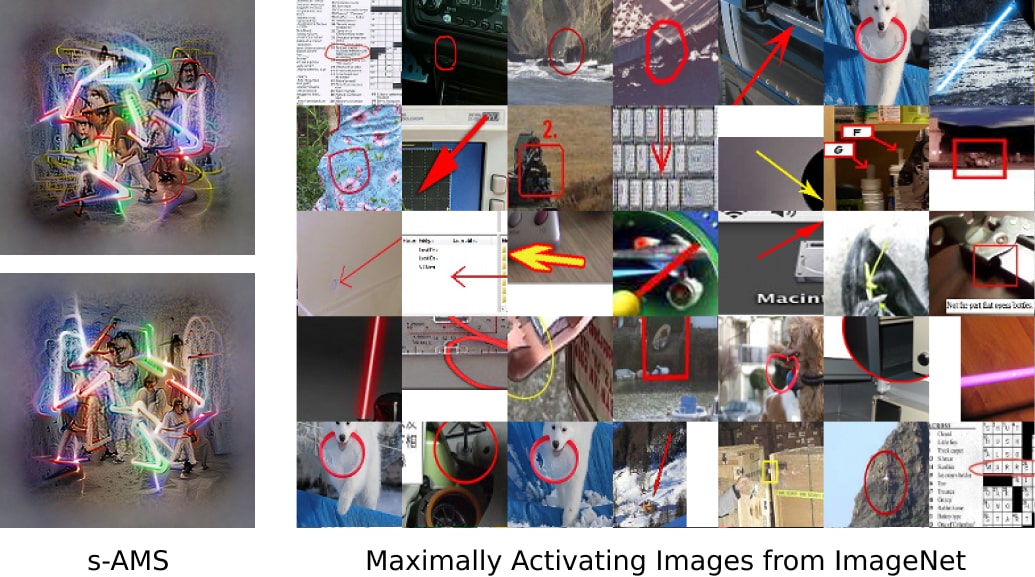}
\vspace{-0.5cm}
\caption{\textbf{Failing to explain ``Star Wars'' representation with natural images.} Comparison of the s-AMS (left) and most activated images collected from the ImageNet dataset (right) for unit 744 in the last convolutional layer of the CLIP ResNet50 model. Due to the inaccessibility of the training dataset and lack of specific images due to copyright restrictions, natural images do not reveal the underlying concept of the ``Star Wars'' neuron.
}
\label{fig:starwars}
\end{wrapfigure}

The optimization procedure for s-AMS generation has several adjustable hyperparameters, including the optimization method and transformations applied to signals during the procedure. A key parameter is the number of optimization steps (or epochs) denoted as $m$. This parameter can be seen as analogous to the parameter $d$ used in n-AMS generation. Since different random initializations in the parameter space can lead to the convergence of s-AMS generation into different local solutions, the resulting s-AMS can vary. This variability mirrors that observed when sampling n-AMS. Therefore, analogous to Definition \ref{def:n_ams}, for the representation $f_i$, we introduce a set of s-AMS, comprised of $n$ signals, denoted as $\tilde{S}_i = \{\tilde{s}^i_1,..., \tilde{s}^i_n\}.$

Figure \ref{fig:starwars} demonstrates the comparison between s-AMS and most activating signals from the ImageNet dataset for one unit in CLIP ResNet 50 network\footnote{Signals were obtained from \texttt{OpenAI Microscope}.}. The analysis based solely on natural signals leads to erroneous conclusions about the learned concept due to the absence of the true concept in the dataset. As the original training dataset remains undisclosed, explaining the concepts learned by the representation via identifying the most activating images from ImageNet may lead to misinterpretation, given the absence of ``Star Wars''-related images within the ImageNet dataset. In contrast, synthetic Activation-Maximization Signals can depict the learned concepts without any dependency on data.

The \textit{Synthetic Extreme-Activation} distance, or EA$_s$, is defined in a manner analogous to the EA$_n$ distance (Definition \ref{def:EAn}), with the key distinction being that Representation Activation Vectors are calculated using s-AMS instead of n-AMS. Hence, in contrast to EA$_n$, which evaluates the co-activation of representations based on each other's natural Activation-Maximization signals, EA$_s$ assesses how two representations activate in response to each other's \textit{synthetic} Activation-Maximization signals, i.e., signals generated through an artificial optimization process.

\begin{definition}[Synthetic Extreme-Activation distance]
Let $\mathcal{F} = \{f_1, ..., f_k\}$ be a layer including $k$ neural representations, and $\tilde{\mathcal{S}} = \{\tilde{S}_1,..., \tilde{S}_k\}$ be a collection of $n$ s-AMS for each of the $k$ representations in the layer. For $\forall i, j \in \{1, ..., k\}$ we introduce synthetic RAVs, by substituting n-AMS with s-AMS in Definition \ref{def:ravs}: $\tilde{r}_{ij}, \tilde{r}_{ji},$ --- pair-wise synthetic RAVs, $\tilde{r}_{i*}, \tilde{r}_{j*}$ --- layer-wise synthetic RAVs.

We define pair-wise and layer-wise synthetic Extreme-Activation distance between $f_i$ and $f_j$ as

\begin{equation}
    d^p_{EA_s}\left(f_i, f_j\right) = \frac{1}{\sqrt{2}}\sqrt{1 - \cos\left(\tilde{r}_{ij}, \tilde{r}_{ji}\right)}, \quad d^l_{EA_s}\left(f_i, f_j\right) = \frac{1}{\sqrt{2}}\sqrt{1 - \cos\left(\tilde{r}_{i*}, \tilde{r}_{j*}\right)}.
\end{equation}

\end{definition}

Notably, EA$_n$ distance is computed using standardized activations. However, due to the data-agnostic nature of the EA$_s$ distance, standardization cannot be performed without accessing the evaluation dataset; hence, we use the raw representation's activations. Although this could be considered as a limitation, since the EA$_s$ distance is not shift-invariant, we found in our practical experiments that the angles between synthetic and natural RAVs are typically maintained.

\subsection{Properties and Limitations}
\label{sec:ea_properties}
\begin{figure*}[h]
\begin{center}
\centerline{\includegraphics[width=\textwidth]{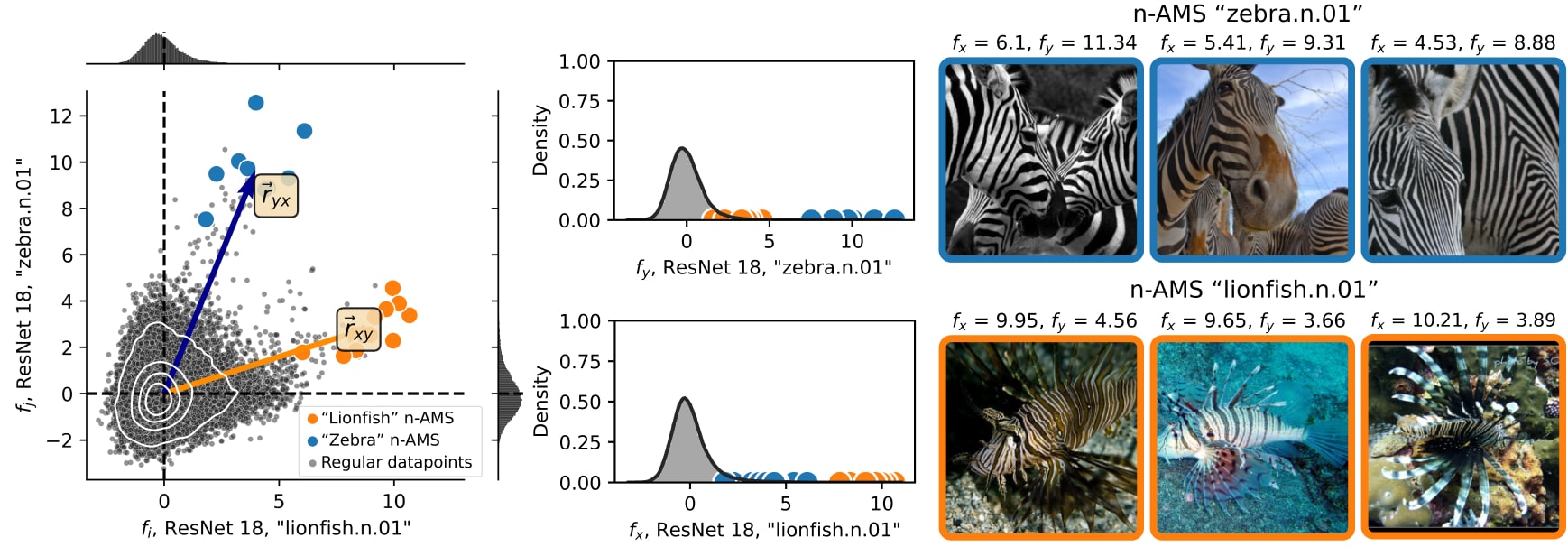}}
\vspace{-0.25cm}
\caption{\textbf{Explaining the Extreme-Activation Distance.} Figure illustrates EA$_n$ distance between two 
logit representations from ResNet18 network, corresponding to ``Zebra'' and ``Lionfish'' classes. Left figure demonstrates that RAVs experience colinearity, implying mutual co-activation on each other n-AMS ($n = 10, d = 10000$). The center part of the figure presents the activation distributions of representations across the evaluation dataset, where the highlighted points correspond to n-AMS. We can observe that both representations are strongly activated by each other's n-AMS, with the ``Lionfish'' represented in orange and the ``Zebra'' in blue. The right-hand portion of the figure presents several n-AMS examples, highlighting the shared feature between the images -- the black and white stripe pattern present in both animals -- which explains their mutual co-activation.}
\label{fig:n-ams_explainable}
\end{center}
\end{figure*}

A key characteristic of the Extreme-Activation distance, applicable to both natural and synthetic contexts, in comparison to other metrics, such as Minkowski, Spearman, and Pearson metrics, is that its computation is based on the activations of two representations given a small subset of AMS, enabling a manual examination of these data points. Such analysis of AMS provides insights into the shared concepts between the two representations. Figure \ref{fig:n-ams_explainable} illustrates the procedure of calculating EA$_n$ distance between two representations obtained from the logit layer of the ResNet18 network, specifically for the ``Zebra'' and ``Lionfish'' representations. Examination of the angle between Representation Activation Vectors (RAVs) reveals a mutual co-activation between the two representations on each other's n-AMS. By analyzing the signals themselves, which are displayed on the right section of the figure, we can observe the unique shared concepts between these two classes --- notably the specific black and white striped pattern exhibited by both animals. Thus, by relying on a limited number of \textit{anchor} data points, the EA distance facilitates the interpretation of the reasons behind the functional similarity between representations.

EA distance, as discussed above, operates on the premise that the representations could be explained by the features in input, that maximally activate them.  This viewpoint could be seen as an oversimplification, as it neglects certain attributes that moderately activate or even de-activate representations. Depending on the specific problem and the type of representations in question, such nuances could hold significant importance for understanding the internal decision-making mechanisms of the network. Nonetheless, our experimental findings underline that the EA distance, even when solely focusing on the most activating signals, offers a potent and practical framework for interpreting Deep Neural Networks. This is particularly applicable when examining representations with output confined to the positive realm, for instance, post-ReLU activation function.

One limitation common to all distance metrics between neural representations, including the proposed EA distance, is their inability to account for the \textit{multisemanticity} or \textit{multimodality} \cite{goh2021multimodal} of neural representations – the capacity of a single representation to detect various concepts. In the case of the EA distance, this behavior can be reflected in the high variance of the RAVs, which results from the fact that n-AMS originate from multiple modalities, i.e., different concepts. While this remains an open avenue for future research, our work demonstrates that simply averaging activations for RAVs computation provides an effective distance measure.

\section{Representation Analysis}

Traditionally, local XAI methods are used to analyze a model's decision-making process, particularly when assessing potential reliance on unwanted spurious concepts. However, these techniques often fall short when tasked with identifying novel, unfamiliar correlations. In this context, we introduce \textit{Representation Analysis} --- a global approach for interpreting a model's decision-making process. This approach is based on the analysis of the internal representations within the models, as well as their interrelationships. After selecting a particular layer, relationships between representations are measured using the proposed EA distance measure. This enables the visualization of the representation space, the analysis of groups of neurons that have learned similar concepts, and the identification of representations that encode anomalous abstractions.

Depending on the problem specifics, the EA distance computation may be executed in a data-aware mode using the EA$_n$ distance metric or in a data-agnostic mode using the EA$_s$ distance metric. We denote the latter scenario as \textbf{DORA} --- \textit{Data-agnOstic Representation Analysis}. The choice between a data-aware and a data-agnostic scenario primarily hinges on the availability of data for analysis. DORA reduces the dependence on data, proving particularly advantageous for interpreting models for which the training data is either unavailable or exceedingly difficult to obtain. Conversely, in practical applications, n-AMS are generally easier to comprehend compared to s-AMS.

\begin{figure*}
\begin{center}
\centerline{\includegraphics[width=\textwidth]{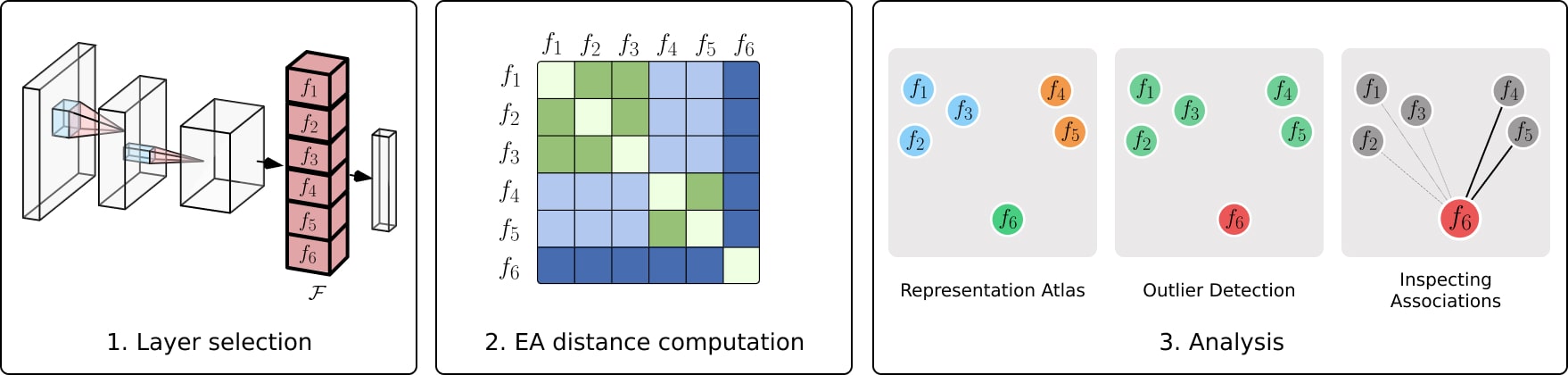}}

\caption{\textbf{Outline of the Representation Analysis:} The figure depicts the three fundamental stages of the analysis: 1. Selection of the layer of interest within the specified model. 2. Calculation of EA distances between representations. 3. Analysis of the relationships between representations, encompassing the visualization of representation space, identification of outlier representations, and, when necessary, manual scrutiny of suspicious relationships between representations.}
\label{fig:ra_main}
\end{center}
\end{figure*}

Figure \ref{fig:ra_main} provides an overview of the three primary stages of the Representation Analysis pipeline, each of which is further elaborated in the subsequent sections. Initially, a layer of interest is selected from the model. Subsequently, the Extreme-Activation distances between representations are computed. Finally, the relationships between the representations are analyzed. This includes visualizing the representation space via \textit{Representation Atlases}, automatically identifying outlier representations, and, if necessary, conducting a thorough manual examination of the causes behind the relationships between representations. As we will show in the following experiments, these outlier representations often encode undesired concepts.

\subsection*{Visualizing Representation Spaces with Representation Atlases}

Inspired by \cite{carter2019exploring}, the visual examination of the functional diversity within one layer can be done by employing the dimensionality reduction method based on the pre-computed EA distance matrix. Such a visualization, referred to as \textit{representation atlas}, allows researchers to visually examine the topological
\begin{wrapfigure}{r}{0.35\textwidth}
\centering
\includegraphics[width=0.35\textwidth]{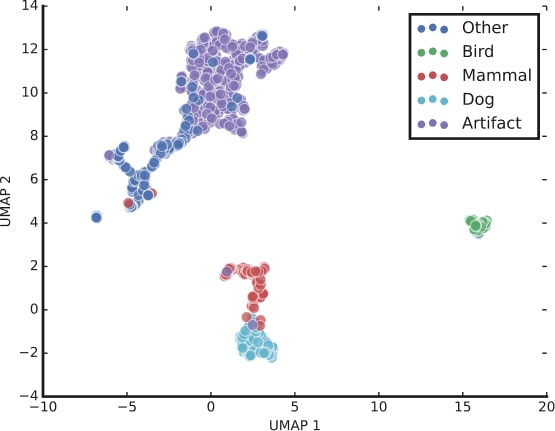}
\vspace{-0.5cm}
\caption{\textbf{Representation Atlas of the ResNet18 Logit Layer.} The figure presents a UMAP visualization of the layer-wise EA$_s$ distances between output logit representations from a ResNet18 trained on ImageNet. Each point represents a unique class representation and is color-coded according to its corresponding WordNet hypernym, i.e., the broader category.
}
\label{fig:resnet_18_representation atlas}
\end{wrapfigure}
landscape of learned representations and identify clusters of semantically similar representations. In the scope of this paper, we employed the widely used UMAP dimensionality reduction algorithm \cite{McInnes2018UMAP}, which has established itself in recent years as an effective method for visualizing relationships between data points. Figure \ref{fig:resnet_18_representation atlas} depicts the representation atlas of the output logit layers of the ResNet18 model trained on ImageNet. Each point in the figure corresponds to an individual neural representation among the 1000 representations in the output layer. The color of each point reflects the WordNet hypernym, a high-level synset, that corresponds to the learned concept of the particular representation. The UMAP visualization, based on the computed EA$_s$ distances, reveals the clusters of semantically similar representations that are preserved, which can be observed in Figure \ref{fig:resnet_18_representation atlas}.

In comparison with other dimensionality reduction methods, such as t-SNE \cite{van2008visualizing} and PCA \cite{jolliffe2016principal}, UMAP is scalable, exhibits a faster computation time \cite{McInnes2018UMAP, trozzi2021umap,becht2019dimensionality, wu2019comparison}, and has fewer parameters to tune. Qualitatively, compared to the other methods, UMAP was reported to improve visualizations and accurately represent the data structure on the projected components \cite{trozzi2021umap, becht2019dimensionality, wu2019comparison}.

\subsection*{Identifying Outlier Representations}

\begin{wrapfigure}{r}{0.57\textwidth}
\vspace{-0.45cm}
\includegraphics[width=0.57\textwidth]{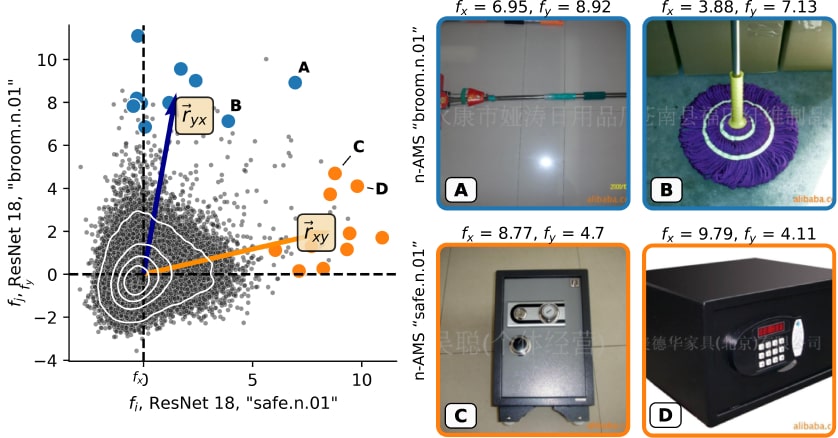}
\vspace{-0.4cm}
\caption{\textbf{Functional similarity between representations due to shared watermark dependencies.} Two representations from the ResNet18 model, specifically ``Broom'' and ``Safe'', show significant co-activation on some n-AMS, despite their visual dissimilarity, which would typically suggest orthogonal RAVs. A closer inspection of the mutual n-AMS reveals that the common feature is the presence of watermarks on the images, which is shown among 4 n-AMS, mutually activating both representations.}
\vspace{0.25cm}
\label{fig:inspecting_associations}
\end{wrapfigure}

Despite their proven effectiveness across various applications, Deep Neural Networks remain susceptible to learning unintended artifacts and undesired concepts from data. One potential application of our proposed distance measure is the identification of anomalous representations, which deviate from the majority of representations within a layer. This can be achieved by applying Outlier Detection methods \cite{ruff2021unifying} based on the computed distance matrix. We hypothesize that such anomalies include representations that have learned spurious concepts from the data. In subsequent experiments, we demonstrate the efficacy of the EA distance measure in detecting such outlier representations under controlled conditions. During our practical experiments, we found that while some outlier representations learn unique, task-relevant concepts, these divergent representations often encode undesirable concepts, demonstrating \textit{shortcut learning} or \textit{Clever Hans} behavior.

\subsection*{Investigating Individual Relationships}

In order to determine the cause behind a specific EA distance, individual relationships between representations can be examined manually. This can be accomplished through visualizing RAVs and AMS, as discussed in Section \ref{sec:ea_properties}. For instance, Figure \ref{fig:inspecting_associations} illustrates the spurious correlation between the ``Broom'' and ``Safe'' representations from the output logit layer of ResNet18, which accounts for the low EA distance between these representations. Upon examining the shared n-AMS, it becomes clear that the n-AMS, which maximally activate both representations, contain a single common feature — watermarks. Such visualizations can aid in investigating the underlying functional similarity between representations and offer insight into the reasons behind such similarities.

\section{Evaluation}
\label{sec:eval}

The practical utility of a distance metric between representations fundamentally depends on its capacity to gauge the similarity between concepts that various model representations have learned. The manner in which this similarity is assessed can take on multiple forms, but from a standpoint of explainability, it is important that the distance metric is aligned \cite{muttenthaler2022human,gabriel2020artificial} with human perception - the computed distances should resonate with our human senses of similarity and difference. Assuming that we know what abstractions two representations are detecting, an effective distance metric should label representations that detect concepts perceived as similar by humans as alike, and those identifying concepts perceived as distinct by humans as dissimilar. Such attributes of the distance metric facilitate the clustering of representations that are conceptually similar and the identification of outlier representations that encode concepts anomalous to the task at hand. This has advantages as it aids in revealing shortcuts and Clever Hans effects, which often appear as out-of-distribution anomalies and are unnatural to the assigned task - for example, textual watermarks in the context of object classification \cite{lapuschkin2019unmasking} or specific tokens in medical image classification \cite{geirhos2020shortcut}.

To quantitatively evaluate the alignment, we compared human-defined semantic distances between concepts, which we refer to as \textit{semantic baselines}, with distance matrices computed between representations trained to learn these concepts. For our study, we utilized two prevalent computer vision datasets, namely ILSVRC-2012 \cite{russakovsky2015imagenet} and CIFAR-100 \cite{krizhevsky2009learning}. We established a measure of distance between concepts by utilizing semantic distances between the class labels. This was accomplished by mapping the classification labels to entities within the WordNet taxonomy database \cite{miller1995wordnet}, a lexical database that organizes English words into a taxonomy of synonym sets, or synsets. In this taxonomy, each synset represents a group of words that are synonyms or have the same meaning. WordNet organizes these synsets into a hierarchy, with more specific concepts being nested under more general ones.

For the ImageNet dataset, class labels were automatically mapped to the corresponding WordNet synsets due to their inherent linkage, whereas for the CIFAR-100 dataset, the labels were manually matched to their respective synsets. It is important to note that semantic distance does not directly equate to the visual similarity between concepts. However, positive correlations between semantic and visual similarities have been reported, thereby demonstrating a significant positive relationship between semantic and visual distances \cite{deselaers2011visual, brust2019not}.

Given the WordNet taxonomy in a form of a graph $\mathcal{G} = (V, E)$ with root $r \in V$, the baseline semantic distances between entities from the WordNet database were computed using the following three distance measures:

\begin{itemize}
\item \textbf{Shortest-Path distance}

Given two vertices $c_i, c_j \in V$ the distance between vertices is determined by the length of the shortest path that connects the two entities in the taxonomy
\begin{equation*}
d_{SP}(c_i, c_j) = l(c_i, c_j).
\end{equation*}
where $l(c_i, c_j)$ is the function that returns the minimal number of edges that need to be traversed to get from $c_i$ to $c_j.$

\item \textbf{Leacock-Chodorow distance \cite{leacock1998combining}}

Given two vertices $c_i, c_j \in V$ the distance between vertices is determined by a logarithm of the shortest-path distance with additional scaling by the taxonomy depth:
\begin{equation*}
d_{LC}(c_i, c_j) = \log{\frac{l(c_i, c_j) + 1}{2T}} - \log{\frac{1}{2T}},
\end{equation*}
where $T = \max_{c \in V} l(r, c)$ is the taxonomy depth.

\item \textbf{Wu-Palmer distance \cite{wu1994verb}}

Given two vertices $c_i, c_j \in V$ the Wu-Palmer distance is defined as:
\begin{equation*}
d_{SP}(c_i, c_j) = 1 - 2\frac{l(r, lcs(c_i, c_j))}{l(r, c_i) + l(r, c_j)},
\end{equation*}
where $lcs(c_i, c_j)$ is the Least Common Subsumer \cite{pedersen2004wordnet} of two concepts $c_i$ and $c_j.$
\end{itemize}

Furthermore, we have utilized the textual labels from both ImageNet and CIFAR100 datasets and calculated the Word2Vec \cite{mikolov2013efficient} similarity between class labels.
\begin{itemize}
    \item \textbf{Word2Vec distance}

    Given textual labels $t_i, t_j$ of two concepts $c_i, c_j$, we define Word2Vec distance as
    \begin{equation*}
        d_{W2V} = \frac{1}{\sqrt{2}}\sqrt{1 - \cos_{W2V}(t_i, t_j)},
    \end{equation*}
    where $\cos_{W2V}(A, B)$ is the cosine of the angle between Word2Vec embeddings of the words $A, B.$
\end{itemize}

\begin{figure}[h!]
\centering
\includegraphics[width=\linewidth]{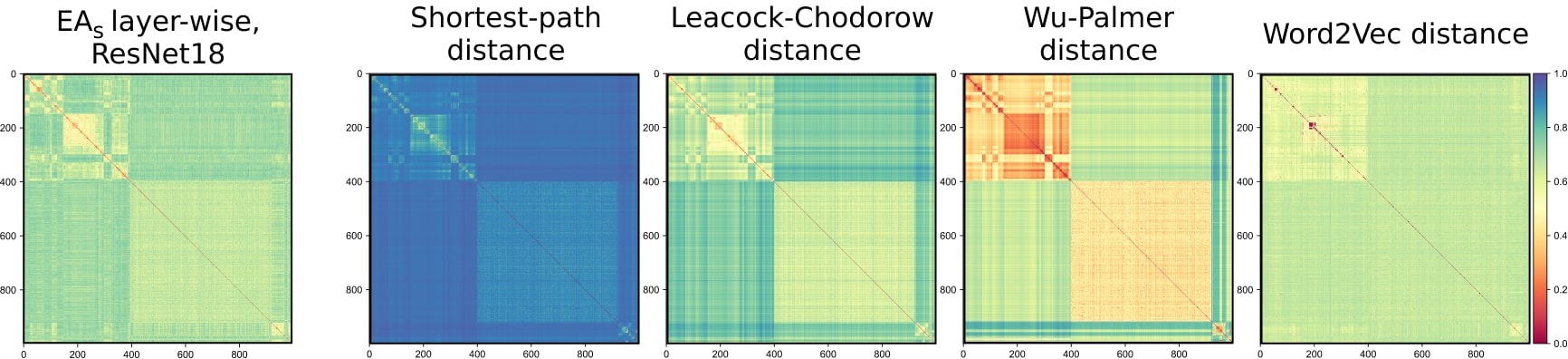}
\vspace{-0.25cm}
  \caption{\textbf{Comparison between EA$_s$ distance matrix and Seamntic Baselines}: From left to right: the EA$_s$ distance metric computed for the output logits of the ImageNet pre-trained ResNet18 model, Shortest-Path, Leacock-Chodorow, Wu-Palmer distances from WordNet taxonomy, and Word2Vec distance.}
  \label{fig:semantic_baseline}

\end{figure}

Figure \ref{fig:semantic_baseline} illustrates the comparison between the functional EA$_s$ distance matrix, derived from 1000 representations from an ImageNet-trained ResNet18 network, and semantic baselines — these are the human-defined distances between ImageNet concepts. To evaluate the alignment between the proposed distance metric and human-defined baselines, we employed the Mantel Test \cite{mentel1967detection}, which is often applied in ecology and evolutionary biology to measure the correlation between two distance matrices. The test calculates the correlation coefficient $\rho$, which indicates the strength of the relationship between the two matrices, and the $p$-value of the test, which describes the statistical significance of the correlation.

It is essential to note that while we evaluate the alignment based on human-defined semantic benchmarks, optimizing such metrics should not be the ultimate objective when proposing new distance metrics between representations. This is because DNNs can naturally employ different decision-making strategies than humans, and these differences may not always be attributed to spurious correlations. For instance, taxonomy-based approaches might be sub-optimal compared with attributing freedom to the models to train for the desired tasks \cite{binder2012taxonomies}. Conversely, in Computer Vision, network representations are expected to be aligned to some extent due to the correlations between the visual and semantic similarity of classes.

\subsection{Hyperparameter selection}
\label{sec:hyperparameter_search}

This section examines the selection of parameters in terms of their ability to attain optimal alignment with the semantic baselines. In our empirical analysis, we utilized a pre-trained ResNet18 model on ImageNet, along with the ILSVRC-2012 validation set consisting of 50,000 images and 1,000 classes, employed for the data-aware metrics. Herein, we calculated the distance metrics between output logit representations, that is, the pre-softmax representations. For data-aware metrics, outputs of representations underwent normalization as discussed in Section \ref{sec:distance_metrics}. Conversely, for the data-agnostic distance metric, specifically EA$_s$, no normalization procedure was undertaken.

\subsubsection*{Minkowski distance}

\begin{wrapfigure}{r}{0.4\textwidth}
\vspace{-0.3cm}
\includegraphics[width=0.4\textwidth]{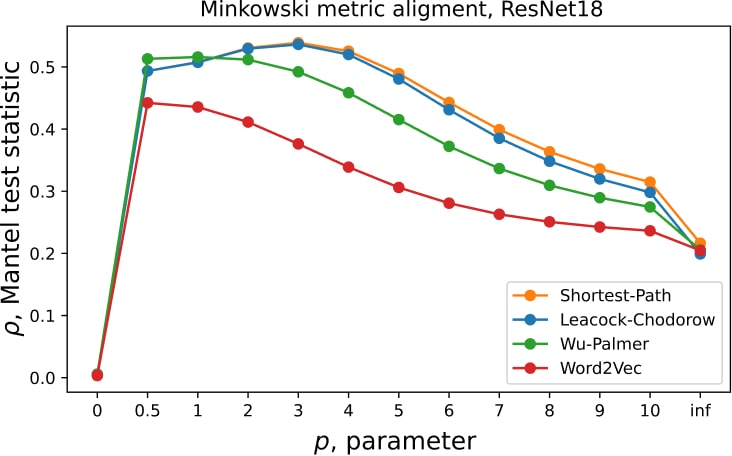}
\vspace{-0.5cm}
\caption{\textbf{Impact of Parameter Selection in Minkowski Distance on Alignment with Semantic Baselines.} To assess the alignment with respect to the four semantic baselines, we calculated the Minkowski distance on the output logits of the ResNet18 network while varying the parameter $p$. The Mantel correlation statistic is reported for each semantic baseline at each parameter value.
}
\vspace{-0.5cm}
\label{fig:minkowski_hyperparamenter}
\end{wrapfigure}

To investigate how different values of the parameter $p$ affect the coherence to semantic baselines, we varied the parameter and evaluated the alignment with four semantic baselines for each case. Figure \ref{fig:minkowski_hyperparamenter} shows the effect of parameter selection on the Mantel test statistic. We observed that the optimal average value of the statistic across the four baselines was achieved for $p = 2$. However, for future experiments, we selected the second-best parameter choice with $p = 1$ due to the natural connection between Euclidean distance and Pearson correlation. We also observed that higher values of $p$ generally result in lower alignment, possibly due to sensitivity to the large amplitudes of individual datapoints.

\subsubsection*{EA$_n$ distance}

\begin{figure}
\centering
\includegraphics[width=\linewidth]{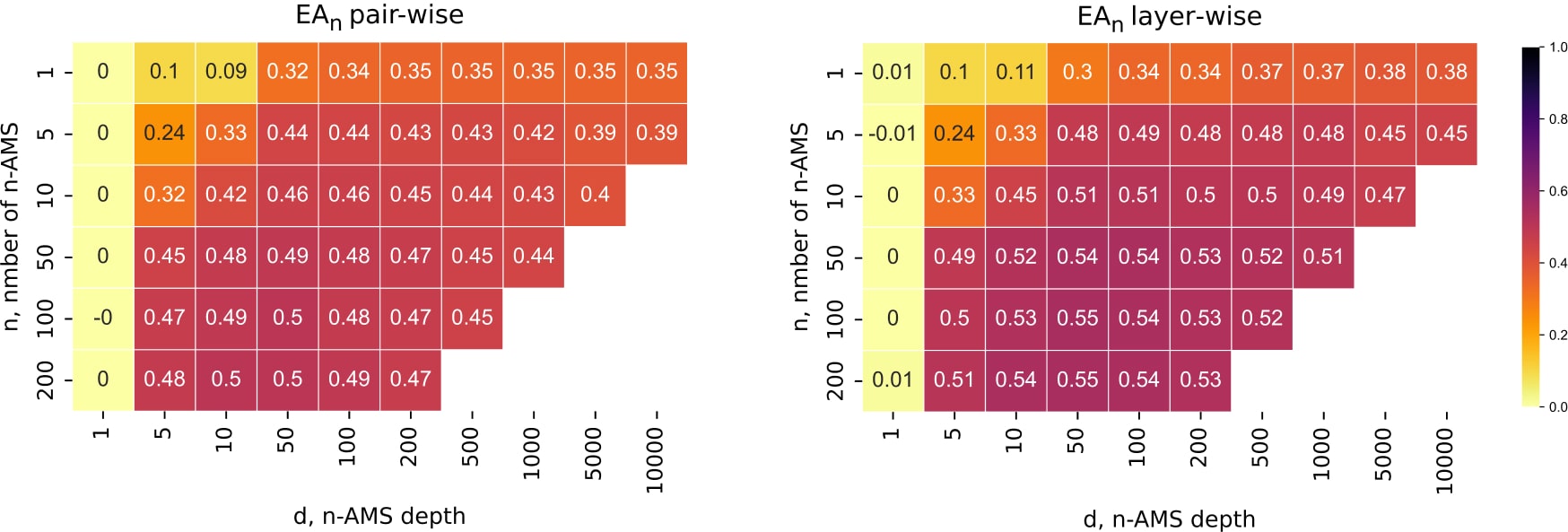}
\vspace{-0.25cm}
  \caption{\textbf{Impact of parameter selection in EA$_n$ distance on alignment with semantic baselines.} To assess alignment with four semantic baselines, we calculated data-aware EA distance on the output logits of the ResNet18 network while varying the parameters $n$ and $d$ for both pair-wise and layer-wise options. The average Mantel correlation statistic across four semantic baselines is reported at each cell.}
  \label{fig:nams_hyperparameter_search}
\end{figure}

EA$_n$ is influenced by two key parameters: $n,$ which denotes the number of n-AMS signals gathered, and $d,$ which represents the size of the subset collected from each signal. To investigate the impact of parameter selection, we varied these parameters for both pair-wise and layer-wise modes. Figure \ref{fig:nams_hyperparameter_search} shows the average Mantel correlation statistic across four semantic baselines for each hyperparameter choice. Our observations reveal that, in general, increasing the number of collected n-AMS, irrespective of the parameter $d$, has a positive impact on the alignment. However, the optimal depth $d$ is achieved when n-AMS are taken from subsets of $d = 50$ datapoints.

\subsubsection*{EA$_s$ distance}

\begin{figure}[h!]
\centering
\includegraphics[width=\linewidth]{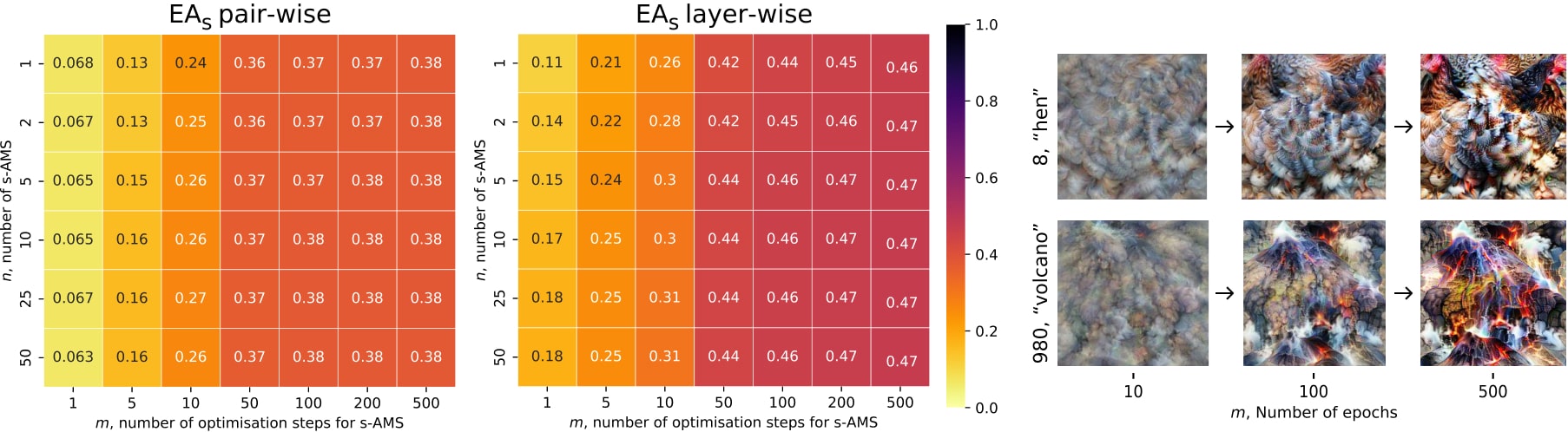}
\vspace{-0.25cm}
  \caption{\textbf{Impact of the parameter selection in EA$_s$ distance on alignment with semantic baselines.} To evaluate the alignment between the EA$_s$ distance and four semantic baselines, we computed the data-agnostic EA$_s$ distance using the ResNet18 network's output logits while varying the hyperparameters $n$ and $m$ for both pair-wise and layer-wise options. For each cell, we reported the average Mantel correlation statistic across the four semantic baselines. The effect of the hyperparameter $m$, which corresponds to the number of optimization steps taken for s-AMS generation, on two neural representations from the ResNet18 output logit layer is shown on the right.}
  \label{fig:sams_hyperparameter_search}
\end{figure}

In the data-agnostic version of the Extreme-Activation distance, the choice of hyperparameters depends on the s-AMS generation method used. In our study, we employed the Feature Visualisation method to generate s-AMS, and we identified two critical hyperparameters: $n$, which is the number of generated s-AMS per representation, and $m$, which is the number of optimization epochs per signal. Figure \ref{fig:sams_hyperparameter_search} depicts the impact of the EA$_s$ distance measure's hyperparameter selection on semantic baseline alignment. We observed that while increasing the number of generated s-AMS generally has a positive effect, this effect is negligible compared to the positive impact of increasing the number of optimization epochs per representation. This is likely due to the generation algorithms' convergence to better local optima, resulting in improved visual preciseness of the images, as illustrated on the right side in Figure \ref{fig:sams_hyperparameter_search}.

\subsection{EA$_s$ Preserves the Angles Between Natural Representation Activation Vectors}

\begin{figure*}[t]
\begin{center}
\centerline{\includegraphics[width=\textwidth]{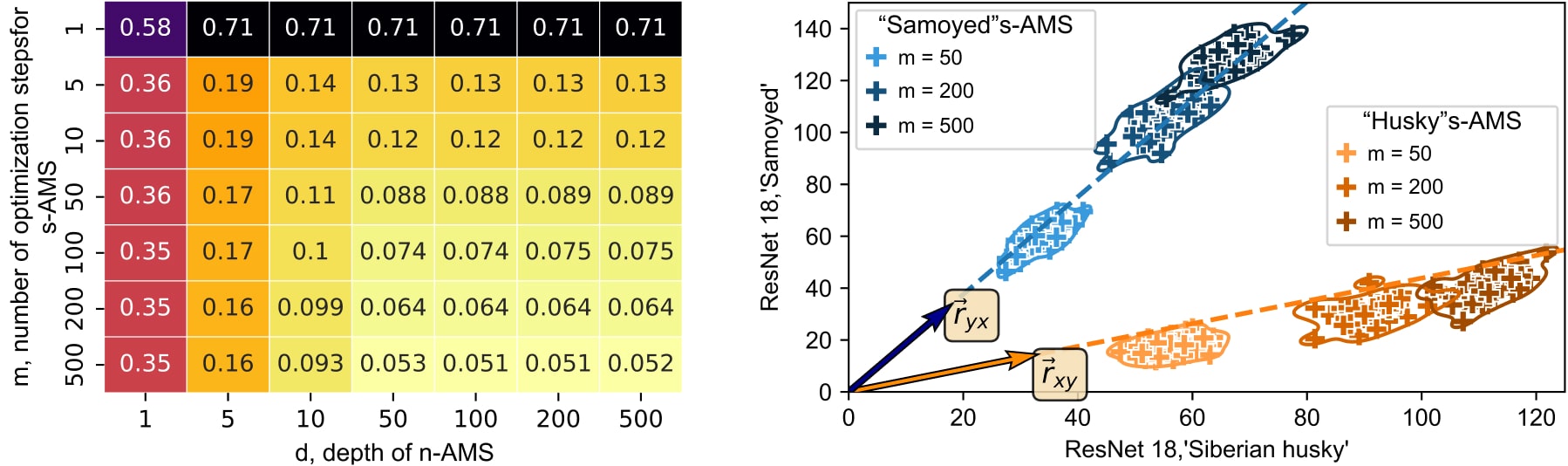}}
\caption{\textbf{Similarity and angle preservation between EA$_n$ and EA$_s$ distance measures.} The left part of the figure shows the RMSE (lower is better) between pair-wise EA$_n$ and EA$_s$ distances on the output layer of the ResNet18 network, with a fixed parameter $n=50$ for both metrics, while varying parameters $d,$ corresponding to the subset size in n-AMS sampling, and $m,$ number of epochs for s-AMS generation. On the right part of the figure, the distributions of pair-wise activations of s-AMS signals are visualized with different parameters $m$ for two neural representations, namely ``Samoyed'' and ``Siberian husky'', overlayed with the direction of natural RAVs computed with $n=50$ and $m=1000$. The length of natural RAVs was extended to enhance visibility.}
\label{fig:angle_conservation}
\end{center}
\end{figure*}

Although both n-AMS and s-AMS activate specific neural representations maximally, the adversarial nature of synthetic signals needs to be considered. In our experiments, we observed that while the generated s-AMS are far from the original \textit{natural} image domain, the angles between natural and synthetic RAVs are consistent, providing additional evidence to the utility of the EA$_s$ distance metric.

To evaluate the angle conservation quantitatively, we employed a ResNet18 pre-trained on the ImageNet dataset and computed EA$_n$ and EA$_s$ distances between the output logit representations, i.e., all 1000 ImageNet classes.  It's important to mention that the EA$_n$ distance is calculated over normalized representations, while EA$_s$ is based on the unnormalized output of representations. In this experiment, we kept the number of signals constant at $n=50$ for both distance metrics, yet varied the parameter $d$ for n-AMS generation and the parameter $m$ for s-AMS generation. Given $\mathcal{F}=\{f_1,\ldots,f_k\}$, which corresponds to the ResNet18 output layer with $k=1000$ neural representations, we evaluated the Root Mean Square Error (RMSE) between pairwise EA$_n$ and EA$_s$ distances:

\begin{equation}
RMSE = \sqrt{\frac{\sum_{i=1}^k \sum_{j=i+1}^k \left(d^p_{EA_n}(f_i,f_j) - d^p_{EA_s}(f_i,f_j)\right)^2}{k(k-1)/2}},
\end{equation}
where $k(k-1)/2$ corresponds to the number of all unique pairs of two different functions from a set of $k$ functions.

Figure \ref{fig:angle_conservation} illustrates the similarity between the computed EA$_n$ and EA$_s$ distances between the representations of the 1000 ImageNet classes. In the left part of the figure, which shows RMSE between the two distance measures for different parameters, we observe that for each parameter $m$ for EA$_s$ distance, the lowest error is achieved with an EA$_n$ distance with high values of $d$.This suggests that the EA$_s$ distance accurately captures the angle between RAVs corresponding to the images with the highest activation percentile. Additionally, we observed that increasing the parameter $m$ is beneficial to lowering the RMSE between natural and synthetic measures. Furthermore, the right part of the figure shows the direction of natural RAVs and activations of s-AMS for ``Samoyed'' and ``Siberian husky'' representations from ResNet18. From this figure, we can observe that the angle between natural RAVs and synthetic RAVs is conserved.

\subsection{Evaluating the alignment with human judgment}
\label{sec:evaluation_alignment}

In this experiment, we quantitatively assess the alignment of the discussed distance metrics with the human-defined distance measures across different datasets and architectures. To this end, we employed eight different architectures for two datasets, ImageNet and CIFAR100. For ImageNet, we employed ResNet18 \cite{he2016deep}, AlexNet \cite{krizhevsky2017imagenet}, ViT \cite{dosovitskiy2020image}, BEiT \cite{bao2021beit}, Inception V3 \cite{szegedy2016rethinking}, DenseNet 161 \cite{huang2017densely}, MobileNet V2 \cite{sandler2018mobilenetv2}, ShuffleNet V2 \cite{ma2018shufflenet}, while for CIFAR-100, we used ResNet 18, ResNet 9, MobileNet V2, ShuffleNet V1, and V2, as well as NASNet \cite{qin2019nasnet}, SqueeeNet  \cite{iandola2016squeezenet} and VGG 11 \cite{simonyan2014very}.

We computed functional distances with optimal hyperparameters found in Section \ref{sec:hyperparameter_search}, including Minkowski $p = 1$, Pearson, Spearman, EA$_n$ with $n = 50, d = 200$, and EA$_s$ with $n = 3, m = 500$, on the output logit layer for each model. We then compared each distance matrix with four semantic baselines: Shortest-Path, Leacock-Chodorow, Wu-Palmer distances from WordNet taxonomy, and Word2Vec distance. This comparison yielded four Mantel test statistics per distance metric. The results of the evaluation are presented in Table \ref{tab:alignment_evaluation_imagenet} for ImageNet-trained models and in Table \ref{tab:alignment_evaluation_cifar} for CIFAR100 models, where we averaged the four Mantel correlation test statistics for each model and distance metric. Our analysis indicates that the layer-wise EA$_n$ metric's distance is generally more favorable due to its stronger linear relationship with all four baseline metrics. Furthermore, we observed that the data-agnostic EA$_s$ metric is on par with data-aware metrics in terms of alignment with the semantic baselines.

\begin{table}[]
\centering
\caption{\textbf{Alignment of Distance Metrics in ImageNet Trained Models}: Each cell represents the average Mantel test statistic (higher is better) across four semantic baselines: Shortest-Path, Leacock-Chodorow, Wu-Palmer distances, and Word2Vec distance. All results demonstrate statistical significance with $p < 0.001$.}
\label{tab:alignment_evaluation_imagenet}
\vspace{2.5mm}
\begin{tabular}{@{}rrccccccccccccc@{}}
\toprule
\textit{} &
  \multicolumn{1}{c}{\textit{}} &
  \textit{Minkowski} &
  \textit{} &
  \textit{Pearson} &
  \textit{} &
  \textit{Spearman} &
  \textit{} &
  \multicolumn{3}{c}{\textit{EA$_n$}} &
  \textit{} &
  \multicolumn{3}{c}{\textit{EA$_s$}} \\
\textit{} &
  \multicolumn{1}{c}{\textit{}} &
  \textit{$p = 1$} &
  \textit{} &
  \textit{} &
  \textit{} &
  \textit{} &
  \textit{} &
  \textit{p-w} &
  \textit{} &
  \textit{l-w} &
  \textit{} &
  \textit{p-w} &
  \textit{} &
  \textit{l-w} \\ \midrule
\textit{ResNet18}     &                               & 0.49 &  & 0.50 &  & 0.48 &  & 0.49 &  & 0.55 &  & 0.38 &  & 0.47 \\
\textit{BeIT}         &                               & 0.32 &  & 0.36 &  & 0.29 &  & 0.44 &  & 0.50 &  & 0.39 &  & 0.47 \\
\textit{MobilenetV2}  &                               & 0.46 &  & 0.46 &  & 0.45 &  & 0.47 &  & 0.52 &  & 0.40 &  & 0.50 \\
\textit{DenseNet161}  &                               & 0.46 &  & 0.47 &  & 0.44 &  & 0.49 &  & 0.54 &  & 0.32 &  & 0.39 \\
\textit{ShuffleNetV2} &                               & 0.21 &  & 0.21 &  & 0.19 &  & 0.29 &  & 0.30 &  & 0.19 &  & 0.16 \\
\textit{InceptionV3}  & \textbf{}                     & 0.31 &  & 0.34 &  & 0.32 &  & 0.38 &  & 0.49 &  & 0.22 &  & 0.27 \\
\textit{AlexNet}      & \multicolumn{1}{l}{\textbf{}} & 0.52 &  & 0.53 &  & 0.52 &  & 0.52 &  & 0.55 &  & 0.42 &  & 0.45 \\
\textit{ViT}          & \multicolumn{1}{l}{}          & 0.53 &  & 0.54 &  & 0.52 &  & 0.54 &  & 0.58 &  & 0.48 &  & 0.53 \\
\textbf{Mean} &
  \multicolumn{1}{l}{} &
  \textbf{0.41} &
  \textbf{} &
  \textbf{0.43} &
  \textbf{} &
  \textbf{0.40} &
  \textbf{} &
  \textbf{0.45} &
  \textbf{} &
  \textbf{0.50} &
  \textbf{} &
  \textbf{0.35} &
  \textbf{} &
  \textbf{0.40} \\ \bottomrule
\end{tabular}
\end{table}

\begin{table}[]
\centering
\caption{\textbf{Alignment of Distance Metrics in CIFAR100 Trained Models}: Each cell represents the average Mantel test statistic (higher is better) across four semantic baselines: Shortest-Path, Leacock-Chodorow, Wu-Palmer distances, and Word2Vec distance. All results demonstrate statistical significance with $p < 0.001$.}
\vspace{2.5mm}
\label{tab:alignment_evaluation_cifar}
\begin{tabular}{@{}rrccccccccccccc@{}}
\toprule
\textit{} &
  \multicolumn{1}{c}{\textit{}} &
  \textit{Minkowski} &
  \textit{} &
  \textit{Pearson} &
  \textit{} &
  \textit{Spearman} &
  \textit{} &
  \multicolumn{3}{c}{\textit{EA$_n$}} &
  \textit{} &
  \multicolumn{3}{c}{\textit{EA$_s$}} \\
\textit{} &
  \multicolumn{1}{c}{\textit{}} &
  \textit{$p = 1$} &
  \textit{} &
  \textit{} &
  \textit{} &
  \textit{} &
  \textit{} &
  \textit{p-w} &
  \textit{} &
  \textit{l-w} &
  \textit{} &
  \textit{p-w} &
  \textit{} &
  \textit{l-w} \\ \midrule
\textit{ResNet9}      &                               & 0.32 &  & 0.37 &  & 0.33 &  & 0.41 &  & 0.52 &  & 0.27 &  & 0.30 \\
\textit{ShuffleNetV2} &                               & 0.49 &  & 0.52 &  & 0.49 &  & 0.53 &  & 0.59 &  & 0.43 &  & 0.47 \\
\textit{MobileNetV2}  &                               & 0.50 &  & 0.51 &  & 0.49 &  & 0.52 &  & 0.59 &  & 0.40 &  & 0.44 \\
\textit{ResNet18}     &                               & 0.43 &  & 0.47 &  & 0.45 &  & 0.48 &  & 0.57 &  & 0.30 &  & 0.37 \\
\textit{ShuffleNet}   &                               & 0.48 &  & 0.51 &  & 0.49 &  & 0.52 &  & 0.58 &  & 0.42 &  & 0.46 \\
\textit{VGG11}        & \textbf{}                     & 0.30 &  & 0.31 &  & 0.31 &  & 0.36 &  & 0.43 &  & 0.23 &  & 0.23 \\
\textit{NasNet}       & \multicolumn{1}{l}{\textbf{}} & 0.48 &  & 0.51 &  & 0.48 &  & 0.52 &  & 0.59 &  & 0.36 &  & 0.41 \\
\textit{SqueezeNet}   & \multicolumn{1}{l}{}          & 0.50 &  & 0.52 &  & 0.51 &  & 0.53 &  & 0.59 &  & 0.45 &  & 0.51 \\
\textbf{Mean} &
  \multicolumn{1}{l}{} &
  \textbf{0.44} &
  \textbf{} &
  \textbf{0.46} &
  \textbf{} &
  \textbf{0.44} &
  \textbf{} &
  \textbf{0.48} &
  \textbf{} &
  \textbf{0.56} &
  \textbf{} &
  \textbf{0.36} &
  \textbf{} &
  \textbf{0.40} \\ \bottomrule
\end{tabular}
\end{table}

\subsection{Evaluating Anomaly-Identification capabilities}

The alignment of distance metrics between neural representations and human judgment of concepts presents an intriguing potential application. Specifically, we can identify representations that are semantically anomalous compared to the majority of learned representations, based on the functional distance. While these representations may simply learn unique individual concepts, we demonstrate in further experiments that in real-life scenarios they might correspond to the undesired concepts from spurious correlations in the training data that diverge from the typical (intended) decision-making strategy.

To assess the usefulness of the alignment between distance metrics and human-defined semantic baseline, we conducted the experiment, where we measured the ability of the distance metrics to detect anomalous representations. For this purpose, we trained a ResNet18 network on a combination of two conceptually different datasets. The combined dataset comprised the Tiny Imagenet \cite{le2015tiny}, containing 200 ImageNet classes, and the MNIST handwritten-numbers dataset \cite{deng2012mnist}, containing 10 handwritten numbers, resulting in a total of 210 classes. MNIST images were upsampled to the size of 3 $\times$ 64 $\times$ 64 pixels to match the size of images in Tiny ImageNet. After training on the combined dataset in the image classification task, we computed functional distances between the output logits and evaluated the ability of different Outlier Detection (OD) methods to detect MNIST logits, given the computed distance matrices only. For this, we utilized five different Outlier Detection methods: the Angle-based Outlier Detector (ABOD) \cite{kriegel2008angle}, Feature Bagging (FB) \cite{lazarevic2005feature}, Isolation Forest (IF) \cite{liu2008isolation}, Local Outlier Factor (LOF) \cite{breunig2000lof} and One-class SVM (OCSVM) \cite{scholkopf2001estimating}. We evaluated the performance of the Outlier Detection methods using the AUC ROC metric for the binary classification between Tiny ImageNet and MNIST representations. To ensure stability in light of the stochastic nature of some outlier detection methods, the results of the outlier detection were repeated 100 times with different random states.

\begin{figure}[h!]
\centering
\includegraphics[width=\linewidth]{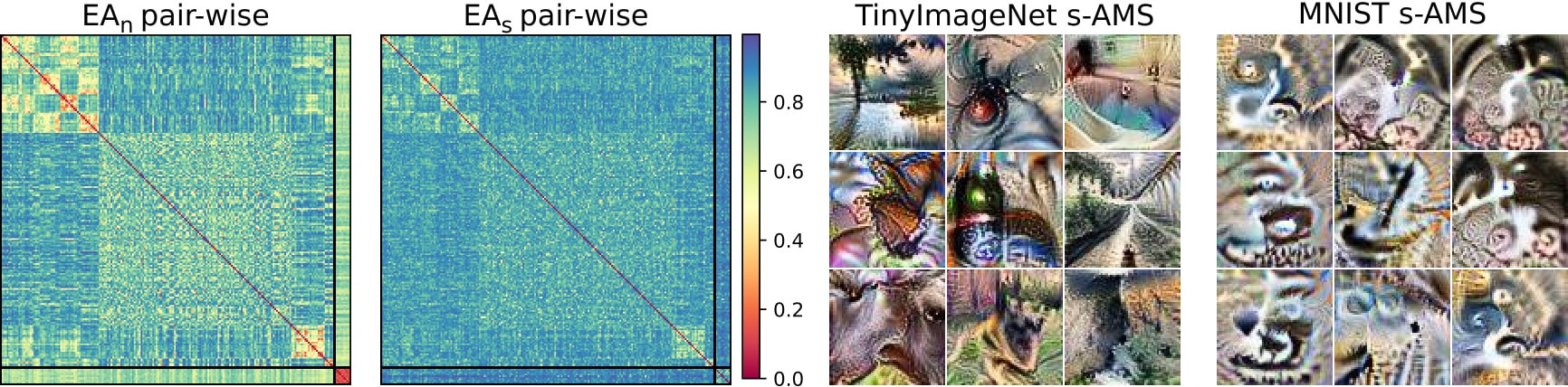}
\vspace{-0.25cm}
  \caption{\textbf{Anomaly-Detection Evaluation Experiment.} From left to right, pair-wise EA$_n$ distance matrix between the output logits of the network trained on the combined dataset, the EA$_s$ distance matrix, s-AMS for Tiny ImageNet logits, and s-AMS for MNIST logits. MNIST representations are highlighted on both distance matrices in the bottom right corner, revealing a block structure in both distance metrics that suggests a high degree of functional differences between Tiny ImageNet representations and semantically distinct MNIST representations. On the left, we can visually observe differences between the s-AMS of Tiny ImageNet and MNIST representations.}
  \label{fig:anomaly_detection_vizualization}
\end{figure}

We utilized the same hyperparameter configuration for distance computation as described in Section \ref{sec:evaluation_alignment}. The effectiveness of EA distances, both natural and synthetic, in distinguishing between representations of Tiny ImageNet and MNIST is demonstrated in Figure \ref{fig:anomaly_detection_vizualization}, as evidenced by the block structure of the distance matrices. This behavior can be attributed to the visual dissimilarities between the classes, where Tiny ImageNet classes exhibit natural and diverse features that are typical for natural images, while MNIST images consist of white digits on a black background. In the case of synthetic EA distance, the ability to detect MNIST representations is based on the visual differences in the s-AMS, which are depicted in the right-hand portion of Figure \ref{fig:anomaly_detection_vizualization}. The s-AMS-based EA distance measure depends on the network's ability to perceive self-generated s-AMS, and we can observe distinct dissimilarities between the patterns of s-AMS for Tiny ImageNet classes, which contain high-level natural concepts, and the more data-specific patterns for MNIST classes, which illustrate the network's perception of white-on-black handwritten digits and letters.

The results of the described experiment are presented in Table \ref{tab:evaluating_od_capabilities}, which indicate that, in general, all distance metrics are capable of detecting MNIST representations. However, the EA distance metrics are generally more effective in detecting semantically artifactual representations, whereas the pairwise EA$_n$ metric is the most effective.

\begin{table}[]
\centering
\caption{\textbf{Anomaly Identification Performance of Distance Metrics.} The table displays the average AUC ROC (higher is better) binary classification performance of the Outlier Detection methods across 100 re-trials, in the task of detecting MNIST representations among the combined Tiny ImageNet and MNIST representations, specifically in the output layer of the trained network.}
\vspace{2.5mm}
\label{tab:evaluating_od_capabilities}
\begin{tabular}{rrccccccccccccc}
\hline
\multicolumn{1}{c}{\textit{}} &
  \multicolumn{1}{c}{\textit{}} &
  \textit{Minkowski} &
  \textit{} &
  \textit{Pearson} &
  \textit{} &
  \textit{Spearman} &
  \textit{} &
  \multicolumn{3}{c}{\textit{EA$_n$}} &
  \textit{} &
  \multicolumn{3}{c}{\textit{EA$_s$}} \\
\multicolumn{1}{c}{\textit{}} &
  \multicolumn{1}{c}{\textit{}} &
  \textit{$p = 1$} &
  \textit{} &
  \textit{} &
  \textit{} &
  \textit{} &
  \textit{} &
  \textit{p-w} &
  \textit{} &
  \textit{l-w} &
  \textit{} &
  \textit{p-w} &
  \textit{} &
  \textit{l-w} \\ \hline
\textit{ABOD}  &  & 0.56 &  & 0.63 &  & 0.58 &  & 0.91 &  & 1.00 &  & 0.82 &  & 0.71 \\
\textit{FB}    &  & 0.97 &  & 0.99 &  & 0.81 &  & 1.00 &  & 1.00 &  & 0.89 &  & 0.87 \\
\textit{IF}    &  & 0.83 &  & 0.87 &  & 0.64 &  & 0.94 &  & 0.70 &  & 0.76 &  & 0.61 \\
\textit{LOF}   &  & 0.65 &  & 0.53 &  & 0.55 &  & 0.67 &  & 0.96 &  & 1.00 &  & 0.87 \\
\textit{OCSVM} &  & 1.00 &  & 1.00 &  & 0.95 &  & 1.00 &  & 0.67 &  & 1.00 &  & 0.72 \\
\textbf{Mean} &
   &
  \textbf{0.80} &
  \textbf{} &
  \textbf{0.80} &
  \textbf{} &
  \textbf{0.71} &
  \textbf{} &
  \textbf{0.90} &
  \textbf{} &
  \textbf{0.87} &
  \textbf{} &
  \textbf{0.89} &
  \textbf{} &
  \textbf{0.76} \\ \hline
\end{tabular}
\end{table}

\section{Experiments: Finding Outlier Representations with DORA}
\label{sec:synthetic}

In this section, we illustrate the broad applicability of the DORA framework and demonstrate that outlier representations, often found in intermediate layers, can frequently encode malicious and undesirable concepts.

 \subsection{ImageNet pre-trained networks}

Pre-trained networks on ImageNet have become an essential component in the field of Computer Vision. Their capability to recognize a diverse set of objects and scenes makes them particularly useful as a starting point for a wide range of computer vision tasks. They are frequently utilized for fine-tuning to specific tasks or as a feature extractor, where the images are encoded by the networks for further computations \cite{zhuang2020comprehensive, weiss2016survey}.

In the following, we explore the feature extractor representations of three widely-used pre-trained models: ResNet18 \cite{he2016deep}, MobileNetV2 \cite{sandler2018mobilenetv2}, and DenseNet121 \cite{huang2017densely}. Using LOF outlier detection, we found latent layers with representations that appear to be watermark detectors, e.g., detecting Chinese and Latin text patterns. As ImageNet does not have a specific category for watermarks, these representations could be seen as Clever-Hans artifacts and deviate from desired decision-making \cite{lapuschkin2019unmasking, anders2022finding}. To verify these representations can detect watermarks, we created two binary classification datasets, for Chinese and Latin watermarks, containing normal images and identical images, with inserted random watermarks, evaluating the sensitivity of individual representations using the AUC ROC classification measure. To ensure the detection of characters and not specific words/phrases (unlike CLIP models \cite{goh2021multimodal}), the probing datasets were generated with random characters (for more details we refer to the Appendix). Our results show that not only the reported outliers but also neighboring representations in EA distance are affected by artifactual behavior. Lastly, we find that this behavior persists during transfer learning, posing a risk for safety-critical fields like medicine.

\subsubsection*{ImageNet ResNet18}
\label{paragraph:renset18}

We applied DORA to analyze the Average Pooling layer of the ResNet18 model, which consists of the $k = 512$ high-level representations that are commonly used without further modification during transfer learning. Following the DORA approach, we calculated EA$_s$ layer-wise distance with  $n=5$ s-AMS per each representation and with $m=500$, based on our findings in the section \ref{sec:hyperparameter_search}. After calculating the EA$_s$ distances, we used the LOF method with a contamination parameter $p = 0.01$ (corresponding to the top 1\% of representations), and the number of neighbors was set to 20.

\begin{figure*}[h!]
\begin{center}
\centerline{\includegraphics[width=\textwidth]{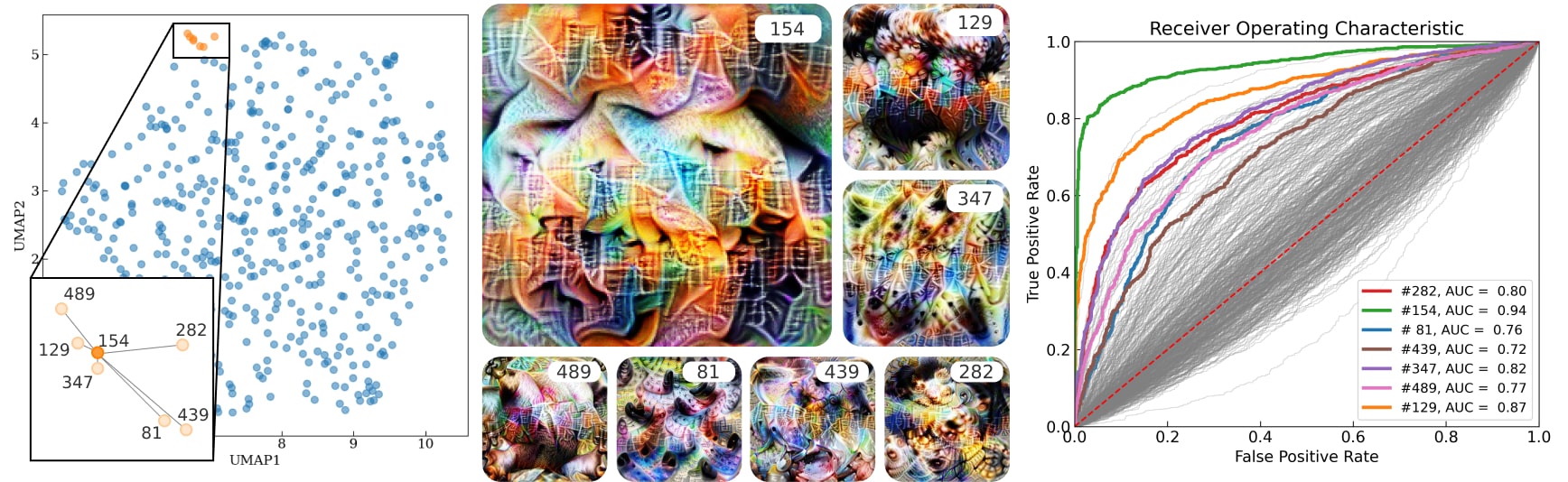}}
\caption{\textbf{Cluster of Clever-Hans representations in the ResNet18 feature extractor.}
From left to right: representation atlas of the ResNet18 average pooling layer with the highlighted cluster of Clever-Hans representations (left), s-AMS of the representations in the cluster (middle), and AUC ROC sensitivity scores for the detection of images with Chinese watermarks in the binary classification problem (right), where colored curves correspond to the behavior of representations in the cluster and gray curves for other representations. From the s-AMS of neuron 154, we can observe symbolic patterns resembling Chinese logograms learned by the neuron as well as by its closest neighbor neurons. We can observe that the outlier neuron 154 exhibits the highest AUC value (green curve), followed by its nearest neighbors.
}
\vspace{-0.5cm}
\label{fig:resnet_ch_cluster}
\end{center}
\end{figure*}

LOF identified five outlier representations, namely neurons 7, 99, 154, 160, 162, and 393. The outlier neuron 154, displayed a specific, recognizable pattern in s-AMS that could be perceived as the presence of Chinese logograms. By probing the network on a binary classification problem between images watermarked with Chinese logograms vs normal images, Neuron 154 showed the highest detection rate (AUC ROC of 0.94) towards the class with watermarked images, providing significant evidence that this representation is susceptible to the Clever-Hans effect. Further analysis of neighboring representations in EA$_s$ distance showed that they also exhibit similar behavior. The results of the analysis of the ResNet 18 average pooling layer are shown in Figure \ref{fig:resnet_ch_cluster}, illustrating the cluster of Clever-Hans representations found, along with their s-AMS and AUC ROC performance on the binary classification problem. Additional information on the dataset generation and the identified outlier representations can be found in the Appendix. Note that in general, the presence of such artifacts could indeed pose serious risks and may lead to a degradation in classifier performance (see \cite{anders2022finding}).

In the further investigation of the model, we inferenced s-AMS signals of representations in the reported CH-cluster and obtained their predictions by the model. Among the selected signals, the model predominantly predicted an affiliation of these signals with the classes ``carton'', ``swab'', ``apron'', ``monitor'' and ``broom'', which is in line with the reported spurious correlation of the ``carton'' class and Chinese watermarks \cite{li2022Dilemma}. Upon computing the corresponding s-AMS signals for these logits, we were able to confirm their association with CH-behaviour, as they displayed clear, visible logographic patterns, specific to Chinese character detectors, in their corresponding s-AMS. Corresponding signals and additional information could be found in Appendix.

\subsubsection*{ImageNet MobileNetV2}
\label{paragraph:renset18}

\begin{figure*}[h!]
\begin{center}
\centerline{\includegraphics[width=\textwidth]{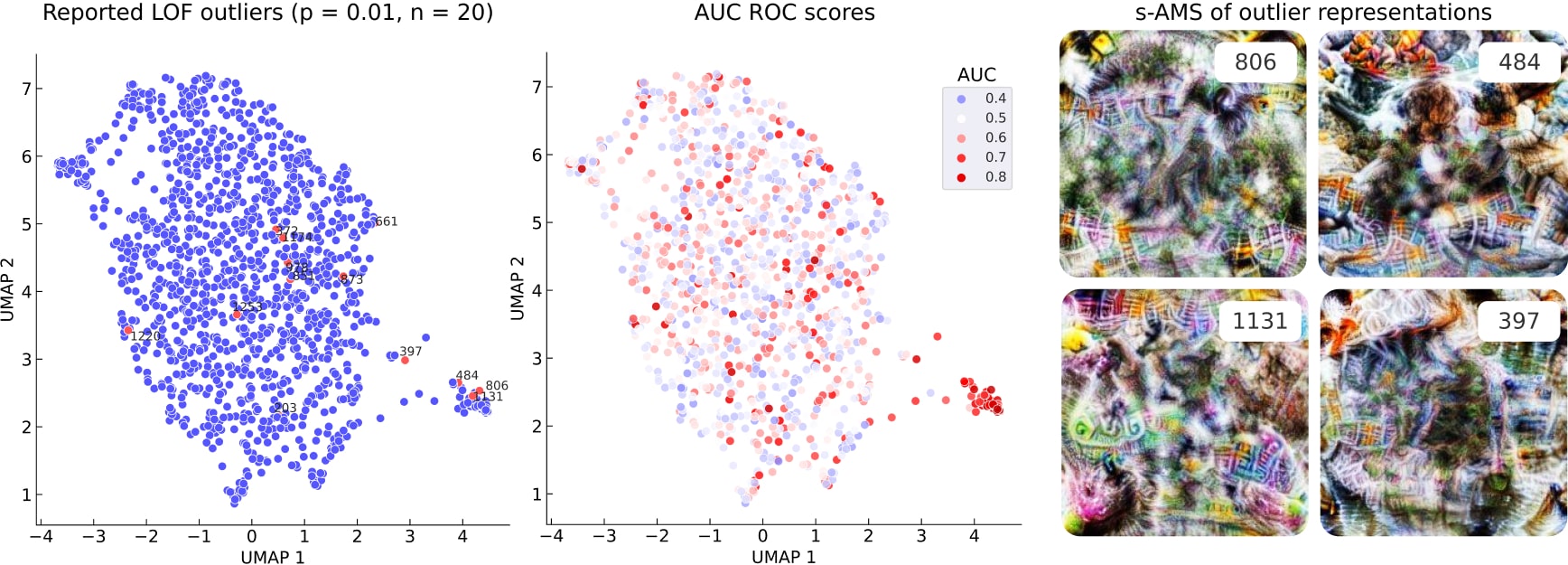}}
\caption{\textbf{Cluster of Clever-Hans representations in the MobileNet V2 feature extractor.} The left figure illustrates the outlier representations as identified by the LOF OD method, overlaid on the DORA representation atlas. The middle figure displays the sensitivity of the neural representations to Chinese watermarks, where the highly-sensitive cluster of neurons can be clearly observed in the bottom-right part of the atlas, including 3 reported outlier representations. The right graph illustrates the s-AMS of several of the reported outlier neurons, which exhibit a distinctive logographic pattern typical of Chinese character detectors.
}
\vspace{-0.5cm}
\label{fig:mobilenet_ch_cluster}
\end{center}
\end{figure*}

We used DORA with the same parameters as in the previous experiment ($n=5$ s-AMS per each representation and  $m=500$ epochs for s-AMS generation) to analyze the ``features'' layer of MobileNetV2 network \cite{sandler2018mobilenetv2}, which consists of $k = 1280$ channels with $7 \times 7$ activation maps. The analysis was performed on channels by averaging the resulting activation maps of neurons. We calculated the EA distances between representations and applied the LOF method with a contamination parameter of 0.01 which yielded 13 outlier representations. Upon visual inspection of the s-AMS of these representations, we observed distinct patterns specific to Chinese character detectors in neurons 397, 484, 806, and 1131. Figure \ref{fig:mobilenet_ch_cluster} illustrates the s-AMS of these neurons, as well as the sensitivity of neurons in the Chinese-character detection task. We can observe that the neighbors of these neurons (397, 484, 806, 1131) are sensitive to CH artifacts and form a distinctive cluster visible in the representation atlas.

\subsubsection*{ImageNet DenseNet 121}
\label{repr_detector}

\begin{figure*}[h!]
\begin{center}
\centerline{\includegraphics[width=\textwidth]{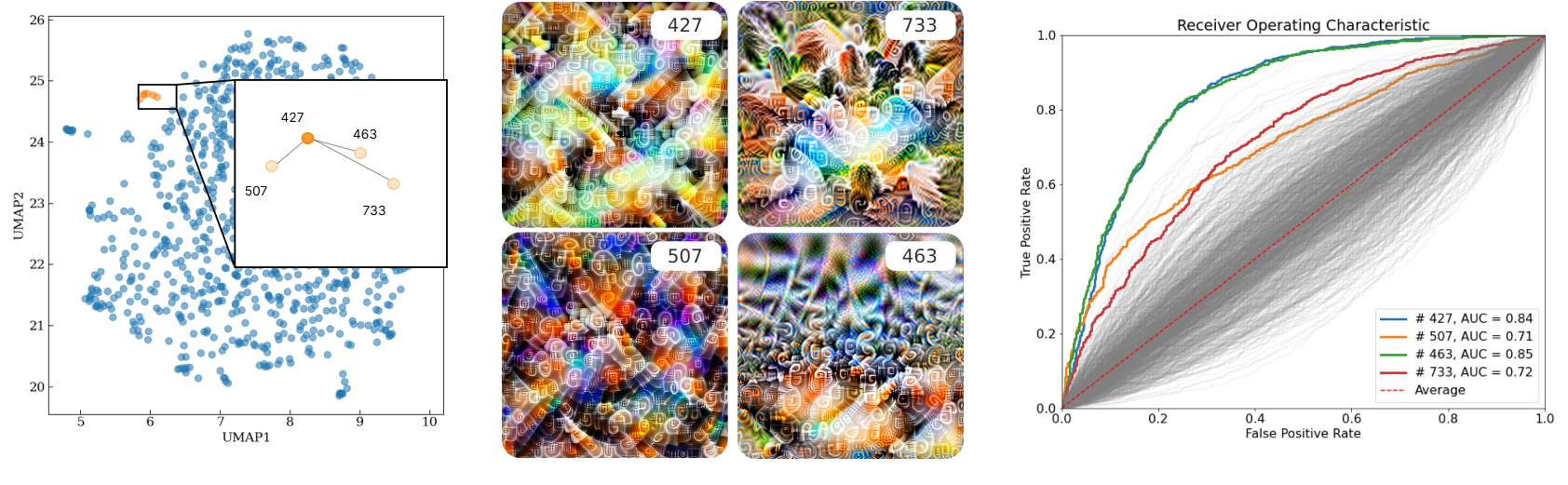}}
\vspace{-0.25cm}
\caption{\textbf{DenseNet121 --- Latin text detector.} Applying DORA to the last layer of the feature extractor of DenseNet121 yields, among others, Neuron 427 as an outlier, which corresponds to the upper left of the 4 feature visualizations. From neuron 427 as well as from its three closest neighbors (shown left), we can observe semantic concepts resembling Latin text characters. The AUC values were computed using the average channel activations on the Latin probing dataset. As shown, the AUCs are high for the representation outliers found by DORA, compared to most of the other representations, which indicates that they indeed learned to detect Latin text patterns.}
\label{fig:densenet121_outliers_original_latin}
\vskip -0.3in
\end{center}
\end{figure*}

We conducted a similar analysis on the last layer of the feature extractor of the ImageNet pre-trained DenseNet121 model, which consists of $k = 1024$ channel representations with $7 \times 7$ activation maps. We calculated $n = 5$ s-AMS per representation with $m=150$ optimization steps for faster experimentation. The LOF outlier detection method with a contamination parameter of $p = 0.01$ identified 10 outlier representations. One of these, neuron 768, was found to be a Chinese character detector (more information can be found in the Appendix). By increasing the contamination parameter to $p = 0.035$ (corresponding to the top 3.5\% or 35 representations), we also identified neuron 427, which is susceptible to the detection of Latin text and watermarks. Figure \ref{fig:densenet121_outliers_original_latin} illustrates the representation atlas, highlighting representation 427 along with several neighboring representations, namely neurons 733, 507, and 463, which also exhibit a high detection rate for unintended concepts.

\label{subsubsection:clever_hans_densenet121}
Given the widespread use of pre-trained models in safety-critical areas, it is essential that the artifacts embodied in a pre-trained model are made ineffective or unlearned during the transfer learning task (see also \cite{anders2022finding}).
To this end, we examined the effect of fine-tuning the pre-trained DenseNet121 model on the CheXpert challenge \cite{irvin2019chexpert}, which benchmarks classifiers on a multi-label chest radiograph dataset. Despite the modification of all model parameters during fine-tuning, neurons 427 and 768, which were Latin and Chinese characters detectors in the pre-trained model, retained their original artifact-detection capabilities and remained outliers after applying DORA. We studied neuron 427's ability to detect Latin text and found that it had an AUC value of 0.84 in the pre-trained model and 0.81 in the fine-tuned model. Similar behavior was observed with neuron 768, indicating that the Clever-Hans effect persisted after fine-tuning.

\subsection{CLIP ResNet50}

\begin{wrapfigure}{r}{0.4\textwidth}
\vspace{-0.45cm}
\centering
\includegraphics[width=0.4\textwidth]{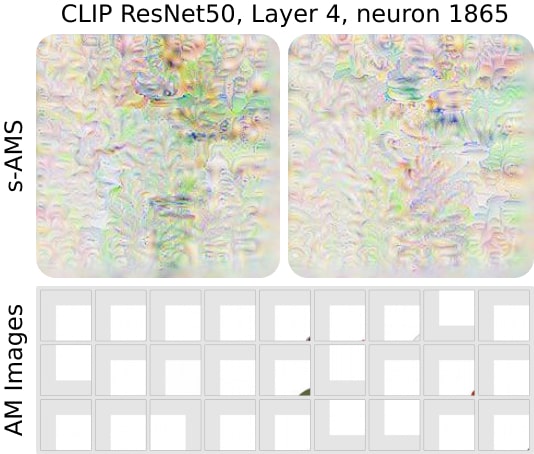}
\vspace{-0.55cm}
\caption{\textbf{AMS for reported outlier representation}. LOF identified neuron 1865 as the strongest outlier. Analysis of s-AMS and most activating images from ImageNet (obtained from \texttt{OpenAI Microscope}) indicate that it primarily detects white images/backgrounds, which is atypical compared to other high-level representations.}

\label{fig:clip_outlier}
\end{wrapfigure}

CLIP (Contrastive Language-Image Pre-training) models, which are designed to predict the associations between text and images, are trained using a contrastive learning objective \cite{dai2017contrastive, hjelm2018learning} on extensive datasets. They are often fine-tuned for tasks like image classification \cite{agarwal2021evaluating} or text-to-image synthesis, where CLIP models frequently function as text encoders (e.g., Stable Diffusion \cite{Rombach_2022_CVPR}).

In this experiment, we explored the representation space of the CLIP ResNet50 model \cite{radford2021learning}, with particular emphasis on the final layer of its image feature extractor (referred to as ``layer 4''). While the training dataset was not publicly revealed, it is known to be significantly larger than standard computer vision datasets like ImageNet, leading to a broader range of concepts compared to ImageNet networks. We applied DORA to the 2048 channel representations from "layer 4", generating $n=3$ signals per representation with $m=512$ and employing settings akin to those in \citep{goh2021multimodal}.

Analysis of the outlier representations with contamination parameter $p = 0.0025$ yielded 6 outlier neurons, namely 631,  658,  838, 1666, 1865, and 1896. Representation 1865 -- neuron with the highest outlier score --  was found to detect the unusual concept of white images/background, as shown by s-AMS and most activating images  (collected from \texttt{OpenAI Microscope}) in the Figure \ref{fig:clip_outlier}. However, the other outlier representations could not be concluded to be undesirable as they seemed to detect rare but natural concepts. Further details and analysis of the other outlier representations can be found in the Appendix.

\begin{figure}
\begin{center}
\centerline{\includegraphics[width=\columnwidth]{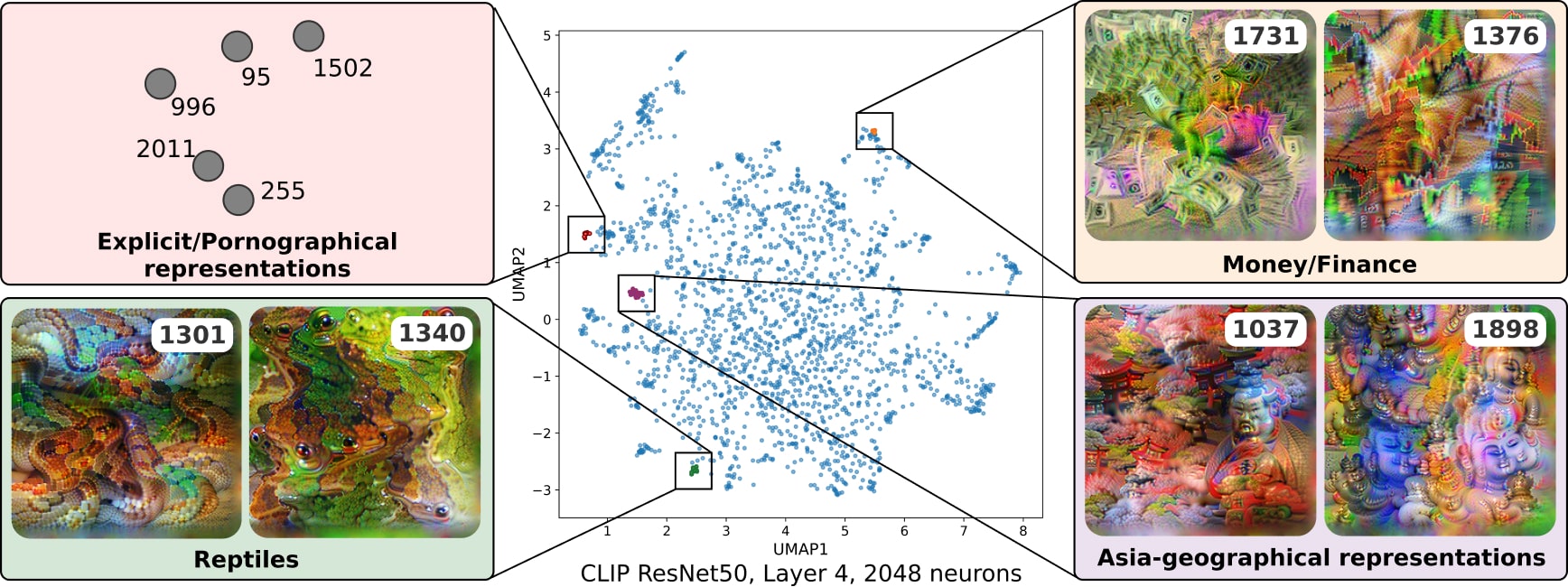}}
\caption{\textbf{Representation atlas of CLIP ResNet50 ``layer 4''.} Representation atlas for CLIP ResNet50 ``layer 4'', where several clusters of representations are highlighted. Activation-Maximization signals associated with the Explicit/Pornographic representations were omitted due to the presence of explicit concepts in the signals.
}
\label{fig:clip_atlas}
\end{center}
\vspace{-0.5cm}
\end{figure}

After computing the representation atlas for ``layer 4'', we manually investigated several distinctive clusters. Figure \ref{fig:clip_atlas} illustrates the representation atlas alongside several reported clusters of semantically similar representations. With our analysis, we found a cluster of Explicit/Pornographic representations. Furthermore, we were able to confirm the presence of geographical neurons, as reported in \citep{goh2021multimodal} and we noted that representations from neighboring geographical regions, such as India, China, and Japan, were located close to one another. Additional information and more detailed visualizations can be found in the Appendix.

\section{Discussion and Conclusion}

The popularity of Deep Neural Networks (DNNs) across diverse fields has brought to light the significant challenge posed by their inherent opacity, particularly for the fair and responsible deployment of DNNs. The presence of artifacts, spurious correlations, or biases in datasets is not a rare occurrence. Therefore, it has become increasingly crucial to audit these models using Explainable Artificial Intelligence (XAI) methods to avert potential undesirable or even harmful behavior. To date, audits have primarily employed local XAI methods, which necessitate data access to elucidate the predictions of a given model and are often found to be limited when it comes to uncovering new potential biases \cite{adebayo2022post}. Prior to our research, as far as our knowledge extends, there existed no methods for identifying representations that had inadvertently or malevolently learned unintended concepts.

In our work, we introduced a general \textit{Representation Analysis} pipeline for exploration of the relationships between neural representations within a specific layer and introduced DORA as a special case of data-agnostic analysis. The core of our framework is the newly introduced \textit{Extreme-Activation} (EA) distance measure, which allows us to measure the similarity between concepts, learned by representations within the model. This distance measure is easy to interpret and it allows us to analyze relationships between representations in the layer, including the identification of the anomalous representations, that we demonstrate in practice often correspond to the undesired spurious concepts.

Although we have demonstrated the broad applicability of DORA, there exist several limitations that require attention. Firstly, the proposed approach assumes that undesired behavior in representations is not systematic. Consequently, DORA may not be able to identify infected representations if such behavior is widespread across a large number of representations, as it would no longer be considered anomalous. Another limitation pertains to the potential semantic multimodality of representations \cite{goh2021multimodal} and additional research is necessary to address this issue.

In summary, DORA expands the scope for auditing "black-box" systems, thereby offering a methodology that enhances the understanding of learned representations within the model and their interrelationships. By facilitating a deeper dive into these complex systems, Representation Analysis makes it possible to demystify the intricacies of the internal model behavior and consequently leads to a more transparent and accountable machine learning process, encouraging robustness and trustworthiness in the deployment of these models.

\section*{Acknowledgements}
We would like to thank Filip Rejmus for his analysis regarding the visualization of the Clever Hans (CH) behavior in representations with global explanation methods. Furthermore, we would like to thank Sebastian Lapuschkin for fruitful discussions about CH behavior in Deep Neural Networks. 
\section*{Funding}
This work was partly funded by the German Ministry for Education and Research through the project Explaining 4.0 (ref. 01IS200551), the German Research Foundation (ref. DFG KI-FOR 5363), the Investitionsbank Berlin through BerDiBa (grant no. 10174498), and the European Union’s Horizon 2020 programme through iToBoS (grant no. 965221). KRM was partly funded by the German Ministry for Education and Research (under refs 01IS14013A-E, 01GQ1115, 01GQ0850, 01IS18056A, 01IS18025A, and 01IS18037A) and BBDC/BZML and BIFOLD and by the Institute of Information \& Communications Technology Planning \& Evaluation (IITP) grants funded by the Korea Government (MSIT) (No. 2019-0-00079, Artificial Intelligence Graduate School Program, Korea University and No. 2022-0-00984, Development of Artificial Intelligence Technology for Personalized Plug-and-Play Explanation and Verification of Explanation).

{\small
\bibliographystyle{abbrvnat}
\bibliography{main}
}
\appendix
\include{appendix}

\end{document}

%% file: math_commands.tex
\usepackage{amsmath,amsfonts,bm}

\def\eqref#1{equation~\ref{#1}}

\def\1{\bm{1}}

\DeclareMathAlphabet{\mathsfit}{\encodingdefault}{\sfdefault}{m}{sl}
\SetMathAlphabet{\mathsfit}{bold}{\encodingdefault}{\sfdefault}{bx}{n}

\DeclareMathOperator*{\argmax}{arg\,max}

%% file: appendix.tex
\appendix
\section{Appendix}

\subsection{Evaluation}
In the evaluation, two datasets were used: ILSVRC2012 (ImageNet 2012) \cite{deng2009imagenet} and CIFAR-100 \cite{krizhevsky2009learning}. For ImageNet, we employed eight different pre-trained models: ResNet18 \cite{he2016deep}, AlexNet \cite{krizhevsky2017imagenet}, Inception V3 \cite{szegedy2016rethinking}, DenseNet 161 \cite{huang2017densely}, MobileNet V2 \cite{sandler2018mobilenetv2}, ShuffleNet V2 \cite{ma2018shufflenet}, obtained from the \texttt{torchvision-models} package \cite{marcel2010torchvision}, as well as ViT \cite{dosovitskiy2020image} and BEiT \cite{krizhevsky2017imagenet}, obtained from the \texttt{pytorch-vision-models} library \cite{rw2019timm}. For the CIFAR-100 dataset, we trained seven networks: ResNet 18, MobileNet V2, ShuffleNet V1, and V2, NASNet \cite{qin2019nasnet}, SqueeeNet \cite{iandola2016squeezenet}, and VGG 11 \cite{simonyan2014very}, using the \texttt{Pytorch-cifar100} GitHub repository \cite{githubGitHubWeiaicunzaipytorchcifar100}, while the ResNet9 network was trained using a publicly available Kaggle notebook \cite{kaggleCIFAR100Resnet}.

The semantic baseline distances between concepts for both datasets were obtained using the \texttt{NLTK} package \cite{bird2009natural}. There is a cross-connection between class labels and WordNet entities for ILSVRC2012, as the classes are inherently connected with WordNet synsets. For CIFAR-100, we manually connected the labels to synsets by matching class label names with WordNet synset names. For 98 classes, WordNet synsets were found. For the remaining two classes, ``aquarium fish'' and ``maple tree'', WordNet synsets for ``fish'' and ``maple'' were used, respectively, due to the absence of a direct name match.

For the Word2Vec distance, we used the WordNet synset name as the textual label. If a textual label contained multiple words, the distance between two classes was determined as the maximum distance among all possible word pairs between the textual labels of the two classes.

\subsection{Experiments}
\subsubsection{Probing dataset}

\begin{figure*}[h!]
\begin{center}
\centerline{\includegraphics[width=\textwidth]{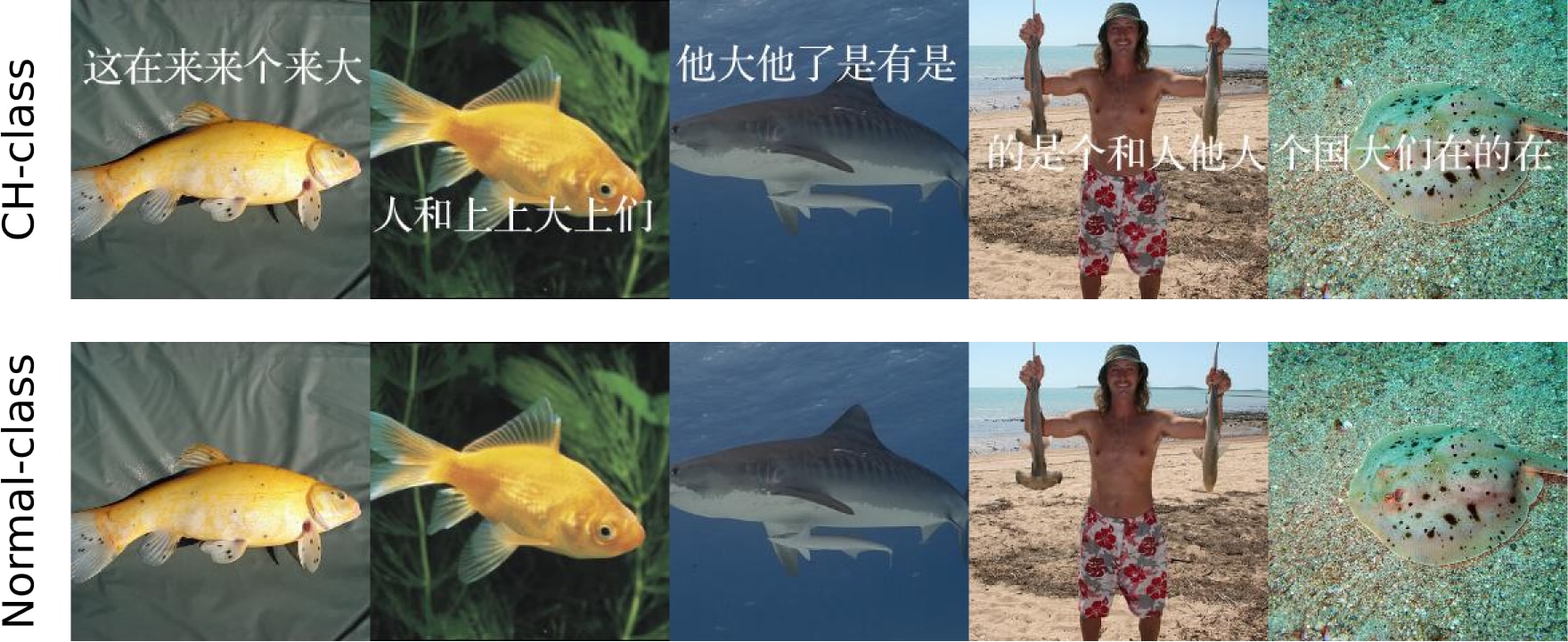}}
\caption{\textbf{Illustration of the Probing Dataset.} The figure depicts images from the probing dataset utilized to evaluate the representation's capacity to distinguish between watermarked (CH) and non-watermarked (normal) images. The watermarked class images are identical to the normal class images, except for the addition of a random test string at a random location on the image.}
\label{fig:ch_dataset}
\vskip -0.3in
\end{center}
\end{figure*}

To assess the ability of the identified representations to detect undesirable concepts, we created two probing datasets for the binary classification of Chinese and Latin text detection. We modified one class of images by adding specific watermarks while leaving the other class unchanged. We used a baseline dataset of 998 ImageNet images \footnote{Images were obtained from \url{https://github.com/EliSchwartz/imagenet-sample-images}, with the exception of two images (of the class "carton" and "terrapin") that already exhibit watermarks.} to create 2 probing datasets (Chinese and Latin) by inserting random textual watermarks, as shown in Figure \ref{fig:ch_dataset}. For the Chinese characters detection problem, the watermarks were generated by randomly selecting 7 out of the 20 most commonly used Chinese characters \cite{da2004corpus}, and a similar process was followed using the English alphabet for the Latin text detection problem. The font size for all watermarks has been set to 30, while the image dimensions remain standard at 224 $\times$ 224 pixels. AUC ROC was used as the performance metric to evaluate the representations' ability to differentiate between watermarked and normal classes. The true labels provided by the two datasets were used, where class 1 represents images with a watermark and class 0 represents images without. We computed the scalar activations for all images from both classes for a specific neural representation and then calculated the AUC ROC classification score based on the differences in activations using the binary labels. A score of 1 indicates a perfect classifier, consistently ranking watermarked images higher than normal ones, while a score of 0.5 indicates a random classifier.

\subsubsection{ImageNet ResNet18}

In the following, we provide additional details on the ResNet18 \cite{he2016deep} experiment, discussed in the main paper. The model was downloaded from the Torchvision library \cite{marcel2010torchvision} and s-AMS were generated with parameters $n=5$ and $m = 500$ using the \texttt{DORA} package.

\begin{figure}
\begin{center}
\centerline{\includegraphics[width=\textwidth]{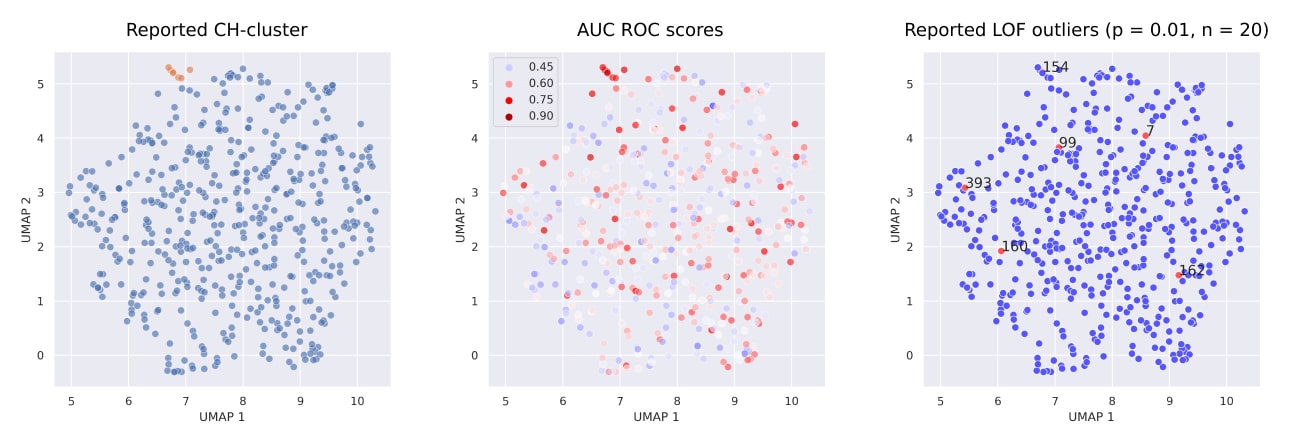}}
\caption{\protect\textbf{Detailed illustration of the cluster of malicious representations found. } All of the figures illustrate the representation atlas of the average pooling layer of ResNet18, calculated using the DORA distance metric. From left to right: illustration of the reported Chinese detector cluster, the sensitivity of different representations for detecting Chinese watermarks, and a set of reported outliers among the representations using the LOF method. From the middle figure, it can be observed that the cluster of reported representations exhibits high sensitivity towards the artifactual concept of the desired task, and the closer the representations are to the cluster in the representation atlas, the more they are able to detect malicious concepts in the data.}
\label{fig:appendix:resnet_cluster_detailed}
\end{center}
\end{figure}

\begin{figure}[h!]	
\subfigure[Neuron 154]{\includegraphics[width=0.48\textwidth]{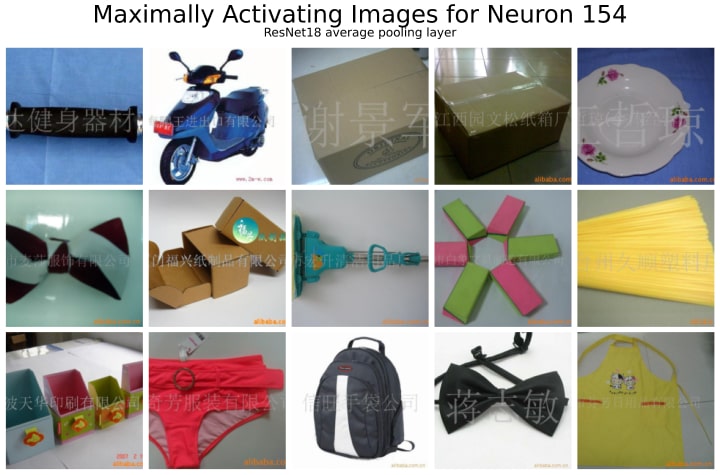}}
\hfill
\subfigure[Neuron 347]{\includegraphics[width=0.48\textwidth]{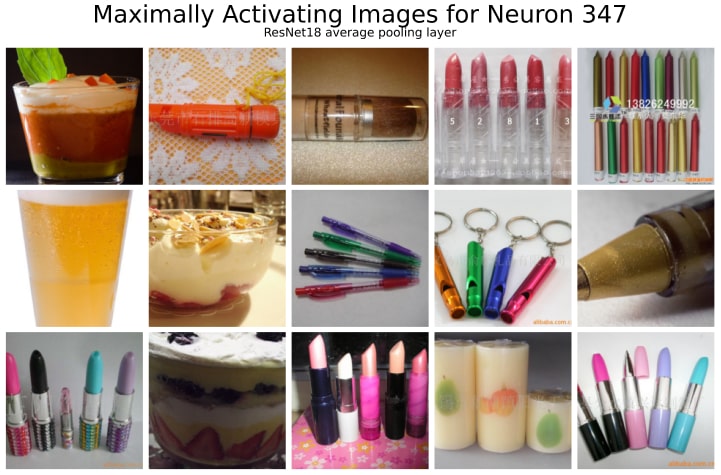}}
\subfigure[Neuron 129]{\includegraphics[width=0.48\textwidth]{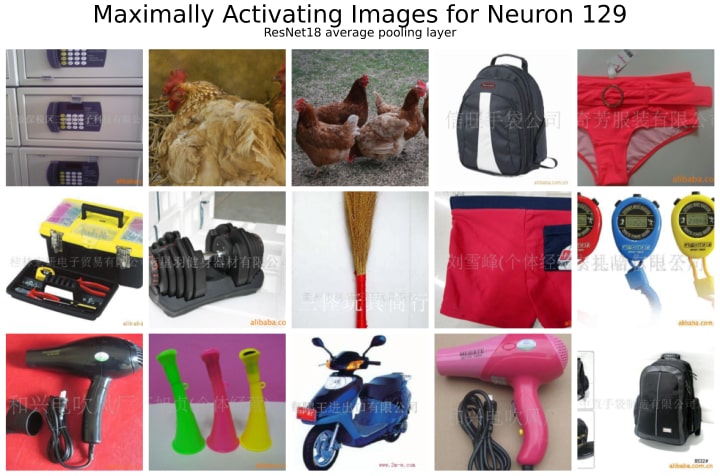}}
\hfill
\subfigure[Neuron 489]{\includegraphics[width=0.48\textwidth]{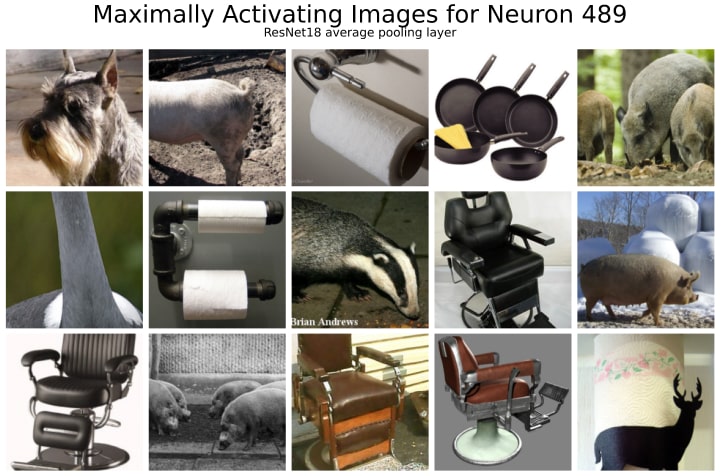}}
\subfigure[Neuron 81]{\includegraphics[width=0.48\textwidth]{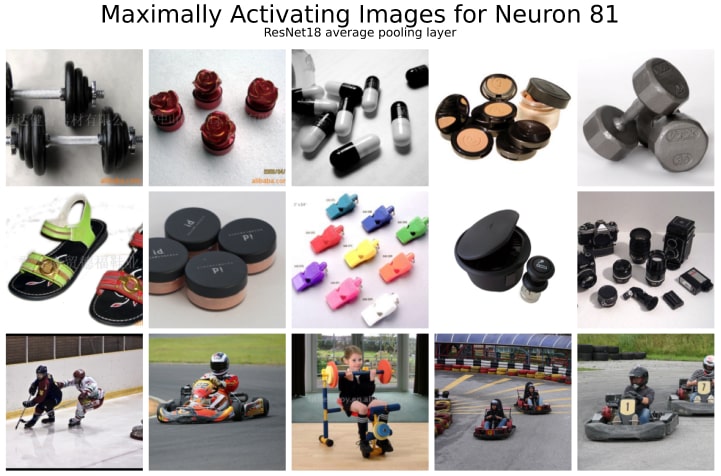}}
\hfill
\subfigure[Neuron 282]{\includegraphics[width=0.49\textwidth]{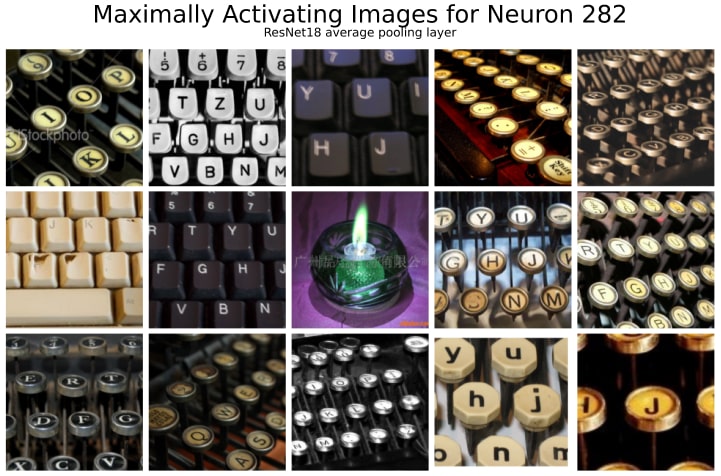}}
\caption{\textbf{Maximally Activating Images for different ResNet18 neurons, reported in the cluster of malicious representations.} The figure shows the $15$ highest Activating Images for various neurons in the "avgpool" layer of the ResNet18 network, which were identified as being in the cluster of malicious representations. The signals were calculated using a subset of 1 million images from the ImageNet 2012 training dataset. It can be observed that among the top activating images, there are images of Chinese watermarks, supporting the hypothesis that these neurons have learned undesirable concepts.}
\label{fig:appendix:resnet_n-AMS}
\end{figure}

Figure \ref{fig:appendix:resnet_cluster_detailed} illustrates the cluster of reported representations in the average pooling layer of the model, specifically neurons 154, 129, 347, 489, 81, 439, and 282, along with the sensitivity of other neurons to Chinese watermarks. It can be seen that representations close to the reported cluster also exhibit sensitivity towards malicious concepts. For additional context, Figure \ref{fig:appendix:resnet_n-AMS} shows the natural Activation-Maximisation signals (n-AMS) for the reported representations, obtained using 1 million subsamples of the ImageNet 2012 train dataset. The presence of Chinese watermarks in the n-AMS further supports our hypothesis of the Clever-Hans nature of these representations.

To examine which output class logits may be compromised by Clever-Hans (CH) behavior, we used the s-AMS of the reported neurons to obtain class predictions on these signals. Figure \ref{fig:appendix:resnet18_probing} shows several s-AMS for the reported representations along with the network's predictions for the corresponding data points. We observed that certain classes, such as ``carton'' (478), ``apron'' (411), ``swab, swob, mop'' (840), ``monitor'' (664), and ``broom'' (462) were frequently predicted with high scores. When we computed the s-AMS for selected output logits, we found similar Chinese patterns, similar to those observed in the reported neurons of the average pooling layer (see Figure \ref{fig:appendix:resnet18_probing}). These results suggest that such artifacts learned by the network pose a potential threat to applications due to the network's tendency to classify images with added watermarks as belonging to one of these classes.

\subsubsection{ImageNet DenseNet121}

\begin{figure*}
\vspace{-0.25cm}
\centering
\includegraphics[width=0.45\textwidth]{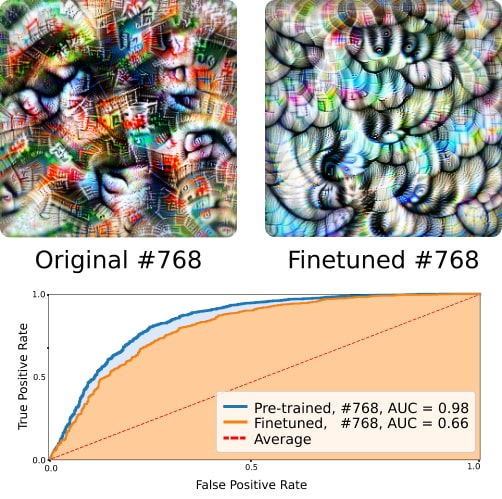}
\caption{\protect\textbf{Survived Chinese-characters detector}. Neuron 768 learns to detect Chinese logographic symbols during pre-training (top left) and does not unlearn this behavior during fine-tuning on the CheXpert dataset (top right). The AUC values of the neurons' activation on images corrupted with Chinese watermarks are still high after pre-training.}
\label{fig:densenet121_outliers_chinese_finetuned}
\end{figure*}

\begin{figure*}[h!]
\begin{center}
\centerline{\includegraphics[width=\textwidth]{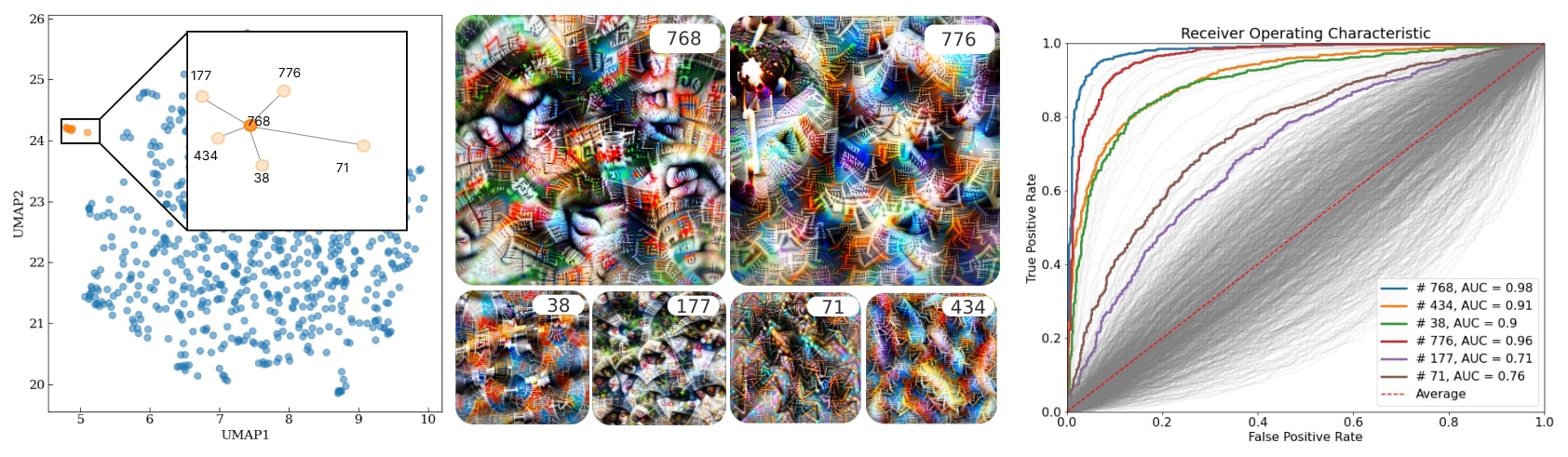}}
\caption{\textbf{DenseNet121 --- Chinese-characters detector.} Applying DORA to the last layer of the feature extractor of DenseNet121 yields, among others, Neuron 768, which corresponds to the upper left of the 6 feature visualizations. From Neuron 768 as well as from its five closest neighbors (shown left), we can observe semantic concepts resembling Chinese logograms. The AUC values were computed using the channel activations on a data set that was corrupted with watermarks written in Chinese. As shown, the AUCs are high for the representation outliers found by DORA, compared to most of the other representations, which indicates that they indeed learned to detect Chinese logograms.}
\label{fig:densenet121_outliers_original_chinese}
\vskip -0.3in
\end{center}
\end{figure*}

\begin{figure}[h!]
\hfill

\subfigure[Neuron 768]{\includegraphics[width=0.5\textwidth]{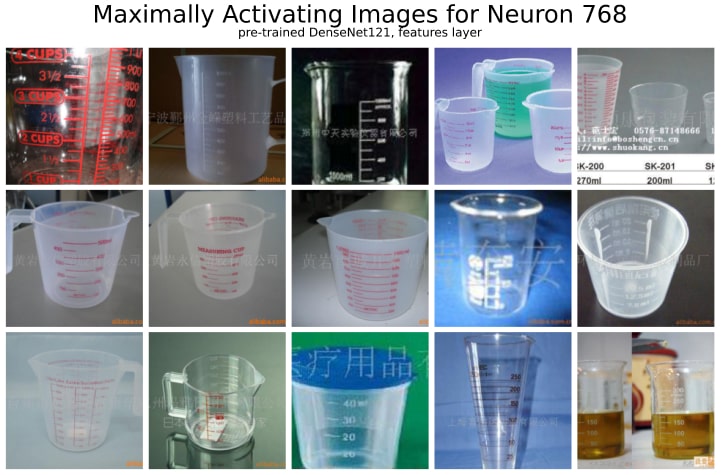}}
\subfigure[Neuron 427]{\includegraphics[width=0.5\textwidth]{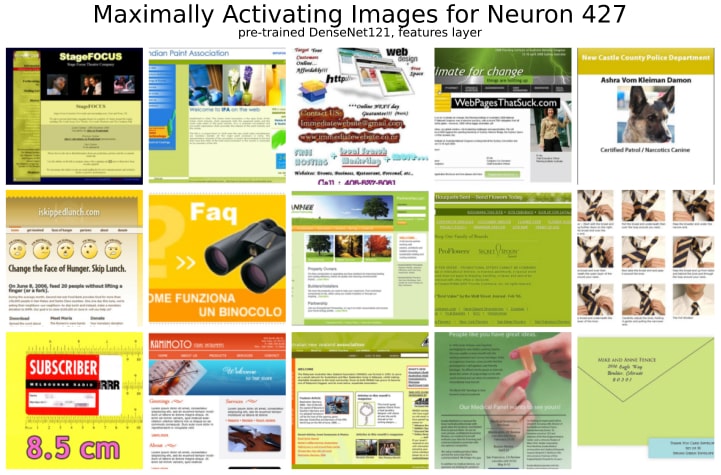}}
\caption{\textbf{Maximally Activating images for different neurons in DenseNet121.} Illustration of the top $15$ images that trigger the highest activations for the Chinese watermark detector (neuron 768) and the Latin text detector (neuron 427) in the 'features' layer of DenseNet121.}
\label{fig:densenet121_nams}
\end{figure}

The DORA framework was employed to investigate the pre-trained DenseNet121 on the ImageNet dataset \cite{huang2017densely}. Specifically, attention was focused on the last layer of the feature extractor, which comprised 1024 channel representations. The study primarily examined two outliers detected by DORA: neuron 768 and neuron 427, along with some of their nearest neighbors in the EA distance. Following an analysis of the s-AMS for both neurons, specific symbolic patterns were observed, which were characteristic of character detectors. Neuron 768 was identified as a Chinese character detector, while neuron 427 was identified as a Latin text detector. Figure \ref{fig:densenet121_outliers_original_latin} in the main paper and Figure \ref{fig:densenet121_outliers_original_chinese} depict these neurons, along with their closest neighbors in EA distance, which exhibited similar properties. The hypothesis was further supported by visualizing the n-AMS across the ImageNet dataset, as demonstrated in Figure \ref{fig:densenet121_nams}.

As mentioned in Section \ref{repr_detector}, we find that the outliers found by DORA are maintained during fine-tuning on another dataset, e.g. the CheXpert challenge.
The CheXpert challenge benchmarks various deep learning models on the task of classifying multilabel chest radiographs and additionally provides human experts, e.g. radiologists, with performance metrics for comparison. The data set itself consists of 224,316 training, 200 validation, and 500 test data points. The current best approach in terms of AUC-ROC score uses an ensemble of five DenseNet121's \cite{huang2017densely} that were pre-trained on the ImageNet dataset and fine-tuned by optimizing a special surrogate loss for the AUC-ROC score \cite{Yuan2021auc}. The training code can be found in this public repository \url{https://github.com/Optimization-AI/LibAUC/}. We choose to finetune one DenseNet121 using this approach on a downsampled version of the CheXpert data with a resolution of 256x256x3. The converged model yields an AUC-ROC score of 87.93\% on the validation dataset. Having the finetuned DenseNet121 and the outlier neuron 768 at hand we show the Feature Visualizations and the AUC-ROC curves for both the pre-trained and fine-tuned channel on an ImageNet subset with both uncorrupted and corrupted images with Chinese watermarks in Figure \ref{fig:densenet121_outliers_chinese_finetuned}.

\subsubsection{CLIP ResNet 50}

The s-AMS for the CLIP ResNet 50 was computed using the same parameters as \citet{goh2021multimodal} with the Lucent library. The number of optimization steps $m$ was set to 512. The analysis was conducted on representations (channels) from the ``layer 4'' layer of the model. (Details on the s-AMS generation parameters can be found at \url{https://github.com/openai/CLIP-featurevis} and Lucent library at \url{https://github.com/greentfrapp/lucent})

\textbf{Star Wars representation}

\begin{figure*}[h!]
\begin{center}
\centerline{\includegraphics[width=\textwidth]{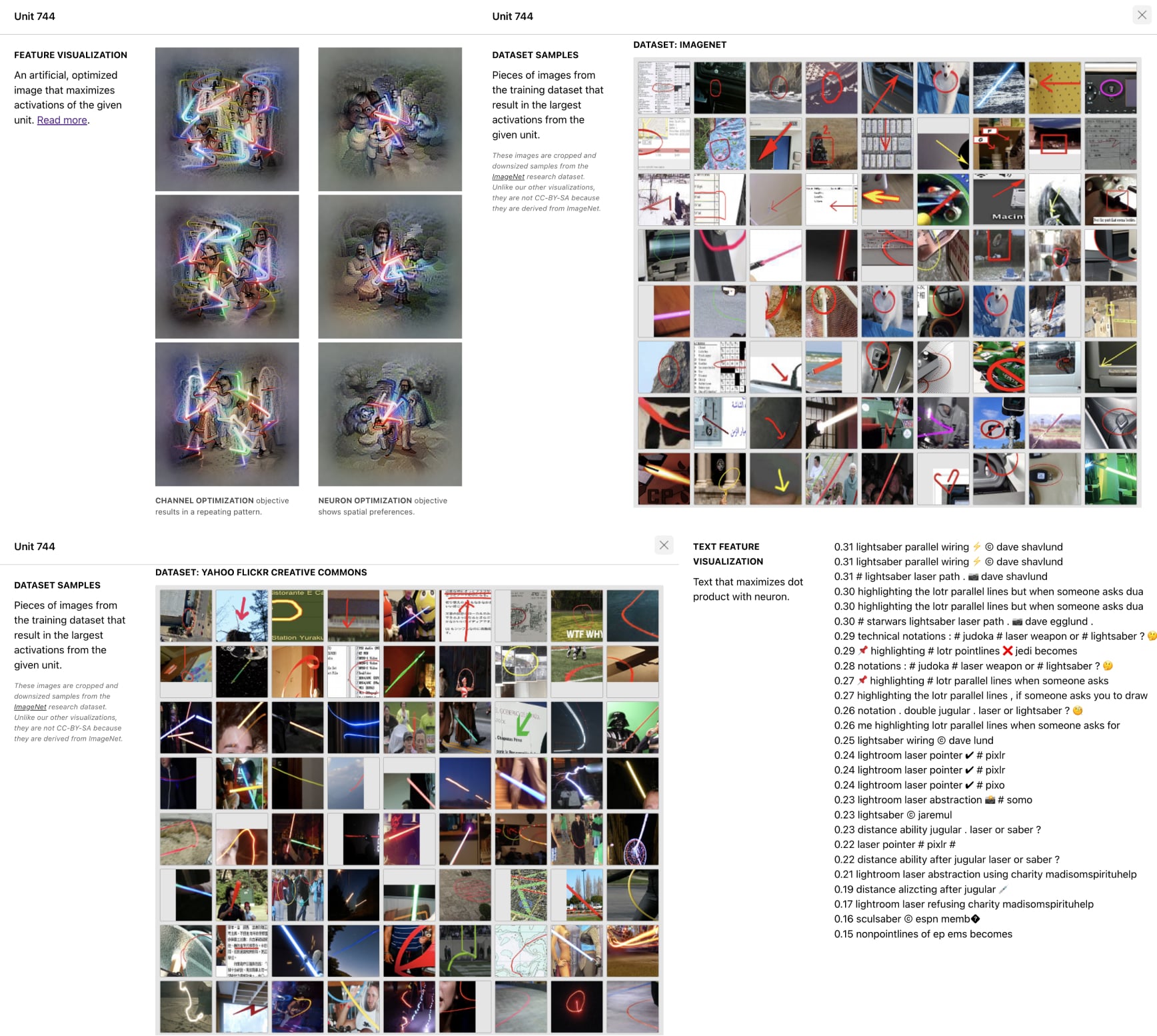}}
\caption{\textbf{CLIP ResNet Neuron 744.} 
The figure shows s-AMS and Maximally Activating Images for neuron 744 in the ``layer 4'' layer of the model, computed for 2 different data corpora. The observed signals and explanations from Text Feature Visualization confirm that the neuron can detect Star Wars-related concepts. Results obtained from \texttt{OpenAI Microscope}.
}
\label{fig:appendix:star-wars}
\end{center}
\end{figure*}

Figure \ref{fig:starwars} shows the limitations of the n-AMS approach when the data corpus for analysis differs from the training dataset. Figure \ref{fig:appendix:star-wars} further illustrates n-AMS collected from ImageNet and Yahoo Creative Commons \cite{thomee2016yfcc100m} datasets via \texttt{OpenAI Microscope}. Text Feature Visualization \cite{goh2021multimodal} supports our hypothesis that the model is a detector of Star Wars-related concepts.

\textbf{Outlier representations}

\begin{figure*}[h!]
\begin{center}
\centerline{\includegraphics[width=0.9\textwidth]{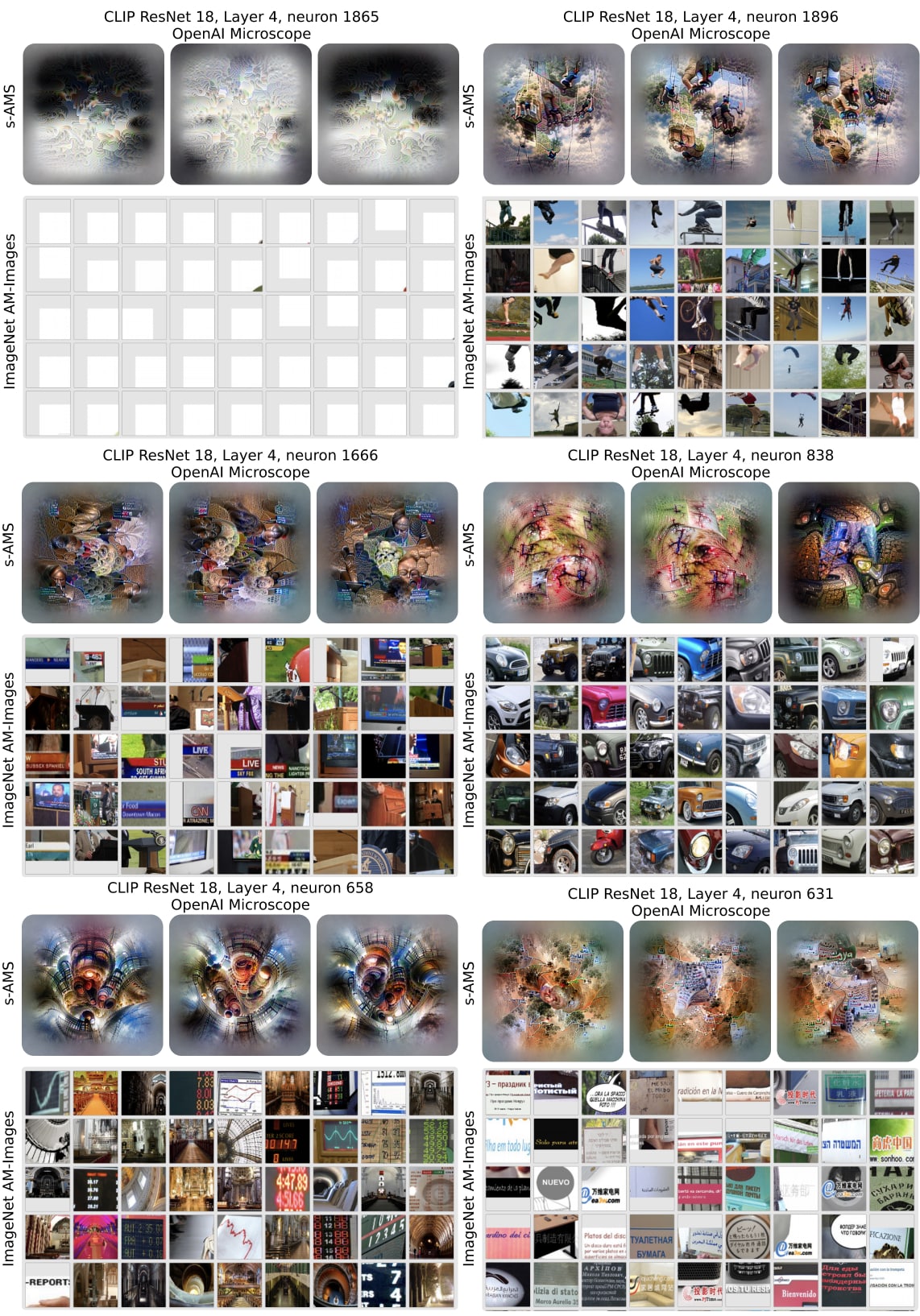}}
\caption{\textbf{s-AMS and Maximally Activating Images for reported outlier neurons.} Figure illustrates s-AMS and Maximally Activating Images for the reported outlier neurons in the ``layer 4'' layer of the CLIP ResNet 50 model, collected from \texttt{OpenAI Microscope}.
}
\label{fig:appendix:clip-outliers}
\end{center}
\end{figure*}

Analysis of the representations space of the CLIP model yielded a number of potential candidates to be considered outlier representations, namely neurons 631,  658,  838, 1666, 1865, and 1896. In Figure \ref{fig:appendix:clip-outliers} we illustrate 3 s-AMS signals, alongside n-AMS images, collected from the ImageNet dataset per each reported representation, collected using \texttt{OpenAI Microscope}. While it is hard to explain the anomalous nature of neurons 631,  658,  838, 1666, and 1896, we can clearly observe how different the concept of neuron 1865 is.

\textbf{Clusters of representations}

\begin{figure*}[h!]
\begin{center}
\centerline{\includegraphics[width=0.9\textwidth]{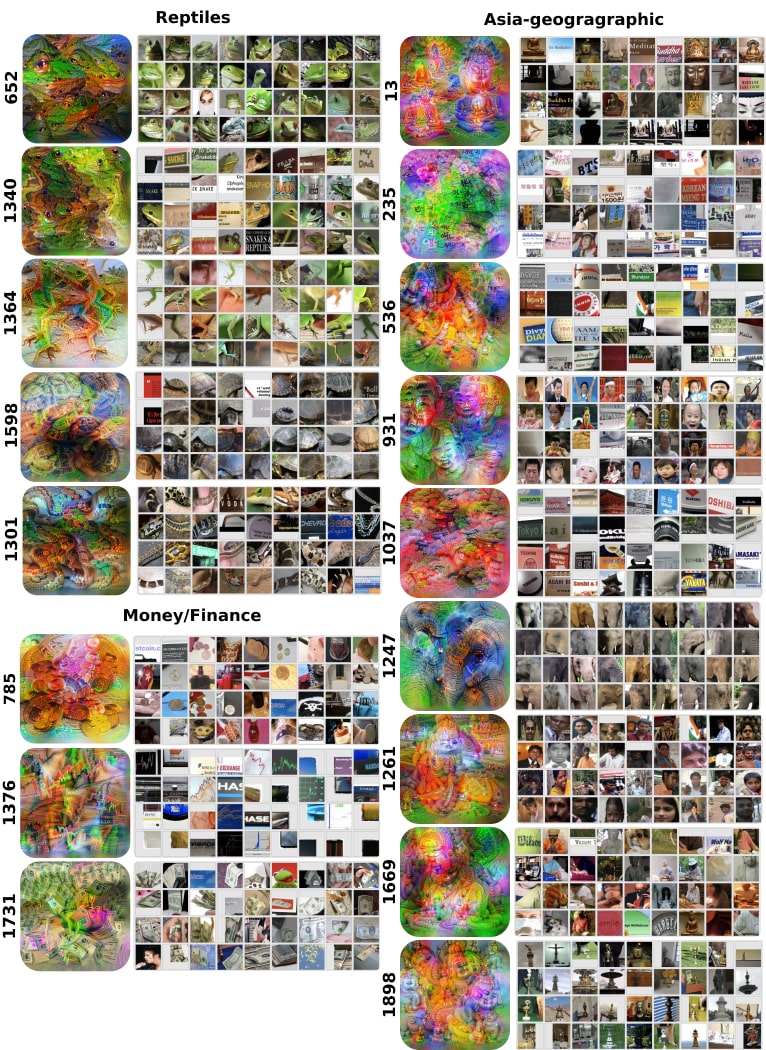}}
\caption{\textbf{s-AMS and maximally activating images from ImageNet for the neurons in the reported clusters.} This figure shows the s-AMS and maximally activating images for representations assigned to the various reported clusters. The s-AMS were generated, while the maximally activating images from ImageNet were collected via the \texttt{OpenAI Microscope}. Representations of explicit or pornographic content were excluded due to the presence of obscene images.
}
\label{fig:appendix:clip-cluster-representations}
\end{center}
\end{figure*}

We manually examined several distinctive classes of representations in ``layer4''  of the CLIP model after computing the representation atlas for the channel representations. Table \ref{tab:appendix:clip_clusters} summarizes the results of our analysis and shows interesting clusters found along with the associated neurons. Figure \ref{fig:appendix:clip-cluster-representations} shows synthetic and natural AMS, providing evidence for the assignment of neurons to their respective clusters.

\begin{table}
\centering
\caption{\textbf{Clusters of CLIP ``layer4'' representations.} This table presents several interesting clusters and the indexes of the corresponding representations that were examined through manual inspection of the s-AMS and most activating images.}

\label{tab:appendix:clip_clusters}
\begin{tabular}{lll} 
\toprule
\textit{Cluster}      &  & \textit{Representations}                                                                                                                                              \\ 
\cline{1-1}\cline{3-3}
                      &  &                                                                                                                                                                       \\
Explicit/Pornographic &  & 95, 255, 996, 1502, 2011                                                                                                                                              \\
Money/Finance         &  & 785, 1376, 1731                                                                                                                                                       \\
Reptiles              &  & \begin{tabular}[c]{@{}l@{}}230, 250, 417, 521, 652, 654, 694, 1008, 1234,~1301,~1340,\\1364, ~1445, 1598~\end{tabular}                                                 \\
Fish/Aquarium         &  & 1193, 1384                                                                                                                          \\
Asia-geographic~      &  & \begin{tabular}[c]{@{}l@{}}13,~165,~235,~536,~780,~931,~1037,~1261,~1247,~1423,\\1669,~1761,~1874,~1898\end{tabular}  \\
\bottomrule
\end{tabular}
\end{table}

\subsection{Experimental setup}
All described experiments, if not stated otherwise, were performed on the Google Colab Pro \cite{bisong2019google} environment with the GPU accelerator.

\begin{figure}
\centerline{\includegraphics[width=\textwidth]{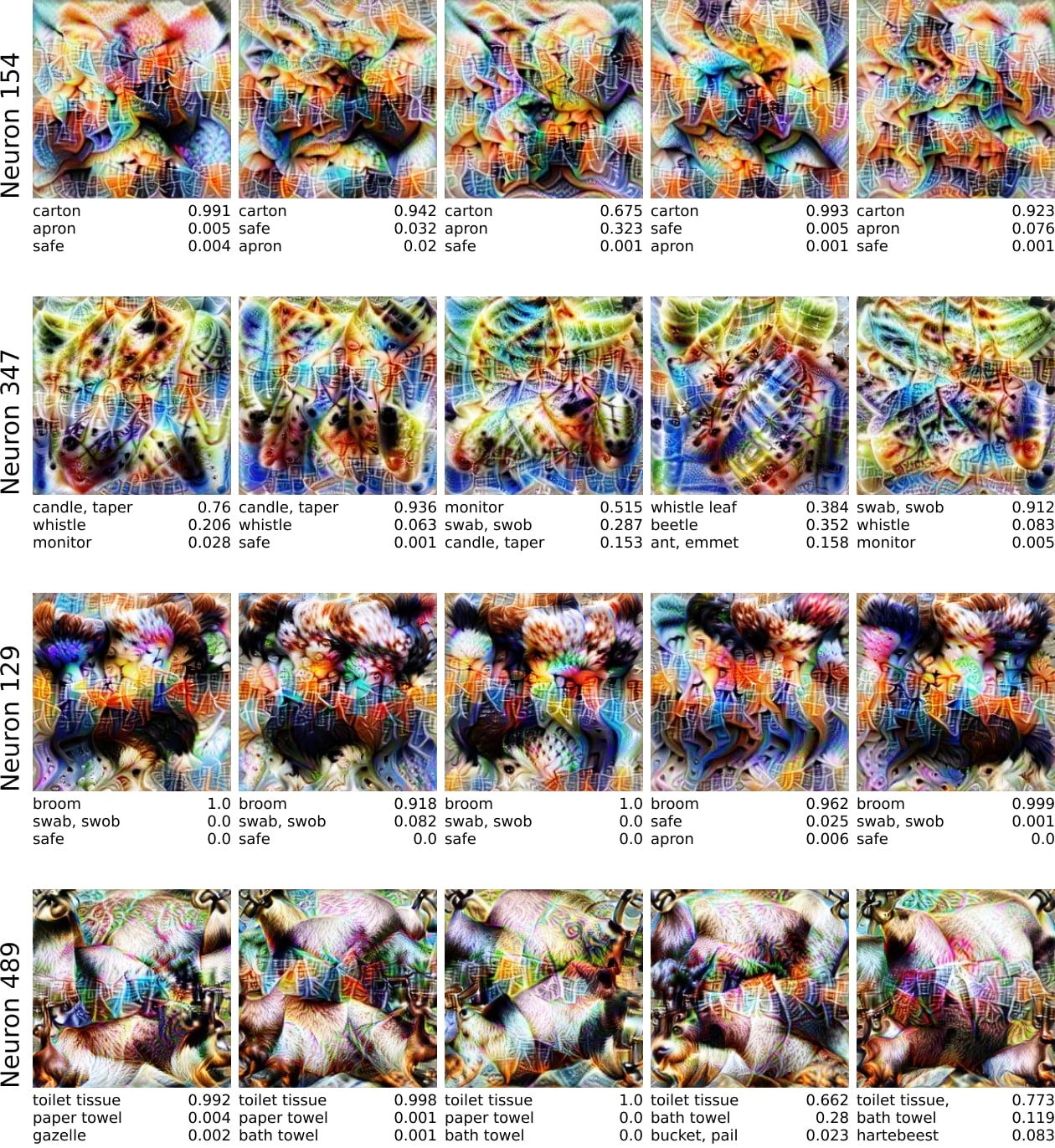}}
\caption{\protect\textbf{s-AMS and model predictions for reported neurons in ResNet18.} Figure illustrates the s-AMS signals for four different reported neurons in the average pooling layer of ImageNet-trained ResNet18, along with the model's predictions for the top three classes with their respective softmax scores.}
\label{fig:appendix:resnet18_probing}
\end{figure}

\begin{figure}
\centerline{\includegraphics[width=\textwidth]{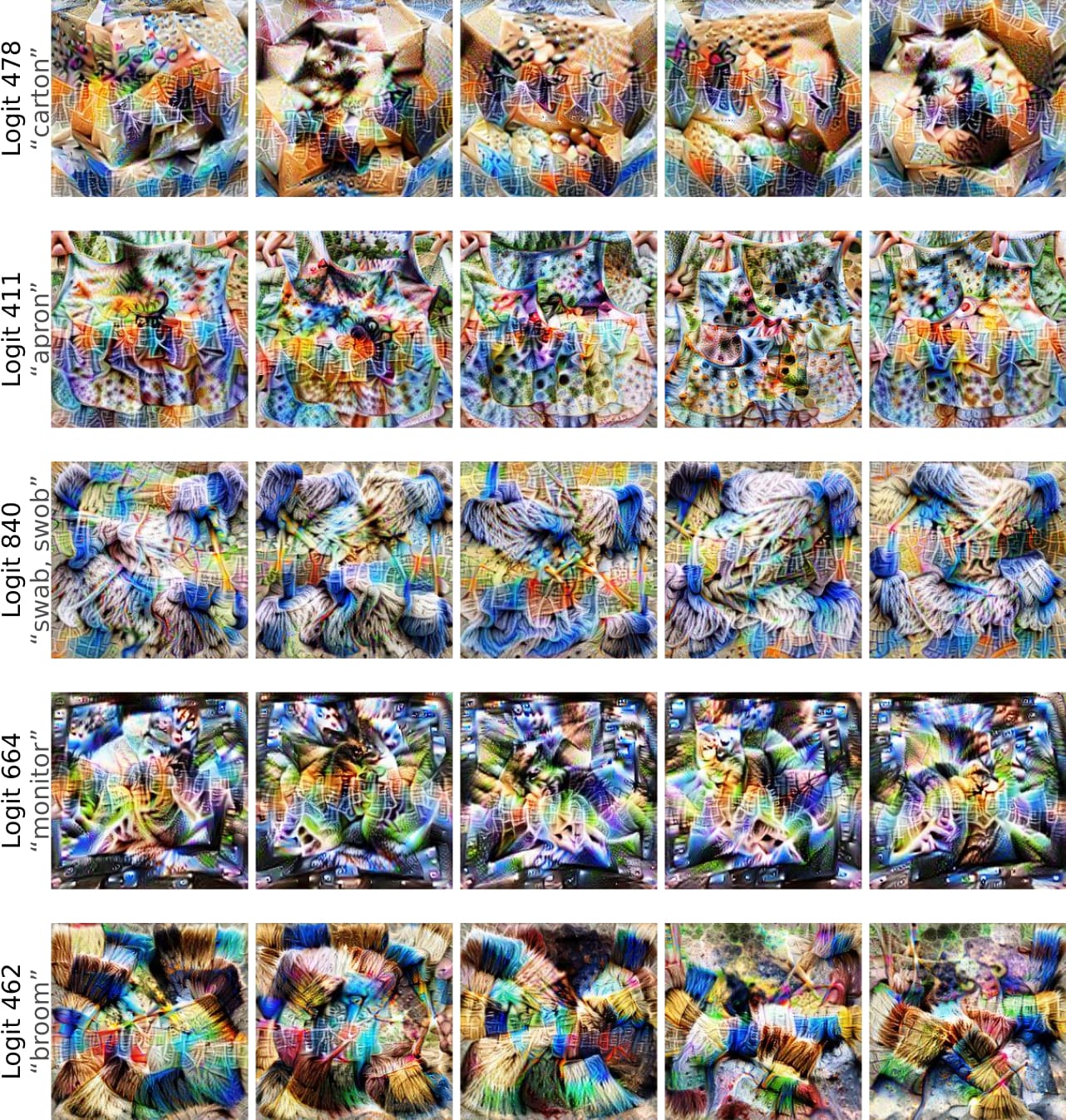}}
\caption{\protect\textbf{s-AMS for several ResNet18 logits.} Figure shows s-AMS for the output logit representations of ResNet18. Similar to the reported neurons from the average pooling layer, the logits display logographic patterns, logographic patterns specific to Chinese character detectors, suggesting that these classes may be particularly affected by CH behavior.}
\label{fig:appendix:resnet18_probing}
\end{figure}

%% file: main.bbl
\begin{thebibliography}{143}
\providecommand{\natexlab}[1]{#1}
\providecommand{\url}[1]{\texttt{#1}}
\expandafter\ifx\csname urlstyle\endcsname\relax
  \providecommand{\doi}[1]{doi: #1}\else
  \providecommand{\doi}{doi: \begingroup \urlstyle{rm}\Url}\fi

\bibitem[git(2020)]{githubGitHubWeiaicunzaipytorchcifar100}
{G}it{H}ub - weiaicunzai/pytorch-cifar100: {P}ractice on {CIFAR}100---
  github.com.
\newblock \url{https://github.com/weiaicunzai/pytorch-cifar100}, 2020.
\newblock [Accessed 08-Jan-2023].

\bibitem[Adebayo et~al.(2022)Adebayo, Muelly, Abelson, and
  Kim]{adebayo2022post}
J.~Adebayo, M.~Muelly, H.~Abelson, and B.~Kim.
\newblock Post hoc explanations may be ineffective for detecting unknown
  spurious correlation.
\newblock In \emph{International Conference on Learning Representations}, 2022.

\bibitem[Agarwal et~al.(2021)Agarwal, Krueger, Clark, Radford, Kim, and
  Brundage]{agarwal2021evaluating}
S.~Agarwal, G.~Krueger, J.~Clark, A.~Radford, J.~W. Kim, and M.~Brundage.
\newblock Evaluating {CLIP}: towards characterization of broader capabilities
  and downstream implications.
\newblock \emph{arXiv preprint arXiv:2108.02818}, 2021.

\bibitem[Anders et~al.(2022)Anders, Weber, Neumann, Samek, M{\"u}ller, and
  Lapuschkin]{anders2022finding}
C.~J. Anders, L.~Weber, D.~Neumann, W.~Samek, K.-R. M{\"u}ller, and
  S.~Lapuschkin.
\newblock Finding and removing clever hans: Using explanation methods to debug
  and improve deep models.
\newblock \emph{Information Fusion}, 77:\penalty0 261--295, 2022.

\bibitem[Bach et~al.(2015)Bach, Binder, on, Klauschen, M{\"u}ller, and
  Samek]{bach2015pixel}
S.~Bach, A.~Binder, G.~on, F.~Klauschen, K.-R. M{\"u}ller, and W.~Samek.
\newblock On pixel-wise explanations for non-linear classifier decisions by
  layer-wise relevance propagation.
\newblock \emph{PloS one}, 10\penalty0 (7):\penalty0 e0130140, 2015.

\bibitem[Baehrens et~al.(2010)Baehrens, Schroeter, Harmeling, Kawanabe, Hansen,
  and M\"uller]{baehrens2010explain}
D.~Baehrens, T.~Schroeter, S.~Harmeling, M.~Kawanabe, K.~Hansen, and K.-R.
  M\"uller.
\newblock How to explain individual classification decisions.
\newblock \emph{Journal of Machine Learning Research}, 11\penalty0
  (Jun):\penalty0 1803--1831, 2010.

\bibitem[Bao et~al.(2021)Bao, Dong, and Wei]{bao2021beit}
H.~Bao, L.~Dong, and F.~Wei.
\newblock {BEIT}: {BERT} pre-training of image transformers.
\newblock \emph{arXiv preprint arXiv:2106.08254}, 2021.

\bibitem[Bau et~al.(2017)Bau, Zhou, Khosla, Oliva, and
  Torralba]{bau2017network}
D.~Bau, B.~Zhou, A.~Khosla, A.~Oliva, and A.~Torralba.
\newblock Network dissection: Quantifying interpretability of deep visual
  representations.
\newblock In \emph{Proceedings of the IEEE Conference on Computer Vision and
  Pattern Recognition}, pages 6541--6549, 2017.

\bibitem[Bau et~al.(2018)Bau, Zhu, Strobelt, Zhou, Tenenbaum, Freeman, and
  Torralba]{bau2018gan}
D.~Bau, J.-Y. Zhu, H.~Strobelt, B.~Zhou, J.~B. Tenenbaum, W.~T. Freeman, and
  A.~Torralba.
\newblock {GAN} dissection: Visualizing and understanding generative
  adversarial networks.
\newblock \emph{arXiv preprint arXiv:1811.10597}, 2018.

\bibitem[Becht et~al.(2019)Becht, McInnes, Healy, Dutertre, Kwok, Ng, Ginhoux,
  and Newell]{becht2019dimensionality}
E.~Becht, L.~McInnes, J.~Healy, C.-A. Dutertre, I.~W. Kwok, L.~G. Ng,
  F.~Ginhoux, and E.~W. Newell.
\newblock Dimensionality reduction for visualizing single-cell data using umap.
\newblock \emph{Nature biotechnology}, 37\penalty0 (1):\penalty0 38--44, 2019.

\bibitem[Bengio et~al.(2013)Bengio, Courville, and
  Vincent]{bengio2013representation}
Y.~Bengio, A.~Courville, and P.~Vincent.
\newblock Representation learning: A review and new perspectives.
\newblock \emph{IEEE transactions on pattern analysis and machine
  intelligence}, 35\penalty0 (8):\penalty0 1798--1828, 2013.

\bibitem[Bie{\ss}mann et~al.(2010)Bie{\ss}mann, Meinecke, Gretton, Rauch,
  Rainer, Logothetis, and M{\"u}ller]{biessmann2010temporal}
F.~Bie{\ss}mann, F.~C. Meinecke, A.~Gretton, A.~Rauch, G.~Rainer, N.~K.
  Logothetis, and K.-R. M{\"u}ller.
\newblock Temporal kernel {CCA} and its application in multimodal neuronal data
  analysis.
\newblock \emph{Machine Learning}, 79\penalty0 (1):\penalty0 5--27, 2010.

\bibitem[Binder et~al.(2012)Binder, M{\"u}ller, and
  Kawanabe]{binder2012taxonomies}
A.~Binder, K.-R. M{\"u}ller, and M.~Kawanabe.
\newblock On taxonomies for multi-class image categorization.
\newblock \emph{International Journal of Computer Vision}, 99\penalty0
  (3):\penalty0 281--301, 2012.

\bibitem[Binder et~al.(2023)Binder, Weber, Lapuschkin, Montavon, M{\"u}ller,
  and Samek]{binder2023shortcomings}
A.~Binder, L.~Weber, S.~Lapuschkin, G.~Montavon, K.-R. M{\"u}ller, and
  W.~Samek.
\newblock Shortcomings of top-down randomization-based sanity checks for
  evaluations of deep neural network explanations.
\newblock In \emph{Proceedings of the IEEE/CVF Conference on Computer Vision
  and Pattern Recognition}, pages 16143--16152, 2023.

\bibitem[Bird et~al.(2009)Bird, Klein, and Loper]{bird2009natural}
S.~Bird, E.~Klein, and E.~Loper.
\newblock \emph{Natural language processing with Python: analyzing text with
  the natural language toolkit}.
\newblock O'Reilly Media, Inc., 2009.

\bibitem[Bisong and Bisong(2019)]{bisong2019google}
E.~Bisong and E.~Bisong.
\newblock Google colaboratory.
\newblock \emph{Building machine learning and deep learning models on google
  cloud platform: a comprehensive guide for beginners}, pages 59--64, 2019.

\bibitem[Bommasani et~al.(2021)Bommasani, Hudson, Adeli, Altman, Arora, von
  Arx, Bernstein, Bohg, Bosselut, Brunskill,
  et~al.]{bommasani2021opportunities}
R.~Bommasani, D.~A. Hudson, E.~Adeli, R.~Altman, S.~Arora, S.~von Arx, M.~S.
  Bernstein, J.~Bohg, A.~Bosselut, E.~Brunskill, et~al.
\newblock On the opportunities and risks of foundation models.
\newblock \emph{arXiv preprint arXiv:2108.07258}, 2021.

\bibitem[Borowski et~al.(2020)Borowski, Zimmermann, Schepers, Geirhos, Wallis,
  Bethge, and Brendel]{borowski2020natural}
J.~Borowski, R.~S. Zimmermann, J.~Schepers, R.~Geirhos, T.~S. Wallis,
  M.~Bethge, and W.~Brendel.
\newblock Natural images are more informative for interpreting cnn activations
  than state-of-the-art synthetic feature visualizations.
\newblock In \emph{NeurIPS 2020 Workshop SVRHM}, 2020.

\bibitem[Braun et~al.(2008)Braun, Buhmann, and M{\"u}ller]{braun2008relevant}
M.~L. Braun, J.~M. Buhmann, and K.-R. M{\"u}ller.
\newblock On relevant dimensions in kernel feature spaces.
\newblock \emph{The Journal of Machine Learning Research}, 9:\penalty0
  1875--1908, 2008.

\bibitem[Breunig et~al.(2000)Breunig, Kriegel, Ng, and Sander]{breunig2000lof}
M.~M. Breunig, H.-P. Kriegel, R.~T. Ng, and J.~Sander.
\newblock {LOF}: identifying density-based local outliers.
\newblock In \emph{Proceedings of the 2000 ACM SIGMOD International Conference
  on Management of data}, pages 93--104, 2000.

\bibitem[Brown and Talbert(2022)]{brown2022using}
K.~E. Brown and D.~A. Talbert.
\newblock Using explainable {AI} to measure feature contribution to
  uncertainty.
\newblock In \emph{The International FLAIRS Conference Proceedings}, volume~35,
  2022.

\bibitem[Brown et~al.(2020)Brown, Mann, Ryder, Subbiah, Kaplan, Dhariwal,
  Neelakantan, Shyam, Sastry, Askell, et~al.]{brown2020language}
T.~Brown, B.~Mann, N.~Ryder, M.~Subbiah, J.~D. Kaplan, P.~Dhariwal,
  A.~Neelakantan, P.~Shyam, G.~Sastry, A.~Askell, et~al.
\newblock Language models are few-shot learners.
\newblock \emph{Advances in Neural Information Processing Systems},
  33:\penalty0 1877--1901, 2020.

\bibitem[Brust and Denzler(2019)]{brust2019not}
C.-A. Brust and J.~Denzler.
\newblock Not just a matter of semantics: The relationship between visual and
  semantic similarity.
\newblock In \emph{German Conference on Pattern Recognition}, pages 414--427.
  Springer, 2019.

\bibitem[Buhrmester et~al.(2019)Buhrmester, M{\"u}nch, and
  Arens]{buhrmester2019analysis}
V.~Buhrmester, D.~M{\"u}nch, and M.~Arens.
\newblock Analysis of explainers of black box {D}eep {N}eural {N}etworks for
  {C}omputer {V}ision: A survey.
\newblock \emph{arXiv preprint arXiv:1911.12116}, 2019.

\bibitem[Bykov et~al.(2021)Bykov, H{\"o}hne, Creosteanu, M{\"u}ller, Klauschen,
  Nakajima, and Kloft]{bykov2021explaining}
K.~Bykov, M.~M.-C. H{\"o}hne, A.~Creosteanu, K.-R. M{\"u}ller, F.~Klauschen,
  S.~Nakajima, and M.~Kloft.
\newblock Explaining {B}ayesian {N}eural {N}etworks.
\newblock \emph{arXiv preprint arXiv:2108.10346}, 2021.

\bibitem[Bykov et~al.(2022)Bykov, Hedstr{\"o}m, Nakajima, and
  H{\"o}hne]{bykov2021noisegrad}
K.~Bykov, A.~Hedstr{\"o}m, S.~Nakajima, and M.~M.-C. H{\"o}hne.
\newblock Noise{G}rad—enhancing explanations by introducing stochasticity to
  model weights.
\newblock In \emph{Proceedings of the AAAI Conference on Artificial
  Intelligence}, volume~36, pages 6132--6140, 2022.

\bibitem[Cammarata et~al.(2020)Cammarata, Goh, Carter, Schubert, Petrov, and
  Olah]{cammarata2020curve}
N.~Cammarata, G.~Goh, S.~Carter, L.~Schubert, M.~Petrov, and C.~Olah.
\newblock Curve detectors.
\newblock \emph{Distill}, 5\penalty0 (6):\penalty0 e00024--003, 2020.

\bibitem[Carter et~al.(2019)Carter, Armstrong, Schubert, Johnson, and
  Olah]{carter2019exploring}
S.~Carter, Z.~Armstrong, L.~Schubert, I.~Johnson, and C.~Olah.
\newblock Exploring {N}eural {N}etworks with activation atlases.
\newblock \emph{Distill.}, 2019.

\bibitem[Chen et~al.(2018)Chen, Li, Tao, Barnett, Su, and Rudin]{chen2018looks}
C.~Chen, O.~Li, C.~Tao, A.~J. Barnett, J.~Su, and C.~Rudin.
\newblock This looks like that: deep learning for interpretable image
  recognition.
\newblock \emph{arXiv preprint arXiv:1806.10574}, 2018.

\bibitem[Da(2004)]{da2004corpus}
J.~Da.
\newblock A corpus-based study of character and bigram frequencies in chinese
  e-texts and its implications for chinese language instruction.
\newblock In \emph{Proceedings of the fourth International Conference on new
  technologies in teaching and learning Chinese}, pages 501--511. Citeseer,
  2004.

\bibitem[Dai and Lin(2017)]{dai2017contrastive}
B.~Dai and D.~Lin.
\newblock Contrastive learning for image captioning.
\newblock \emph{Advances in Neural Information Processing Systems}, 30, 2017.

\bibitem[Deng et~al.(2009)Deng, Dong, Socher, Li, Li, and
  Fei-Fei]{deng2009imagenet}
J.~Deng, W.~Dong, R.~Socher, L.-J. Li, K.~Li, and L.~Fei-Fei.
\newblock Imagenet: A large-scale hierarchical image database.
\newblock In \emph{2009 IEEE Conference on Computer Vision and Pattern
  Recognition}, pages 248--255. IEEE, 2009.

\bibitem[Deng(2012)]{deng2012mnist}
L.~Deng.
\newblock The {MNIST} database of handwritten digit images for machine learning
  research.
\newblock \emph{IEEE Signal Processing Magazine}, 29\penalty0 (6):\penalty0
  141--142, 2012.

\bibitem[Deselaers and Ferrari(2011)]{deselaers2011visual}
T.~Deselaers and V.~Ferrari.
\newblock Visual and semantic similarity in imagenet.
\newblock In \emph{CVPR 2011}, pages 1777--1784. IEEE, 2011.

\bibitem[Dosovitskiy et~al.(2020)Dosovitskiy, Beyer, Kolesnikov, Weissenborn,
  Zhai, Unterthiner, Dehghani, Minderer, Heigold, Gelly,
  et~al.]{dosovitskiy2020image}
A.~Dosovitskiy, L.~Beyer, A.~Kolesnikov, D.~Weissenborn, X.~Zhai,
  T.~Unterthiner, M.~Dehghani, M.~Minderer, G.~Heigold, S.~Gelly, et~al.
\newblock An image is worth 16x16 words: Transformers for image recognition at
  scale.
\newblock \emph{arXiv preprint arXiv:2010.11929}, 2020.

\bibitem[Erhan et~al.(2009)Erhan, Bengio, Courville, and
  Vincent]{erhan2009visualizing}
D.~Erhan, Y.~Bengio, A.~Courville, and P.~Vincent.
\newblock Visualizing higher-layer features of a deep network.
\newblock \emph{University of Montreal}, 1341\penalty0 (3):\penalty0 1, 2009.

\bibitem[Fogel and Sagi(1989)]{fogel1989gabor}
I.~Fogel and D.~Sagi.
\newblock Gabor filters as texture discriminator.
\newblock \emph{Biological cybernetics}, 61\penalty0 (2):\penalty0 103--113,
  1989.

\bibitem[Gabriel(2020)]{gabriel2020artificial}
I.~Gabriel.
\newblock Artificial intelligence, values, and alignment.
\newblock \emph{Minds and machines}, 30\penalty0 (3):\penalty0 411--437, 2020.

\bibitem[Gade et~al.(2019)Gade, Geyik, Kenthapadi, Mithal, and
  Taly]{gade2019explainable}
K.~Gade, S.~C. Geyik, K.~Kenthapadi, V.~Mithal, and A.~Taly.
\newblock Explainable {AI} in industry.
\newblock In \emph{Proceedings of the 25th ACM SIGKDD International Conference
  on knowledge discovery \& data mining}, pages 3203--3204, 2019.

\bibitem[Gautam et~al.(2021)Gautam, H{\"o}hne, Hansen, Jenssen, and
  Kampffmeyer]{gautam2021looks}
S.~Gautam, M.~M.-C. H{\"o}hne, S.~Hansen, R.~Jenssen, and M.~Kampffmeyer.
\newblock This looks more like that: Enhancing self-explaining models by
  prototypical relevance propagation.
\newblock \emph{arXiv preprint arXiv:2108.12204}, 2021.

\bibitem[Gautam et~al.(2022{\natexlab{a}})Gautam, Boubekki, Hansen, Salahuddin,
  Jenssen, H{\"o}hne, and Kampffmeyer]{gautam2022protovae}
S.~Gautam, A.~Boubekki, S.~Hansen, S.~Salahuddin, R.~Jenssen, M.~H{\"o}hne, and
  M.~Kampffmeyer.
\newblock Protovae: A trustworthy self-explainable prototypical variational
  model.
\newblock \emph{Advances in Neural Information Processing Systems},
  35:\penalty0 17940--17952, 2022{\natexlab{a}}.

\bibitem[Gautam et~al.(2022{\natexlab{b}})Gautam, Höhne, Hansen, Jenssen, and
  Kampffmeyer]{sg_isbi}
S.~Gautam, M.~M.-C. Höhne, S.~Hansen, R.~Jenssen, and M.~Kampffmeyer.
\newblock Demonstrating the risk of imbalanced datasets in chest x-ray
  image-based diagnostics by prototypical relevance propagation.
\newblock In \emph{2022 IEEE 19th International Symposium on Biomedical Imaging
  (ISBI)}, pages 1--5, 2022{\natexlab{b}}.
\newblock \doi{10.1109/ISBI52829.2022.9761651}.

\bibitem[Geirhos et~al.(2018)Geirhos, Rubisch, Michaelis, Bethge, Wichmann, and
  Brendel]{geirhos2018imagenet}
R.~Geirhos, P.~Rubisch, C.~Michaelis, M.~Bethge, F.~A. Wichmann, and
  W.~Brendel.
\newblock Image{N}et-trained {CNNs} are biased towards texture; increasing
  shape bias improves accuracy and robustness.
\newblock \emph{arXiv preprint arXiv:1811.12231}, 2018.

\bibitem[Geirhos et~al.(2020)Geirhos, Jacobsen, Michaelis, Zemel, Brendel,
  Bethge, and Wichmann]{geirhos2020shortcut}
R.~Geirhos, J.-H. Jacobsen, C.~Michaelis, R.~Zemel, W.~Brendel, M.~Bethge, and
  F.~A. Wichmann.
\newblock Shortcut learning in deep neural networks.
\newblock \emph{Nature Machine Intelligence}, 2\penalty0 (11):\penalty0
  665--673, 2020.

\bibitem[Glorot et~al.(2011)Glorot, Bordes, and Bengio]{glorot2011deep}
X.~Glorot, A.~Bordes, and Y.~Bengio.
\newblock Deep sparse rectifier neural networks.
\newblock In \emph{Proceedings of the fourteenth international conference on
  {A}rtificial {I}ntelligence and {S}tatistics}, pages 315--323. JMLR Workshop
  and Conference Proceedings, 2011.

\bibitem[Goh et~al.(2021)Goh, Cammarata, Voss, Carter, Petrov, Schubert,
  Radford, and Olah]{goh2021multimodal}
G.~Goh, N.~Cammarata, C.~Voss, S.~Carter, M.~Petrov, L.~Schubert, A.~Radford,
  and C.~Olah.
\newblock Multimodal neurons in artificial neural networks.
\newblock \emph{Distill}, 6\penalty0 (3):\penalty0 e30, 2021.

\bibitem[Grinwald et~al.(2022)Grinwald, Bykov, Nakajima, and
  H{\"o}hne]{Grinwald2022}
D.~Grinwald, K.~Bykov, S.~Nakajima, and M.~M.-C. H{\"o}hne.
\newblock Visualizing the diversity of representations learned by bayesian
  neural networks.
\newblock \emph{arXiv preprint arXiv:2201.10859}, 2022.

\bibitem[Gu et~al.(2017)Gu, Dolan-Gavitt, and Garg]{gu2017badnets}
T.~Gu, B.~Dolan-Gavitt, and S.~Garg.
\newblock Badnets: Identifying vulnerabilities in the machine learning model
  supply chain.
\newblock \emph{arXiv preprint arXiv:1708.06733}, 2017.

\bibitem[Guidotti(2021)]{guidotti2021evaluating}
R.~Guidotti.
\newblock Evaluating local explanation methods on ground truth.
\newblock \emph{Artificial Intelligence}, 291:\penalty0 103428, 2021.

\bibitem[Guidotti et~al.(2018)Guidotti, Monreale, Ruggieri, Turini, Giannotti,
  and Pedreschi]{guidotti2018survey}
R.~Guidotti, A.~Monreale, S.~Ruggieri, F.~Turini, F.~Giannotti, and
  D.~Pedreschi.
\newblock A survey of methods for explaining black box models.
\newblock \emph{ACM computing surveys (CSUR)}, 51\penalty0 (5):\penalty0 1--42,
  2018.

\bibitem[Guyon and Elisseeff(2003)]{guyon2003introduction}
I.~Guyon and A.~Elisseeff.
\newblock An introduction to variable and feature selection.
\newblock \emph{Journal of Machine Learning Research}, 3\penalty0
  (Mar):\penalty0 1157--1182, 2003.

\bibitem[Hardoon et~al.(2005)Hardoon, Szedmak, and Shawe-Taylor]{Hardoon2005}
D.~Hardoon, S.~Szedmak, and J.~Shawe-Taylor.
\newblock {C}anonical {C}orrelation {A}nalysis: An overview with application to
  learning methods.
\newblock \emph{Neural computation}, 16:\penalty0 2639--64, 01 2005.
\newblock \doi{10.1162/0899766042321814}.

\bibitem[He et~al.(2016)He, Zhang, Ren, and Sun]{he2016deep}
K.~He, X.~Zhang, S.~Ren, and J.~Sun.
\newblock Deep residual learning for image recognition.
\newblock In \emph{Proceedings of the IEEE Conference on Computer Vision and
  Pattern Recognition}, pages 770--778, 2016.

\bibitem[Hedstr{\"o}m et~al.(2022)Hedstr{\"o}m, Weber, Bareeva, Motzkus, Samek,
  Lapuschkin, and H{\"o}hne]{hedstrom2022quantus}
A.~Hedstr{\"o}m, L.~Weber, D.~Bareeva, F.~Motzkus, W.~Samek, S.~Lapuschkin, and
  M.~M.-C. H{\"o}hne.
\newblock Quantus: an explainable {AI} toolkit for responsible evaluation of
  neural network explanations.
\newblock \emph{arXiv preprint arXiv:2202.06861}, 2022.

\bibitem[Hernandez et~al.(2021)Hernandez, Schwettmann, Bau, Bagashvili,
  Torralba, and Andreas]{hernandez2021natural}
E.~Hernandez, S.~Schwettmann, D.~Bau, T.~Bagashvili, A.~Torralba, and
  J.~Andreas.
\newblock Natural language descriptions of deep visual features.
\newblock In \emph{International Conference on Learning Representations}, 2021.

\bibitem[Hjelm et~al.(2018)Hjelm, Fedorov, Lavoie-Marchildon, Grewal, Bachman,
  Trischler, and Bengio]{hjelm2018learning}
R.~D. Hjelm, A.~Fedorov, S.~Lavoie-Marchildon, K.~Grewal, P.~Bachman,
  A.~Trischler, and Y.~Bengio.
\newblock Learning deep representations by mutual information estimation and
  maximization.
\newblock \emph{arXiv preprint arXiv:1808.06670}, 2018.

\bibitem[Huang et~al.(2017)Huang, Liu, Van Der~Maaten, and
  Weinberger]{huang2017densely}
G.~Huang, Z.~Liu, L.~Van Der~Maaten, and K.~Q. Weinberger.
\newblock Densely {C}onnected {C}onvolutional {N}etworks.
\newblock In \emph{Proceedings of the IEEE Conference on Computer Vision and
  Pattern Recognition}, pages 4700--4708, 2017.

\bibitem[Iandola et~al.(2016)Iandola, Han, Moskewicz, Ashraf, Dally, and
  Keutzer]{iandola2016squeezenet}
F.~N. Iandola, S.~Han, M.~W. Moskewicz, K.~Ashraf, W.~J. Dally, and K.~Keutzer.
\newblock Squeezenet: Alexnet-level accuracy with 50x fewer parameters and< 0.5
  mb model size.
\newblock \emph{arXiv preprint arXiv:1602.07360}, 2016.

\bibitem[Irvin et~al.(2019)Irvin, Rajpurkar, Ko, Yu, Ciurea-Ilcus, Chute,
  Marklund, Haghgoo, Ball, Shpanskaya, Seekins, Mong, Halabi, Sandberg, Jones,
  Larson, Langlotz, Patel, Lungren, and Ng]{irvin2019chexpert}
J.~Irvin, P.~Rajpurkar, M.~Ko, Y.~Yu, S.~Ciurea-Ilcus, C.~Chute, H.~Marklund,
  B.~Haghgoo, R.~Ball, K.~Shpanskaya, J.~Seekins, D.~Mong, S.~Halabi,
  J.~Sandberg, R.~Jones, D.~Larson, C.~Langlotz, B.~Patel, M.~Lungren, and
  A.~Ng.
\newblock Chexpert: A large chest radiograph dataset with uncertainty labels
  and expert comparison.
\newblock \emph{Proceedings of the AAAI Conference on Artificial Intelligence},
  33:\penalty0 590--597, 07 2019.

\bibitem[Izmailov et~al.(2022)Izmailov, Kirichenko, Gruver, and
  Wilson]{izmailov2022spurious}
P.~Izmailov, P.~Kirichenko, N.~Gruver, and A.~G. Wilson.
\newblock On feature learning in the presence of spurious correlations.
\newblock \emph{arXiv preprint arXiv:2210.11369}, 2022.

\bibitem[Jackson()]{jackson1986introduction}
P.~Jackson.
\newblock Introduction to expert systems.
\newblock URL \url{https://www.osti.gov/biblio/5675197}.
\newblock [Accessed 16-Feb-2023].

\bibitem[Jaiswal et~al.(2020)Jaiswal, Babu, Zadeh, Banerjee, and
  Makedon]{jaiswal2020survey}
A.~Jaiswal, A.~R. Babu, M.~Z. Zadeh, D.~Banerjee, and F.~Makedon.
\newblock A survey on contrastive self-supervised learning.
\newblock \emph{Technologies}, 9\penalty0 (1):\penalty0 2, 2020.

\bibitem[Jiang and Nachum(2020)]{pmlr-v108-jiang20a}
H.~Jiang and O.~Nachum.
\newblock Identifying and correcting label bias in machine learning.
\newblock In S.~Chiappa and R.~Calandra, editors, \emph{Proceedings of the
  Twenty Third International Conference on Artificial Intelligence and
  Statistics}, volume 108 of \emph{Proceedings of Machine Learning Research},
  pages 702--712. PMLR, 26--28 Aug 2020.
\newblock URL \url{https://proceedings.mlr.press/v108/jiang20a.html}.

\bibitem[Jolliffe and Cadima(2016)]{jolliffe2016principal}
I.~T. Jolliffe and J.~Cadima.
\newblock Principal component analysis: a review and recent developments.
\newblock \emph{Philosophical transactions of the royal society A:
  Mathematical, Physical and Engineering Sciences}, 374\penalty0
  (2065):\penalty0 20150202, 2016.

\bibitem[Kornblith et~al.(2019)Kornblith, Norouzi, Lee, and
  Hinton]{kornblith2019}
S.~Kornblith, M.~Norouzi, H.~Lee, and G.~Hinton.
\newblock Similarity of neural network representations revisited.
\newblock In \emph{International Conference on Machine Learning}, pages
  3519--3529. PMLR, 2019.

\bibitem[Kriegel et~al.(2008)Kriegel, Schubert, and Zimek]{kriegel2008angle}
H.-P. Kriegel, M.~Schubert, and A.~Zimek.
\newblock Angle-based outlier detection in high-dimensional data.
\newblock In \emph{Proceedings of the 14th ACM SIGKDD International Conference
  on Knowledge discovery and data mining}, pages 444--452, 2008.

\bibitem[Krizhevsky(2009)]{krizhevsky2009learning}
A.~Krizhevsky.
\newblock Learning multiple layers of features from tiny images.
\newblock pages 32--33, 2009.
\newblock URL
  \url{https://www.cs.toronto.edu/~kriz/learning-features-2009-TR.pdf}.

\bibitem[Krizhevsky et~al.(2017)Krizhevsky, Sutskever, and
  Hinton]{krizhevsky2017imagenet}
A.~Krizhevsky, I.~Sutskever, and G.~E. Hinton.
\newblock Imagenet classification with deep convolutional neural networks.
\newblock \emph{Communications of the ACM}, 60\penalty0 (6):\penalty0 84--90,
  2017.

\bibitem[Laakso(2000)]{laakso2000}
A.~Laakso.
\newblock Content and cluster analysis: Assessing representational similarity
  in neural systems.
\newblock \emph{Philosophical Psychology}, 13, 05 2000.
\newblock \doi{10.1080/09515080050002726}.

\bibitem[Lapuschkin et~al.(2016)Lapuschkin, Binder, Montavon, Muller, and
  Samek]{lapuschkin2016analyzing}
S.~Lapuschkin, A.~Binder, G.~Montavon, K.-R. Muller, and W.~Samek.
\newblock Analyzing classifiers: Fisher vectors and deep neural networks.
\newblock In \emph{Proceedings of the IEEE Conference on Computer Vision and
  Pattern Recognition}, pages 2912--2920, 2016.

\bibitem[Lapuschkin et~al.(2019)Lapuschkin, W{\"a}ldchen, Binder, Montavon,
  Samek, and M{\"u}ller]{lapuschkin2019unmasking}
S.~Lapuschkin, S.~W{\"a}ldchen, A.~Binder, G.~Montavon, W.~Samek, and K.-R.
  M{\"u}ller.
\newblock Unmasking clever hans predictors and assessing what machines really
  learn.
\newblock \emph{Nature communications}, 10:\penalty0 1096, 2019.

\bibitem[Lazarevic and Kumar(2005)]{lazarevic2005feature}
A.~Lazarevic and V.~Kumar.
\newblock Feature bagging for outlier detection.
\newblock In \emph{Proceedings of the eleventh ACM SIGKDD International
  Conference on Knowledge discovery in data mining}, pages 157--166, 2005.

\bibitem[Le and Kayal(2021)]{le2021revisiting}
M.~Le and S.~Kayal.
\newblock Revisiting edge detection in convolutional neural networks.
\newblock In \emph{2021 International Joint Conference on Neural Networks
  (IJCNN)}, pages 1--9. IEEE, 2021.

\bibitem[Le and Yang(2015)]{le2015tiny}
Y.~Le and X.~Yang.
\newblock Tiny imagenet visual recognition challenge.
\newblock \emph{Stanford CS 231N}, 7\penalty0 (7):\penalty0 3, 2015.

\bibitem[Leacock and Chodorow(1998)]{leacock1998combining}
C.~Leacock and M.~Chodorow.
\newblock Combining local context and wordnet similarity for word sense
  identification.
\newblock \emph{WordNet: An electronic lexical database}, 49\penalty0
  (2):\penalty0 265--283, 1998.

\bibitem[LeCun and Misra(2021)]{lecun2021self}
Y.~LeCun and I.~Misra.
\newblock Self-supervised learning: The dark matter of intelligence, 2021.
\newblock URL
  \url{https://ai.facebook.com/blog/self-supervised-learning-the-dark-matter-of-intelligence/}.
\newblock [Accessed 08-Jan-2023].

\bibitem[Li et~al.(2015)Li, Yosinski, Clune, Lipson, and Hopcroft]{Li2015}
Y.~Li, J.~Yosinski, J.~Clune, H.~Lipson, and J.~Hopcroft.
\newblock Convergent learning: Do different neural networks learn the same
  representations?
\newblock \emph{arXiv preprint arXiv:1511.07543}, 2015.

\bibitem[Li et~al.(2022)Li, Evtimov, Gordo, Hazirbas, Hassner, Ferrer, Xu, and
  Ibrahim]{li2022Dilemma}
Z.~Li, I.~Evtimov, A.~Gordo, C.~Hazirbas, T.~Hassner, C.~C. Ferrer, C.~Xu, and
  M.~Ibrahim.
\newblock A whac-a-mole dilemma: Shortcuts come in multiples where mitigating
  one amplifies others, 2022.

\bibitem[Liu et~al.(2008)Liu, Ting, and Zhou]{liu2008isolation}
F.~T. Liu, K.~M. Ting, and Z.-H. Zhou.
\newblock Isolation {F}orest.
\newblock In \emph{2008 8-th IEEE International Conference on Data Mining},
  pages 413--422. IEEE, 2008.

\bibitem[Ma et~al.(2018)Ma, Zhang, Zheng, and Sun]{ma2018shufflenet}
N.~Ma, X.~Zhang, H.-T. Zheng, and J.~Sun.
\newblock Shufflenet v2: Practical guidelines for efficient cnn architecture
  design.
\newblock In \emph{Proceedings of the European Conference on Computer Vision
  (ECCV)}, pages 116--131, 2018.

\bibitem[Mantel(1967)]{mentel1967detection}
N.~Mantel.
\newblock The detection of disease clustering and a generalized regression
  approach.
\newblock \emph{Cancer Res.}, 27:\penalty0 175--178, 1967.

\bibitem[Marcel and Rodriguez(2010)]{marcel2010torchvision}
S.~Marcel and Y.~Rodriguez.
\newblock Torchvision the machine-vision package of torch.
\newblock In \emph{Proceedings of the 18th ACM International Conference on
  Multimedia}, pages 1485--1488, 2010.

\bibitem[Marr and Nishihara(1978)]{marr1978representation}
D.~Marr and H.~K. Nishihara.
\newblock Representation and recognition of the spatial organization of
  three-dimensional shapes.
\newblock \emph{Proceedings of the Royal Society of London. Series B.
  Biological Sciences}, 200\penalty0 (1140):\penalty0 269--294, 1978.

\bibitem[McInnes et~al.(2018)McInnes, Healy, Saul, and
  Grossberger]{McInnes2018UMAP}
L.~McInnes, J.~Healy, N.~Saul, and L.~Grossberger.
\newblock Umap: Uniform manifold approximation and projection.
\newblock \emph{Journal of Open Source Software}, 3:\penalty0 861, 09 2018.
\newblock \doi{10.21105/joss.00861}.

\bibitem[Mikolov et~al.(2013)Mikolov, Chen, Corrado, and
  Dean]{mikolov2013efficient}
T.~Mikolov, K.~Chen, G.~Corrado, and J.~Dean.
\newblock Efficient estimation of word representations in vector space.
\newblock \emph{arXiv preprint arXiv:1301.3781}, 2013.

\bibitem[Miller(1995)]{miller1995wordnet}
G.~A. Miller.
\newblock Wordnet: a lexical database for english.
\newblock \emph{Communications of the ACM}, 38\penalty0 (11):\penalty0 39--41,
  1995.

\bibitem[Montavon et~al.(2011)Montavon, Braun, and
  M{\"u}ller]{montavon2011kernel}
G.~Montavon, M.~L. Braun, and K.-R. M{\"u}ller.
\newblock Kernel analysis of deep networks.
\newblock \emph{Journal of Machine Learning Research}, 12\penalty0 (9), 2011.

\bibitem[Montavon et~al.(2018)Montavon, Samek, and
  M{\"u}ller]{montavon2018methods}
G.~Montavon, W.~Samek, and K.-R. M{\"u}ller.
\newblock Methods for interpreting and understanding deep neural networks.
\newblock \emph{Digital Signal Processing}, 73:\penalty0 1--15, 2018.

\bibitem[Morcos et~al.(2018)Morcos, Raghu, and Bengio]{Morcos2018}
A.~Morcos, M.~Raghu, and S.~Bengio.
\newblock Insights on representational similarity in neural networks with
  canonical correlation.
\newblock \emph{Advances in Neural Information Processing Systems}, 31, 2018.

\bibitem[Mordvintsev et~al.(2018)Mordvintsev, Pezzotti, Schubert, and
  Olah]{mordvintsev2018differentiable}
A.~Mordvintsev, N.~Pezzotti, L.~Schubert, and C.~Olah.
\newblock Differentiable image parameterizations.
\newblock \emph{Distill}, 3\penalty0 (7):\penalty0 e12, 2018.

\bibitem[Mu and Andreas(2020)]{mu2020compositional}
J.~Mu and J.~Andreas.
\newblock Compositional explanations of neurons.
\newblock \emph{Advances in Neural Information Processing Systems},
  33:\penalty0 17153--17163, 2020.

\bibitem[Muttenthaler et~al.(2022)Muttenthaler, Dippel, Linhardt, Vandermeulen,
  and Kornblith]{muttenthaler2022human}
L.~Muttenthaler, J.~Dippel, L.~Linhardt, R.~A. Vandermeulen, and S.~Kornblith.
\newblock Human alignment of neural network representations.
\newblock \emph{arXiv preprint arXiv:2211.01201}, 2022.

\bibitem[Nguyen et~al.(2016)Nguyen, Dosovitskiy, Yosinski, Brox, and
  Clune]{nguyen2016synthesizing}
A.~Nguyen, A.~Dosovitskiy, J.~Yosinski, T.~Brox, and J.~Clune.
\newblock Synthesizing the preferred inputs for neurons in neural networks via
  deep generator networks.
\newblock In \emph{Advances in Neural Information Processing Systems}, pages
  3387--3395, 2016.

\bibitem[Nguyen et~al.(2019)Nguyen, Yosinski, and
  Clune]{nguyen2019understanding}
A.~Nguyen, J.~Yosinski, and J.~Clune.
\newblock Understanding neural networks via feature visualization: A survey.
\newblock In \emph{Explainable {AI}: interpreting, explaining and visualizing
  deep learning}, pages 55--76. Springer, 2019.

\bibitem[Nguyen et~al.(2015)Nguyen, Yosinski, and Clune]{nguyen2015innovation}
A.~M. Nguyen, J.~Yosinski, and J.~Clune.
\newblock Innovation engines: Automated creativity and improved stochastic
  optimization via deep learning.
\newblock In \emph{Proceedings of the 2015 annual conference on genetic and
  evolutionary computation}, pages 959--966, 2015.

\bibitem[Nguyen et~al.(2020)Nguyen, Raghu, and Kornblith]{Nguyen2020}
T.~Nguyen, M.~Raghu, and S.~Kornblith.
\newblock Do wide and deep networks learn the same things? {U}ncovering how
  neural network representations vary with width and depth.
\newblock \emph{arXiv preprint arXiv:2010.15327}, 2020.

\bibitem[Nguyen et~al.(2022)Nguyen, Raghu, and Kornblith]{Nguyen2022}
T.~Nguyen, M.~Raghu, and S.~Kornblith.
\newblock On the origins of the block structure phenomenon in neural network
  representations.
\newblock \emph{arXiv preprint arXiv:2202.07184}, 2022.

\bibitem[Olah et~al.(2017)Olah, Mordvintsev, and Schubert]{olah2017feature}
C.~Olah, A.~Mordvintsev, and L.~Schubert.
\newblock Feature visualization.
\newblock \emph{Distill}, 2\penalty0 (11):\penalty0 e7, 2017.

\bibitem[Olah et~al.(2020)Olah, Cammarata, Voss, Schubert, and
  Goh]{olah2020naturally}
C.~Olah, N.~Cammarata, C.~Voss, L.~Schubert, and G.~Goh.
\newblock Naturally occurring equivariance in neural networks.
\newblock \emph{Distill}, 5\penalty0 (12):\penalty0 e00024--004, 2020.

\bibitem[Omeiza et~al.(2019)Omeiza, Speakman, Cintas, and
  Weldermariam]{omeiza2019smooth}
D.~Omeiza, S.~Speakman, C.~Cintas, and K.~Weldermariam.
\newblock Smooth {Grad-Cam}++: An enhanced inference level visualization
  technique for deep convolutional neural network models.
\newblock \emph{arXiv preprint arXiv:1908.01224}, 2019.

\bibitem[Pedersen et~al.(2004)Pedersen, Patwardhan, Michelizzi,
  et~al.]{pedersen2004wordnet}
T.~Pedersen, S.~Patwardhan, J.~Michelizzi, et~al.
\newblock Wordnet:: Similarity-measuring the relatedness of concepts.
\newblock In \emph{AAAI}, volume~4, pages 25--29, 2004.

\bibitem[Qin and Wang(2019)]{qin2019nasnet}
X.~Qin and Z.~Wang.
\newblock Nasnet: A neuron attention stage-by-stage net for single image
  deraining.
\newblock \emph{arXiv preprint arXiv:1912.03151}, 2019.

\bibitem[Radford et~al.(2021)Radford, Kim, Hallacy, Ramesh, Goh, Agarwal,
  Sastry, Askell, Mishkin, Clark, et~al.]{radford2021learning}
A.~Radford, J.~W. Kim, C.~Hallacy, A.~Ramesh, G.~Goh, S.~Agarwal, G.~Sastry,
  A.~Askell, P.~Mishkin, J.~Clark, et~al.
\newblock Learning transferable visual models from natural language
  supervision.
\newblock In \emph{International Conference on Machine Learning}, pages
  8748--8763. PMLR, 2021.

\bibitem[Raghu et~al.(2017)Raghu, Gilmer, Yosinski, and
  Sohl-Dickstein]{Raghu2017}
M.~Raghu, J.~Gilmer, J.~Yosinski, and J.~Sohl-Dickstein.
\newblock {SVCCA}: {S}ingular {V}ector {C}anonical {C}orrelation {A}nalysis for
  deep understanding and improvement.
\newblock \emph{arXiv preprint arXiv:1706.05806}, 2017.

\bibitem[Raghu et~al.(2021)Raghu, Unterthiner, Kornblith, Zhang, and
  Dosovitskiy]{Raghu2021}
M.~Raghu, T.~Unterthiner, S.~Kornblith, C.~Zhang, and A.~Dosovitskiy.
\newblock Do vision transformers see like convolutional neural networks?
\newblock \emph{Advances in Neural Information Processing Systems},
  34:\penalty0 12116--12128, 2021.

\bibitem[Ramsay et~al.(1984)Ramsay, Berge, and Styan]{Ramsay1984}
J.~Ramsay, J.~Berge, and G.~Styan.
\newblock Matrix correlation.
\newblock \emph{Psychometrika}, 49:\penalty0 403--423, 09 1984.
\newblock \doi{10.1007/BF02306029}.

\bibitem[Rombach et~al.(2022)Rombach, Blattmann, Lorenz, Esser, and
  Ommer]{Rombach_2022_CVPR}
R.~Rombach, A.~Blattmann, D.~Lorenz, P.~Esser, and B.~Ommer.
\newblock High-resolution image synthesis with latent diffusion models.
\newblock In \emph{Proceedings of the IEEE/CVF Conference on Computer Vision
  and Pattern Recognition (CVPR)}, pages 10684--10695, June 2022.

\bibitem[Rudin(2019)]{rudin2019stop}
C.~Rudin.
\newblock Stop explaining black box machine learning models for high stakes
  decisions and use interpretable models instead.
\newblock \emph{Nature machine intelligence}, 1\penalty0 (5):\penalty0
  206--215, 2019.

\bibitem[Ruff et~al.(2021)Ruff, Kauffmann, Vandermeulen, Montavon, Samek,
  Kloft, Dietterich, and M{\"u}ller]{ruff2021unifying}
L.~Ruff, J.~R. Kauffmann, R.~A. Vandermeulen, G.~Montavon, W.~Samek, M.~Kloft,
  T.~G. Dietterich, and K.-R. M{\"u}ller.
\newblock A unifying review of deep and shallow anomaly detection.
\newblock \emph{Proceedings of the IEEE}, 109\penalty0 (5):\penalty0 756--795,
  2021.

\bibitem[Russakovsky et~al.(2015)Russakovsky, Deng, Su, Krause, Satheesh, Ma,
  Huang, Karpathy, Khosla, Bernstein, et~al.]{russakovsky2015imagenet}
O.~Russakovsky, J.~Deng, H.~Su, J.~Krause, S.~Satheesh, S.~Ma, Z.~Huang,
  A.~Karpathy, A.~Khosla, M.~Bernstein, et~al.
\newblock Imagenet large scale visual recognition challenge.
\newblock \emph{International journal of computer vision}, 115:\penalty0
  211--252, 2015.

\bibitem[Samek et~al.(2016)Samek, Binder, on, Lapuschkin, and
  M{\"u}ller]{samek2016evaluating}
W.~Samek, A.~Binder, G.~on, S.~Lapuschkin, and K.-R. M{\"u}ller.
\newblock Evaluating the visualization of what a deep neural network has
  learned.
\newblock \emph{IEEE transactions on Neural Networks and Learning Systems},
  28\penalty0 (11):\penalty0 2660--2673, 2016.

\bibitem[Samek et~al.(2019)Samek, Montavon, Vedaldi, Hansen, and
  M{\"u}ller]{samek2019explainable}
W.~Samek, G.~Montavon, A.~Vedaldi, L.~K. Hansen, and K.-R. M{\"u}ller.
\newblock \emph{Explainable {AI}: interpreting, explaining and visualizing deep
  learning}, volume 11700.
\newblock Springer Nature, 2019.

\bibitem[Samek et~al.(2021)Samek, Montavon, Lapuschkin, Anders, and
  M{\"u}ller]{samek2021explaining}
W.~Samek, G.~Montavon, S.~Lapuschkin, C.~J. Anders, and K.-R. M{\"u}ller.
\newblock Explaining deep neural networks and beyond: A review of methods and
  applications.
\newblock \emph{Proceedings of the IEEE}, 109\penalty0 (3):\penalty0 247--278,
  2021.

\bibitem[Sandler et~al.(2018)Sandler, Howard, Zhu, Zhmoginov, and
  Chen]{sandler2018mobilenetv2}
M.~Sandler, A.~Howard, M.~Zhu, A.~Zhmoginov, and L.-C. Chen.
\newblock Mobilenetv2: Inverted residuals and linear bottlenecks.
\newblock In \emph{Proceedings of the IEEE Conference on Computer Vision and
  Pattern Recognition}, pages 4510--4520, 2018.

\bibitem[Sch{\"o}lkopf et~al.(2001)Sch{\"o}lkopf, Platt, Shawe-Taylor, Smola,
  and Williamson]{scholkopf2001estimating}
B.~Sch{\"o}lkopf, J.~C. Platt, J.~Shawe-Taylor, A.~J. Smola, and R.~C.
  Williamson.
\newblock Estimating the support of a high-dimensional distribution.
\newblock \emph{Neural computation}, 13\penalty0 (7):\penalty0 1443--1471,
  2001.

\bibitem[Schramowski et~al.(2020)Schramowski, Stammer, Teso, Brugger, Herbert,
  Shao, Luigs, Mahlein, and Kersting]{schramowski2020making}
P.~Schramowski, W.~Stammer, S.~Teso, A.~Brugger, F.~Herbert, X.~Shao, H.-G.
  Luigs, A.-K. Mahlein, and K.~Kersting.
\newblock Making deep neural networks right for the right scientific reasons by
  interacting with their explanations.
\newblock \emph{Nature Machine Intelligence}, 2\penalty0 (8):\penalty0
  476--486, 2020.

\bibitem[Selvaraju et~al.(2019)Selvaraju, Cogswell, Das, Vedantam, Parikh, and
  Batra]{Selvaraju_2019}
R.~R. Selvaraju, M.~Cogswell, A.~Das, R.~Vedantam, D.~Parikh, and D.~Batra.
\newblock Grad-cam: Visual explanations from deep networks via gradient-based
  localization.
\newblock \emph{International Journal of Computer Vision}, 128\penalty0
  (2):\penalty0 336–359, 10 2019.
\newblock ISSN 1573-1405.
\newblock \doi{10.1007/s11263-019-01228-7}.

\bibitem[Simonyan and Zisserman(2014)]{simonyan2014very}
K.~Simonyan and A.~Zisserman.
\newblock Very deep convolutional networks for large-scale image recognition.
\newblock \emph{arXiv preprint arXiv:1409.1556}, 2014.

\bibitem[Smilkov et~al.(2017)Smilkov, Thorat, Kim, Vi{\'e}gas, and
  Wattenberg]{smilkov2017smoothgrad}
D.~Smilkov, N.~Thorat, B.~Kim, F.~Vi{\'e}gas, and M.~Wattenberg.
\newblock Smoothgrad: removing noise by adding noise.
\newblock \emph{arXiv preprint arXiv:1706.03825}, 2017.

\bibitem[Stanley(2007)]{stanley2007compositional}
K.~O. Stanley.
\newblock Compositional pattern producing networks: A novel abstraction of
  development.
\newblock \emph{Genetic programming and evolvable machines}, 8:\penalty0
  131--162, 2007.

\bibitem[Sundararajan et~al.(2017)Sundararajan, Taly, and
  Yan]{sundararajan2017axiomatic}
M.~Sundararajan, A.~Taly, and Q.~Yan.
\newblock Axiomatic attribution for deep networks.
\newblock In \emph{International Conference on Machine Learning}, pages
  3319--3328. PMLR, 2017.

\bibitem[Szegedy et~al.(2013)Szegedy, Zaremba, Sutskever, Bruna, Erhan,
  Goodfellow, and Fergus]{szegedy2013intriguing}
C.~Szegedy, W.~Zaremba, I.~Sutskever, J.~Bruna, D.~Erhan, I.~Goodfellow, and
  R.~Fergus.
\newblock Intriguing properties of neural networks.
\newblock \emph{arXiv preprint arXiv:1312.6199}, 2013.

\bibitem[Szegedy et~al.(2016)Szegedy, Vanhoucke, Ioffe, Shlens, and
  Wojna]{szegedy2016rethinking}
C.~Szegedy, V.~Vanhoucke, S.~Ioffe, J.~Shlens, and Z.~Wojna.
\newblock Rethinking the inception architecture for computer vision.
\newblock In \emph{Proceedings of the IEEE Conference on Computer Vision and
  Pattern Recognition}, pages 2818--2826, 2016.

\bibitem[Thomee et~al.(2016)Thomee, Shamma, Friedland, Elizalde, Ni, Poland,
  Borth, and Li]{thomee2016yfcc100m}
B.~Thomee, D.~A. Shamma, G.~Friedland, B.~Elizalde, K.~Ni, D.~Poland, D.~Borth,
  and L.-J. Li.
\newblock Yfcc100m: The new data in multimedia research.
\newblock \emph{Communications of the ACM}, 59\penalty0 (2):\penalty0 64--73,
  2016.

\bibitem[Tiddi et~al.(2020)]{tiddi2020directions}
I.~Tiddi et~al.
\newblock Directions for explainable knowledge-enabled systems.
\newblock \emph{Knowledge Graphs for eXplainable Artificial intelligence:
  Foundations Applications and Challenges}, 47:\penalty0 245, 2020.

\bibitem[Tran et~al.(2018)Tran, Li, and Madry]{tran2018spectral}
B.~Tran, J.~Li, and A.~Madry.
\newblock Spectral signatures in backdoor attacks.
\newblock \emph{Advances in Neural Information Processing Systems}, 31, 2018.

\bibitem[Trozzi et~al.(2021)Trozzi, Wang, and Tao]{trozzi2021umap}
F.~Trozzi, X.~Wang, and P.~Tao.
\newblock Umap as a dimensionality reduction tool for molecular dynamics
  simulations of biomacromolecules: a comparison study.
\newblock \emph{The Journal of Physical Chemistry B}, 125\penalty0
  (19):\penalty0 5022--5034, 2021.

\bibitem[Van~der Maaten and Hinton(2008)]{van2008visualizing}
L.~Van~der Maaten and G.~Hinton.
\newblock Visualizing data using t-sne.
\newblock \emph{Journal of machine learning research}, 9\penalty0 (11), 2008.

\bibitem[Vidovic et~al.(2015)Vidovic, G{\"o}rnitz, M{\"u}ller, R{\"a}tsch, and
  Kloft]{vidovic2015opening}
M.~M.-C. Vidovic, N.~G{\"o}rnitz, K.-R. M{\"u}ller, G.~R{\"a}tsch, and
  M.~Kloft.
\newblock Opening the black box: Revealing interpretable sequence motifs in
  kernel-based learning algorithms.
\newblock In \emph{Joint European Conference on Machine Learning and Knowledge
  Discovery in Databases}, pages 137--153. Springer, 2015.

\bibitem[Vidovic et~al.(2016)Vidovic, G{\"o}rnitz, M{\"u}ller, and
  Kloft]{vidovic2016feature}
M.~M.-C. Vidovic, N.~G{\"o}rnitz, K.-R. M{\"u}ller, and M.~Kloft.
\newblock Feature importance measure for non-linear learning algorithms.
\newblock \emph{arXiv preprint arXiv:1611.07567}, 2016.

\bibitem[Wallis and Buvat(2022)]{wallis2022clever}
D.~Wallis and I.~Buvat.
\newblock Clever hans effect found in a widely used brain tumour mri dataset.
\newblock \emph{Medical Image Analysis}, 77:\penalty0 102368, 2022.

\bibitem[Wang et~al.(2021)Wang, Wu, Zhang, He, and Chua]{wang2021towards}
X.~Wang, Y.~Wu, A.~Zhang, X.~He, and T.-S. Chua.
\newblock Towards multi-grained explainability for graph neural networks.
\newblock \emph{Advances in Neural Information Processing Systems},
  34:\penalty0 18446--18458, 2021.

\bibitem[Wang(2021)]{kaggleCIFAR100Resnet}
Y.~Wang.
\newblock {C}{I}{F}{A}{R}-100 {R}esnet {P}y{T}orch 75.17\% {A}ccuracy ---
  kaggle.com.
\newblock
  \url{https://www.kaggle.com/code/yiweiwangau/cifar-100-resnet-pytorch-75-17-accuracy},
  2021.
\newblock [Accessed 08-Jan-2023].

\bibitem[Weiss et~al.(2016)Weiss, Khoshgoftaar, and Wang]{weiss2016survey}
K.~Weiss, T.~M. Khoshgoftaar, and D.~Wang.
\newblock A survey of transfer learning.
\newblock \emph{Journal of Big data}, 3\penalty0 (1):\penalty0 1--40, 2016.

\bibitem[Wightman(2019)]{rw2019timm}
R.~Wightman.
\newblock Pytorch image models.
\newblock \url{https://github.com/rwightman/pytorch-image-models}, 2019.

\bibitem[Wu et~al.(2019)Wu, Poh~Sheng, Su-En, Chevrier, Jie~Hua, Kiat~Hon, and
  Chen]{wu2019comparison}
D.~Wu, J.~Y. Poh~Sheng, G.~T. Su-En, M.~Chevrier, J.~L. Jie~Hua, T.~L.
  Kiat~Hon, and J.~Chen.
\newblock Comparison between umap and t-sne for multiplex-immunofluorescence
  derived single-cell data from tissue sections.
\newblock \emph{BioRxiv}, page 549659, 2019.

\bibitem[Wu and Palmer(1994)]{wu1994verb}
Z.~Wu and M.~Palmer.
\newblock Verb semantics and lexical selection.
\newblock \emph{arXiv preprint cmp-lg/9406033}, 1994.

\bibitem[Xiao et~al.(2020)Xiao, Engstrom, Ilyas, and Madry]{xiao2020noise}
K.~Xiao, L.~Engstrom, A.~Ilyas, and A.~Madry.
\newblock Noise or signal: The role of image backgrounds in object recognition.
\newblock \emph{arXiv preprint arXiv:2006.09994}, 2020.

\bibitem[Xu et~al.(2019)Xu, Uszkoreit, Du, Fan, Zhao, and
  Zhu]{xu2019explainable}
F.~Xu, H.~Uszkoreit, Y.~Du, W.~Fan, D.~Zhao, and J.~Zhu.
\newblock Explainable {AI}: A brief survey on history, research areas,
  approaches and challenges.
\newblock In \emph{CCF International Conference on natural language processing
  and Chinese computing}, pages 563--574. Springer, 2019.

\bibitem[Yuan et~al.(2021)Yuan, Yan, Sonka, and Yang]{Yuan2021auc}
Z.~Yuan, Y.~Yan, M.~Sonka, and T.~Yang.
\newblock Large-scale {R}obust {D}eep {AUC} {M}aximization: A new surrogate
  loss and empirical studies on medical image classification.
\newblock pages 3020--3029, 10 2021.
\newblock \doi{10.1109/ICCV48922.2021.00303}.

\bibitem[Zech et~al.(2018)Zech, Badgeley, Liu, Costa, Titano, and
  Oermann]{zech2018variable}
J.~R. Zech, M.~A. Badgeley, M.~Liu, A.~B. Costa, J.~J. Titano, and E.~K.
  Oermann.
\newblock Variable generalization performance of a deep learning model to
  detect pneumonia in chest radiographs: a cross-sectional study.
\newblock \emph{PLoS medicine}, 15\penalty0 (11):\penalty0 e1002683, 2018.

\bibitem[Zeiler and Fergus(2014)]{zeiler2014visualizing}
M.~D. Zeiler and R.~Fergus.
\newblock Visualizing and understanding convolutional networks.
\newblock In \emph{European Conference on Computer Vision}, pages 818--833.
  Springer, 2014.

\bibitem[Zhuang et~al.(2020)Zhuang, Qi, Duan, Xi, Zhu, Zhu, Xiong, and
  He]{zhuang2020comprehensive}
F.~Zhuang, Z.~Qi, K.~Duan, D.~Xi, Y.~Zhu, H.~Zhu, H.~Xiong, and Q.~He.
\newblock A comprehensive survey on transfer learning.
\newblock \emph{Proceedings of the IEEE}, 109\penalty0 (1):\penalty0 43--76,
  2020.

\end{thebibliography}
